\documentclass{ieeeaccess}
\PassOptionsToPackage{figuresleft}{rotating}
\usepackage{cite}
\usepackage{amsmath,amssymb,amsfonts}
\usepackage{algorithmic}
\usepackage{graphicx}
\usepackage{textcomp}
\usepackage{array}
\newcolumntype{?}[1]{!{\vrule width #1}}
\usepackage{arydshln}
\usepackage{makecell}
\usepackage[utf8]{inputenc}
\usepackage[T1]{fontenc}
\usepackage{hyperref}
\usepackage{cleveref}
\usepackage[hyperpageref]{backref}
\usepackage{enumitem}
\usepackage{rotating}
\usepackage{multirow}
\usepackage{color}
\usepackage{url}
\usepackage{comment}
\usepackage{mathabx}
\ifCLASSOPTIONcompsoc
  \usepackage[caption=false,font=footnotesize,labelfont=sf,textfont=sf]{subfig}
\else
  \usepackage[caption=false,font=footnotesize]{subfig}
\fi
\def\BibTeX{{\rm B\kern-.05em{\sc i\kern-.025em b}\kern-.08em
    T\kern-.1667em\lower.7ex\hbox{E}\kern-.125emX}}
\begin{document}
\history{Date of publication 2018 00, 0000, date of current version 2018 00, 0000.}
\doi{10.1109/ACCESS.2018.2890150.DOI}

\title{FPGA-based Accelerators of Deep Learning Networks for Learning and Classification: A Review}
\author{
\uppercase{Ahmad Shawahna}\authorrefmark{1},
\uppercase{Sadiq M. Sait}\authorrefmark{1, 2}, \IEEEmembership{Senior Member, IEEE},
\uppercase{and Aiman El-Maleh}\authorrefmark{1}, \IEEEmembership{Member, IEEE}}
\address[1]{Department of Computer Engineering, King Fahd University of Petroleum \& Minerals, Dhahran-31261, Saudi Arabia}
\address[2]{Center for Communications and IT Research, Research Institute, King Fahd University of Petroleum \& Minerals,
Dhahran-31261, Saudi Arabia}
\tfootnote{This work was supported by the King Fahd University of Petroleum \& Minerals, Dhahran, Saudi Arabia.
}
\markboth
{Shawahna \headeretal: FPGA-based Accelerators of Deep Learning Networks for Learning and Classification: A Review}
{Shawahna \headeretal: FPGA-based Accelerators of Deep Learning Networks for Learning and Classification: A Review}

\corresp{Corresponding author: Sadiq M. Sait  (e-mail: sadiq@kfupm.edu.sa).}

\begin{abstract}
Due to recent advances in digital technologies, and availability of credible data, an area of artificial intelligence, deep learning, has emerged, and has demonstrated its ability and effectiveness in solving complex learning problems not possible before. In particular, convolution neural networks (CNNs) have demonstrated their effectiveness in image detection and recognition applications. However, they require intensive CPU operations and memory bandwidth that make general CPUs fail to achieve desired performance levels. Consequently, hardware accelerators that use application specific integrated circuits (ASICs), field programmable gate arrays (FPGAs), and graphic processing units (GPUs) have been employed to improve the throughput of CNNs. More precisely, FPGAs have been recently adopted for accelerating the implementation of deep learning networks due to their ability to maximize parallelism as well as due to their energy efficiency. In this paper, we review  recent existing techniques for accelerating deep learning networks on FPGAs. We highlight the key features employed by the various  techniques for improving the acceleration performance. In addition, we provide recommendations for enhancing the utilization of FPGAs for CNNs acceleration. The techniques investigated in this paper represent the recent trends in FPGA-based accelerators of deep learning networks. Thus, this review is expected to direct the future advances on efficient hardware accelerators and to be useful for deep learning researchers.
\end{abstract}

\begin{keywords}
Adaptable Architectures, 
Convolutional Neural Networks (CNNs), 
Deep Learning,
Dynamic Reconfiguration,
Energy-Efficient Architecture,
Field Programmable Gate Arrays (FPGAs), 
Hardware Accelerator, 
Machine Learning, 
Neural Networks, 
Optimization, 
Parallel Computer Architecture,
Reconfigurable Computing.
\end{keywords}

\titlepgskip=-15pt
\maketitle


\section{Introduction}\label{sec:introduction}
\IEEEPARstart{I}{n}
recent years, due to the availability of massive amounts of credible data (Big Data: Text, Video, Audio, etc.), and tremendous advances in the area of digital electronics technologies that  provide immense computing power, there has been a revival in the area of artificial intelligence (AI), particularly in the area of deep learning (DL)~\cite{bengio2009learning, schmidhuber2015deep, goodfellow2016deep}, a sub-field of machine learning (ML).

The field of DL emerged in 2006 after a long pause in the area of neural networks (NNs) research~\cite{zhang2018deep}. A key aspect in DL is that the networks and/or their weights are {\it not} designed by human beings. Instead, they  are learned from data using a general purpose learning procedure~\cite{rumelhart1986learning, rumelhart1988neurocomputing}.

While ML uses algorithms to parse and learn from data, to make informed decisions, DL structures algorithms in layers to create an artificial neural network (ANN) that can learn, and similar to human intelligence, can make accurate decisions on its own~\cite{nielsen2015neural}. Therefore, instead of designing algorithms by hand, systems can be built and trained to implement concepts in a way similar to what comes naturally to humans, and with accuracy sometimes {\it exceeding} human-level performance~\cite{weyand2016planet, Mathworks2018}.

In DL, each layer is designed to detect features at different levels. A layer transforms the representation at one level (starting from input data which maybe images, text, or sound) to a representation at a higher, slightly more abstract level~\cite{lecun2015deep}. For example, in image recognition, where input initially comes in the form of pixels, the first layer   detects low level features such as edges and curves. The output of the first layer becomes input to the second layer which produces higher level features, for example semi-circles, and squares~\cite{AditDeshpande}. The next layer assembles   the output of the previous layer to parts of familiar objects, and a subsequent layer detects the objects.  As we go through more layers, the network yields an   activation map that represents more and more complex features.   The  deeper you go into the network, the filters begin to be more responsive to a larger region of the pixel space. Higher level layers amplify aspects of the received inputs that are important for discrimination and suppress irrelevant variations.

\subsection{Applications of Deep Learning Networks}
With the now widely used convolution neural networks (CNNs)~\cite{dayhoff1990neural, lecun1995convolutional} and deep neural networks (DNNs)~\cite{hauswald2015djinn, yue2015beyond}, 
it is now possible to solve problems in domains where knowledge is not easily expressed explicitly and implicit information is stored in the raw data. Solutions to multifarious problems in the domain of sciences, business, etc., have been possible that were not conceivable for several years, in spite of best attempts by the AI community. This has been primarily possible due to the excellent ability of deep learning in discovering intricate structures in high-dimensional data. Examples include character recognition~\cite{lecun1990handwritten}, gesture recognition~\cite{barros2014multichannel}, speech recognition (e.g., in Google Now, Siri, or click-through prediction on an advertisement)~\cite{graves2013speech, huang2013learning, abdel2014convolutional}, document processing~\cite{simard2003best, lai2015recurrent, kim2014convolutional}, natural language processing~\cite{collobert2008unified, sarikaya2014application}, video classification~\cite{karpathy2014large}, image classification~\cite{mutch2006multiclass, krizhevsky2012imagenet, simonyan2014very,  russakovsky2015imagenet, szegedy2015going, ren2015faster}, face detection and recognition~\cite{korekado2005image, li2015convolutional}, robot navigation~\cite{muller2006off, hadsell2007multi, sermanet2009multirange}, real-time multiple object tracking~\cite{blanco2018deep}, financial forecasting~\cite{mcnelis2005neural}, and medical diagnosis systems~\cite{lisboa2000artificial, mirowski2008comparing, dahl2013improving}, to name a few. 

Other recent areas of applications include automated driving (e.g., learning to detect stop signs, traffic lights, pedestrians, etc.), aerospace and defense (e.g., identify objects from satellites and identify safe or unsafe zones), medical research (e.g., in identification of cancer cells), industrial automation (e.g., to improve worker safety by detecting when people or objects are within an unsafe distance of machines), and electronics (used in automated hearing, speech translation, etc.)~\cite{hadsell2009learning, deng2014deep, girshick2014rich, wu2017squeezedet, Mathworks2018}.

\subsection{Emergence of Deep Learning Networks}
Convolutional neural networks are considered as one of the most influential innovations in the field of computer vision~\cite{he2015delving}. The success of deep learning networks grew to prominence in 2012 when Krizhevsky et al.~\cite{krizhevsky2012imagenet} utilized CNNs to win the annual olympics of computer vision, ImageNet large-scale vision recognition challenge (ILSVRC)~\cite{russakovsky2015imagenet}. 
Using AlexNet model, they achieved an astounding improvement   as the image classification error dropped from 26\% (in 2011) to 15\%. 
ImageNet is a standard benchmark dataset used to evaluate the performance of object detection and image classification algorithms. It consists of millions of different images distributed over tens of thousands of object classes.

CNNs have achieved even better accuracy in classification and various computer vision tasks. The classification accuracy in ILSVRC improved to 88.8\%~\cite{zeiler2014visualizing}, 93.3\%~\cite{szegedy2015going}, and 96.4\%~\cite{he2016deep} in the 2013, 2014, and 2015 competitions, respectively. Fig.~\ref{ImageNet-Winning} shows the accuracy loss for the winners of ImageNet competitions before and after the emergence  of deep learning algorithms.

\Figure[t](topskip=0pt, botskip=0pt, midskip=0pt)[width=0.48\textwidth]{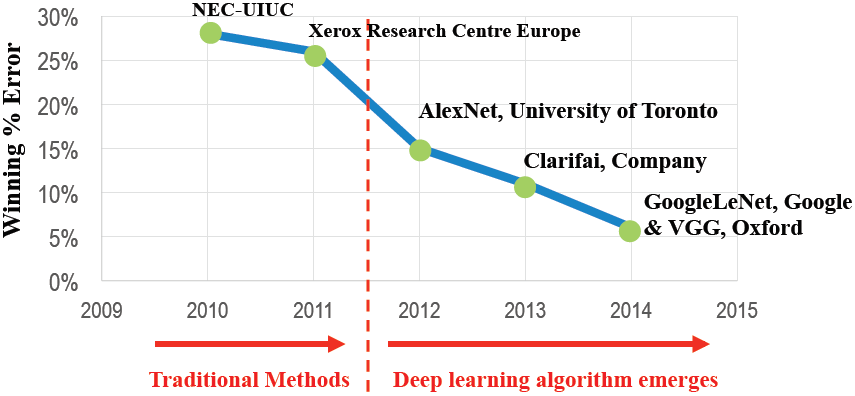}
{ImageNet Competition Results~\cite{ImageNet2018ILSVRC}.
\label{ImageNet-Winning}}

Thereafter, large host companies started using CNNs at the core of their services. Google, Microsoft, Facebook, Amazon, Pinterest, and Instagram are currently using neural networks for their photo search, Bing's image feeds, automatic tagging algorithms, product recommendations, home feed personalization, and for their search infrastructure, respectively~\cite{AditDeshpande}. However, the classic use-case of CNNs is for image and speech processing~\cite{mohamed2012acoustic}. 

A typical CNN is a multi-layered feed-forward ANN with a pipeline-like architecture. Specifically, each layer performs a well-known computation on the outputs of the previous layer to generate the inputs for the next layer. In general, CNNs have two types of inputs; the data to be tested or classified (also named as feature maps), and the weights. Images, audio files, and recorded videos are examples of the input data to be classified using CNNs. On the other hand, the network weights are the data generated from training the CNN on a dataset containing similar inputs to the one being tested.

\subsection{Hardware Acceleration of Deep Learning Networks}
To provide more accurate results as well as real-time object recognition, for example in   applications such as robots and auto-piloted cars, the size of the convolution neural network needs to be increased by adding  more neural network layers~\cite{krizhevsky2012imagenet}. However, evolving more and new type of NN layers results in more complex CNN structures as well as high depth CNN models. Thus, billions of operations and millions of parameters, as well as substantial computing resources are required to train and evaluate the resultant large-scale 
CNN~\cite{nomura2007projection, chilimbi2014project, szegedy2015going}. Such requirements represent a computational challenge for general purpose processors (GPP). Consequently, hardware accelerators such as application specific integrated circuit (ASIC), field programmable gate array (FPGA), and graphic processing unit (GPU) have been employed to improve the throughput of the CNN. In practice, CNNs are trained off-line using the back-propagation process~\cite{lecun1989backpropagation}. Then, the off-line trained CNNs are used to perform recognition tasks using the feed-forward process~\cite{zhang2015optimizing}. Therefore, the speed of feed-forward process is what matters. 

GPUs are the most widely used hardware accelerators for improving both training and classification processes in CNNs~\cite{yazdanbakhsh2015neural}. This is due to their high memory bandwidth and throughput as they are highly efficient in floating-point matrix-based operations~\cite{hinton2012deep, jia2014caffe, vasudevan2017parallel}. However, GPU accelerators consume a large amount of power. Therefore, their use in CNN-based applications implemented as a cloud service on large servers or in battery operated devices becomes a challenge. Furthermore, GPUs gain their performance from their ability to process a large image batch in parallel. For some applications like a video stream, input images should be processed frame by frame as the latency of the result of each frame is critical to the application's performance. For some tracking algorithms, the result of one frame affects the process of the next frame \cite{guo2018angel}.   
Nurvitadhi et al.~\cite{nurvitadhi2017can}  recently evaluated emerging DNN algorithms on latest generations of GPUs (i.e., NVIDIA Titan X Pascal) and FPGAs (i.e., Intel Arria 10 GX 1150 and Intel Stratix 10 2800). The experimental results show that current trends in deep neural networks favor FPGA platforms as they offer higher power efficiency (a.k.a., performance per Watt).

FPGA and ASIC hardware accelerators have relatively limited memory, I/O bandwidths, and computing resources compared with GPU-based accelerators. However, they can achieve at least moderate performance with lower power consumption~\cite{misra2010artificial}. The throughput of ASIC design can be improved by customizing memory hierarchy and assigning dedicated resources~\cite{esmaeilzadeh2012neural}. However, the development cycle, cost, and flexibility are not satisfactory in ASIC-based acceleration of deep learning networks~\cite{han2016eie, du2018reconfigurable}. As an alternative, FPGA-based accelerators are currently in use to provide high throughput at a reasonable price with low power consumption and reconfigurability~\cite{vanderbauwhede2013high, putnam2014reconfigurable}. The availability of high-level synthesis (HLS) tools, using C or C++, from FPGA vendors lowers the programming hurdle and shortens the development time of FPGA-based hardware accelerators~\cite{liang2012high, cong2011high, canis2011legup}.

Convolutional neural networks have a very useful property, that is, each feature map neuron shares its weights with all other neurons~\cite{lecun1998gradient}. The authors in~\cite{hameed2010understanding, keckler2011gpus} proved that the highest energy expense results from accessing the off-chip DRAM memory for data movement rather than computation. In other words, the energy cost of the increased memory accesses and data movement due to the large number of CNN operations  often exceeds the energy cost of computation~\cite{han2016eie, chen2016eyeriss}. Thus, CNN accelerators need to carefully consider this to achieve efficient architecture in terms of time and power.

In this paper, we review the current status of using FPGAs as accelerators for implementing deep learning networks.
We highlight the implementation challenges and design directions used to tackle those challenges. We also provide
future recommendations  to maximize the performance of FPGAs as accelerators for deep learning networks
and simplify their use.

The remainder of the paper is organized as follows. Section \ref{sec:Background_and_Terminology} provides background information about CNNs, their key operations, and some well-known deep learning networks. In addition, it introduces the basic structure of FPGAs and highlights their features enabling them to accelerate computationally intensive applications. It also discusses the implementation challenges of deep learning networks on FPGAs and how these challenges can be overcome. Section \ref{sec:Acceleration_of_Deep_Learning_Networks_Current_Status} reviews existing CNNs compression techniques and presents the current status of accelerating deep learning networks using ASIC-based and FPGA-based accelerators. Section \ref{sec:Metaheuristics_in_the_Design_of_Convolutional_Neural_Networks} describes the use of metaheuristics in the design and optimization of CNNs implementation. Section \ref{sec:Summary_and_Recommendations} summarizes existing design approaches for accelerating deep learning networks and provides recommendations for future directions that will simplify the use of FPGA-based accelerators and enhance their performance. Finally, section \ref{sec:Conclusion} concludes the paper.

\section{Background and Terminology} \label{sec:Background_and_Terminology}

This section gives an overview of the key  operations and terminology used in convolutional neural networks (CNNs) and provides  examples of well-known deep learning networks. In addition, it illustrates the basic structure of field programmable gate arrays (FPGAs) and how deep learning methods can benefit from the capabilities of FPGAs. The last subsection highlights the challenges of implementing deep learning networks on FPGAs.

\subsection{Convolutional Neural Networks (CNNs)}

In this subsection, we describe the key operations and terminology involved in the construction of CNNs including convolution, 
activation functions, normalization, pooling, and characteristics of fully connected layers.

\subsubsection{Convolution (CONV)} \label{Convolution (CONV)}

A convolution operation can be thought of as the production of a matrix smaller in size than the original image matrix, representing pixels, by sliding a small window (called filter, feature identifier, or kernel) of size $k \times k$ over the image (called input feature map (FM)), to produce an output feature neuron value~\cite{serre2007robust}. The filter is an array of numbers called weights or parameters. These weights are computed during the training phase. As the filter slides over the feature map, it multiplies the values in the filter with the original pixel values, that is, it first performs element-wise multiplication, and then sums the products, to produce a single number. The inputs and outputs of the CONV layer are a series of FM arrays.

This operation, starting from the top left corner of the FM, is repeated by moving the window \textit{S} strides at a time,
first in the right direction, until the end of the FM is reached, and then proceeding downwards until the FM is completely scanned and all the elements of the FM are covered. The sliding of the filter window and performing the operation is known by the verb convolving, hence the noun \textit{convolution}~\cite{AditDeshpande, prateekvjoshi2016}. Normally, the size of the kernel is very small, less than or equals to $11 \times 11$. Each output-input FM pair has a set of weights equal to the kernel size and each output FM is computed based on the sum of  the convolution operations performed on all input FMs. Note that different CONV layers in the same CNN model vary considerably in their sizes.

In summary, the convolution operation comprises four levels of loops; the output FMs loop (\textit{Loop-4}), the loop across the input FMs (\textit{Loop-3}), the loop along the dimensions of a single input FM (scan operation, \textit{Loop-2}), and the kernel window size loop (multiply-and-accumulate (MAC) operation, \textit{Loop-1}). CONV layers are dominant in CNN algorithms since they often constitute more than 90\% of the total CNN operations~\cite{krizhevsky2012imagenet,  simonyan2014very, cong2014minimizing, he2016deep, chen2016eyeriss, ma2016scalable}. Therefore, many attempts have been made to speedup CONV operations using loop unrolling technique~\cite{bacon1994compiler, zhang2015optimizing}, as will be discussed later. Loop unrolling maximizes the parallelism of CONV MACs computation which requires a special consideration of processing elements (PEs) and register arrays architecture. Fig.~\ref{CONV_Loops_Unrolling} illustrates the loop unrolling of CONV loops levels.

\subsubsection{Activation Functions (AFs)}

Activation function in neural networks is similar to {\em action potential} in animal cells such as neurons. A neuron is said to {\em fire} if it emits an action potential. 
A popularly used activation function is the {\it sigmoid} function which can be expressed as
\begin{equation}
f(x) = 1/(1+e^{-x})
\end{equation}
where $x$ represents the weighted sum of the neuron inputs and if it is a sufficiently large positive number, the {\it sigmoid} function approximates to unity. For sufficiently large negative values of $x$, the {\it sigmoid} function is close to 0. Another popular activation function is 

\begin{equation}
f(x) = tanh(x)
\end{equation}

The above standard {\it sigmoid} and {\it tanh} non-linear functions require long training time~\cite{krizhevsky2012imagenet}. A recently proposed and commonly used AF in CNNs is rectified linear unit (ReLU) which is defined as 

\begin{equation}
f(x) = max(x, 0)
\end{equation}

{\em ReLU} activation function is known to converge faster in training, and has lesser computational complexity~\cite{suda2016throughput, denil2013predicting} than standard {\it sigmoid} and {\it tanh} functions. In addition, it does not require input normalization to prevent it from saturating~\cite{ krizhevsky2012imagenet,suda2016throughput,nair2010rectified}.

\Figure[t](topskip=0pt, botskip=0pt, midskip=0pt)[width=0.45\textwidth, height = 5.4 cm]{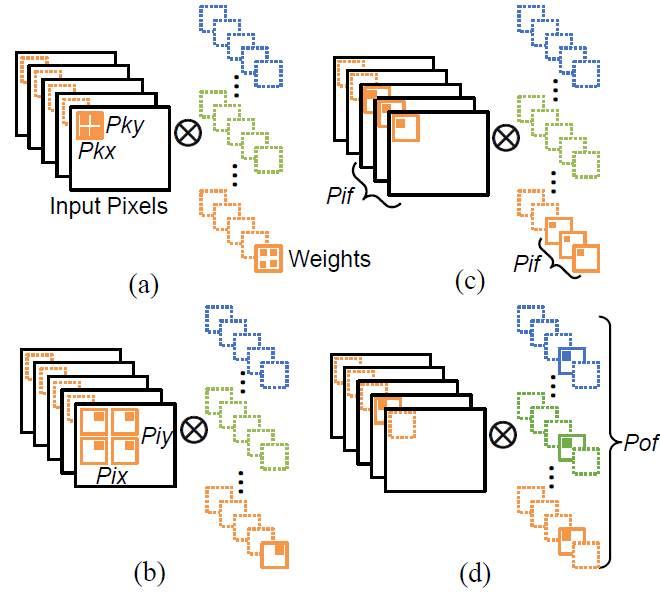}
{CONV Loops Unrolling~\cite{ma2017optimizing}: (a) Unrolling \textit{Loop-1}; (b) Unrolling \textit{Loop-2}; (c) Unrolling \textit{Loop-3}; (d) Unrolling \textit{Loop-4}, where, $Pkx$, $Pky$, $Pix$, $Piy$, $Pif$, and $Pof$ are loop unrolling design variables for the kernel window width, kernel window height, input FM width, input FM height, number of input FMs, and the number of output FMs, respectively.
\label{CONV_Loops_Unrolling}}

\subsubsection{Normalization}

In real life, a phenomenon called ‘lateral inhibition’ appears, which refers to the capacity of an excited neuron to subdue its neighbors, thereby creating a contrast in that area.  In CNNs, to accomplish this, local response normalization (LRN) or simply {\it normalization}  is used, particularly when dealing with ReLU neurons, because they have unbounded activation that needs normalization.
It detects high frequency features with a large response. If we normalize around the local neighborhood of the excited neuron, it becomes even more sensitive as compared to its neighbors. At the same time, it will dampen the responses that are uniformly large in any given local neighborhood. If all the values are large, then normalizing those values will diminish all of them. So, basically it performs some kind of inhibition and boosts the neurons with relatively larger activations.

Normalization can be done within the same feature or across neighboring features by a factor that depends on the neighboring neurons. Expressions to compute the response normalized activity can be found in~\cite{krizhevsky2012imagenet,suda2016throughput}.

\Figure[t](topskip=0pt, botskip=0pt, midskip=0pt)[width=0.68\textwidth]{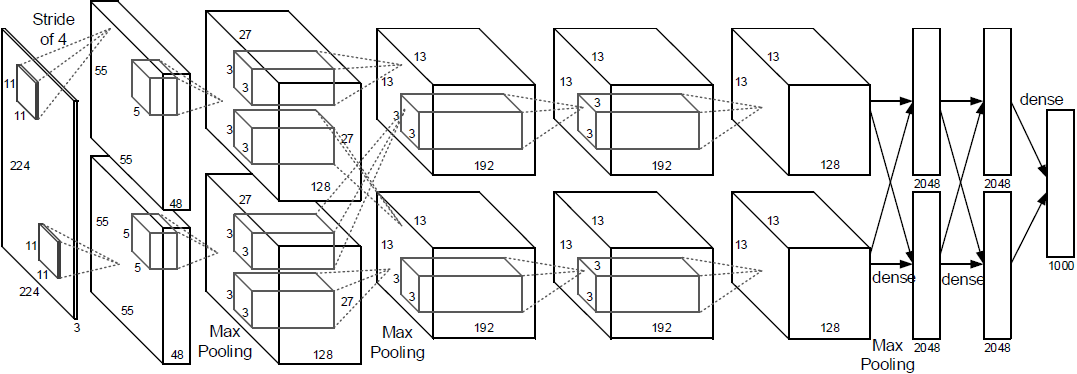}
{AlexNet CNN Architecture~\cite{krizhevsky2012imagenet}.
\label{AlexNet-Architecture}}

\subsubsection{Pooling}

Pooling, also known as {\it subsampling}, is employed to progressively reduce the spatial size of the representation, thereby reducing the amount of parameters and computation in the network.  Pooling layers are periodically inserted in between successive convolutional layers. They operate independently on every depth slice of the input and resize it spatially using the {\em MAX} operation. The most common form is a pooling layer with filters of size $2\times2$ applied where the {\em MAX} operation would be taking a maximum over $4$ samples thereby discarding $75$ percent of the activations~\cite{Karpathy2018CNNS}.
In addition to the popular {\em MAX} pooling, the pooling units in some CNNs are also used to perform other functions, such as {\em AVG} and {\em MIN} operations~\cite{suda2016throughput}.

\subsubsection{Fully Connected Layer (FC)}
A common form of a convolutional neural network architecture comprises stacks of a few convolutional and ReLU layers, followed by layers for pooling, and this pattern is repeated until the image has merged spatially to a small size. This is followed by one or more fully connected layers, also known as  inner-product layers, whose neurons have full connections to all activations in the previous layer, hence the name.
The last fully connected layer is the classification layer and it holds the output such as
the class scores~\cite{suda2016throughput}.

\subsection{Examples of Deep  Learning  Networks}

We list in this subsection some of the well-known deep learning networks.

\begin{itemize}

\item \textbf{AlexNet (2012)} is a convolutional neural network consisting of $5$ convolutional layers, interspersed by $2$ normalization layers, as well as $3$ fully connected layers~\cite{krizhevsky2012imagenet}. Each convolutional layer performs the activation function using ReLU. In addition, $3$ pooling layers are employed with the first, second, and last convolutional layers. The architecture of AlexNet CNN is shown in Fig.~\ref{AlexNet-Architecture}. AlexNet won the 2012 ImageNet challenge by classifying $224 \times 224$ input color images to 1,000 different output classes.

\item \textbf{VGG (2014)} is a convolutional neural network model similar to AlexNet in terms of the number of  fully connected layers. However, it consists of $5$ groups of convolutional layers~\cite{denil2013predicting, simonyan2014very}. The exact number of CONV layers in each group depends on the version of the VGG, visual geometry group, model. Table~\ref{tab_VGG} shows the number of CONV and FC layers for the most commonly used VGG models.

\item \textbf{ResNets (2016)} are  deep residual networks   with  extremely irregular and complex structures compared to AlexNet and VGG CNN models~\cite{he2016deep, he2016identity, szegedy2017inception}. This is due to  having more types of layers, where non-adjacent layers incorporate shortcuts to compute the residual functions, as well as having highly deep structures, that is,  between 50 and  1000  CONV layers. Unlike AlexNet and VGG models where the layers are connected in sequence, the interconnections in  ResNet layers are in the form of a directed acyclic graph (DAG).   ResNet-50 and ResNet-152 are widely used, especially for image classification.  ResNet-50/152 structure contains 53/155 CONV (most of them are followed by batch normalization (BatchNorm), scale, and ReLU layers), 1/1 MAX pooling, 1/1 Average pooling, 1/1 FC, and, 16/50 element-wise (Eltwise) layers, respectively.
\end{itemize}

\begin{table}[b]
	\renewcommand{\arraystretch}{1.2}
	\centering
	\captionsetup{justification=centering}
	\caption{CNN Layers for VGG Models.}
	\begin{tabular}{|l|c|c|c|}
		\hline
		\multicolumn{1}{|c|}{Layers} & VGG-11 & VGG-16 & VGG-19\\
		\hline
		\hline
		CONV (Group 1) & 1 & 2 & 2\\
		\hline
		CONV (Group 2) & 1 & 2 & 2\\
		\hline
		CONV (Group 3) & 2 & 3 & 4\\
		\hline
		CONV (Group 4) & 2 & 3 & 4\\
		\hline
		CONV (Group 5) & 2 & 3 & 4\\
		\hline
		CONV (Total) & \textbf{8} & \textbf{13} & \textbf{16}\\
		\hline
		FC & \textbf{3} & \textbf{3} & \textbf{3}\\
		\hline
		\hline
		\multicolumn{1}{|c|}{\textbf{Total}} & \textbf{11} & \textbf{16} & \textbf{19}\\
		\hline
	\end{tabular}
	
	\label{tab_VGG}
\end{table}

\subsection{Field Programmable Gate Arrays (FPGAs)}

FPGAs are off-the-shelf programmable devices that provide a flexible platform for implementing custom hardware functionality at a low development cost. They consist mainly of a set of programmable logic cells, called configurable logic blocks (CLBs), a programmable interconnection network, and a set of programmable input and output cells around the device~\cite{villasenor1997configurable}. In addition, they have a rich set of embedded components such as digital signal processing (DSP) blocks which are used to perform arithmetic-intensive operations such as multiply-and-accumulate, block RAMs (BRAMs), look-up tables (LUTs), flip-flops (FFs), clock management unit, high speed I/O links, and others. Fig.~\ref{FPGA-Structure} shows a basic structure of an FPGA.

\Figure[t](topskip=0pt, botskip=0pt, midskip=0pt)[width=0.40\textwidth, height = 4.8 cm]{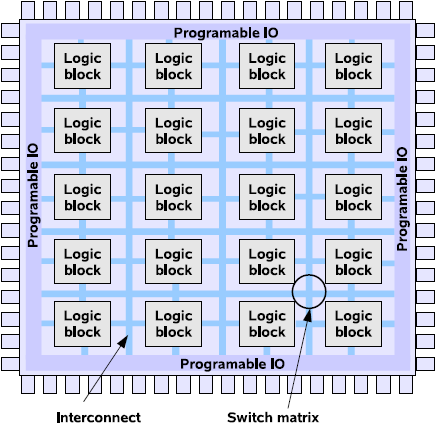}
{FPGA Basic Structure~\cite{villasenor1997configurable}.
\label{FPGA-Structure}}

FPGAs are widely considered as accelerators for computationally-intensive applications as they enable models with highly flexible fine-grained parallelism and associative operations such as broadcast and collective response~\cite{brown2012field}. In~\cite{herbordt2008computing, varma2016architecture}, FPGA computing models used for applications acceleration are presented, including data streaming, associative computing, highly parallel memory access, use of standard hardware structures such as first in first out (FIFO) buffers, stacks and priority queues, and functional parallelism.

FPGAs have the advantage of maximizing performance per Watt of power consumption, reducing costs for large scale operations~\cite{lacey2016deep}. This makes them an excellent choice as accelerators for battery operated devices
and in cloud services on  large  servers. FPGAs have recently been widely used for deep learning acceleration 
given the flexibility in implementing architectures with large degree of parallelism resulting in high execution speeds~\cite{farabet2011large}.

The adoption of software-level programming models such as the open computing language (OpenCL) standard~\cite{munshi2009opencl, stone2010opencl} in FPGA tools made them more attractive to use for deep learning~\cite{omondi2006fpga, waidyasooriya2018design}.
In addition, the feed-forward nature of deep learning algorithms makes FPGAs offer a clear advantage as they can create customized hardware circuits that are deeply pipelined and inherently multithreaded~\cite{lacey2016deep}. FPGAs also have the capability of partial dynamic configuration, which allows part of the FPGA to be reconfigured while the rest is being used. This could be of potential benefit to deep learning methods where the next layer could be reconfigured while the current layer is being used. 

\subsection{Challenges of FPGA-Based Implementation of Deep  Learning  Networks}

Implementation of deep learning networks and, in particular, CNNs on FPGAs has a number of challenges including the requirement of a significant amount of storage, external memory bandwidth, and computational resources on the order of billions of operations per second~\cite{sze2018hardware}. For example,  AlexNet CNN has over 60 million model parameters which need 250MB of memory for storing the weights based on 32-bit floating-point representation as well as requires around 1.5 billion operations for each input image~\cite{suda2016throughput}. This large amount of storage required  is not supported by existing commercial FPGAs and hence the weights have to be stored on external memory and transferred to the FPGA during computation. Without careful implementation of deep learning networks and  maximizing resource sharing, the implementation may not fit on FPGAs due to limited  logic resources.

The problem exacerbates with more complex models such as VGG CNN model which have 16 layers. 
For example, the VGG-16 CNN model has 138 million weights and needs over 30 GOPS~\cite{qiu2016going}.
Although the current trends in implementing CNNs is going toward compressing the entire CNN model with dramatically reducing data bit-width~\cite{han2017ese}, it is expected that future CNN models will get more complex with larger number of layers as the amount of training data continues to grow and the problems to be  solved get more complex.

In addition, different layers in CNNs have different characteristics 
which result in  different parallelism and memory access requirements. 
Different layers in a CNN network exhibit vastly different amounts of intra-output and inter-output parallelism~\cite{chakradhar2010dynamically}. Intra-output parallelism parallelizes the computation of a single output image 
since it is the sum of $n$ input-kernel convolutions. However, inter-output parallelism is based on computing
multiple output FMs in parallel. Furthermore, convolutional layers are computational-centric while fully connected 
layers are memory centric~\cite{qiu2016going}.
For example, the number of operations in each group of convolutional layers in VGG-16 model  are on the 
order of 2 to 9 GOPS while the number of weights are on the order of 0.04 to 7.08 million.
However, the number of operations in  fully connected layers are in the 
order of 0.01 to 0.21 GOPS, while the number of weights are on the order of 4.10 to 102.76 million.
Thus, the developed CNN accelerator must be designed carefully to meet the varying requirements of different layers and
needs to be {\em flexible} to maximize the performance for each CNN layer. 

As  technology advances,  FPGAs continue to grow in size and capabilities. It is crucial to have some mechanisms for addressing the requirements for efficient implementations of deep learning networks. 
Addressing hardware resource limitations requires reuse of  computational resources, and storing of partial results in internal memories.
Data transfer and computational resource usage are significantly impacted by the ordering of operations and selection
of parallelism in the implementation of CNNs on FPGAs. Careful scheduling of operations can result in  significant reduction
in external memory access and internal buffer sizes.
External memory bandwidth requirements can be also decreased by using reduced precision for representing the weights with minimal impact on solution quality, which also results in a better energy efficiency. In addition, the number of external memory accesses can be reduced by utilizing on-chip memory and exploiting data reuse. 
Furthermore, the large number of weights in the fully connected layer
can be reduced, based on utilizing singular value decomposition (SVD)~\cite{van1996matrix} with a small impact on accuracy.
In the next section, we will review various design approaches used to cope with those challenges for implementing deep learning networks.

\section{Acceleration of Deep Learning Networks: Current Status} \label{sec:Acceleration_of_Deep_Learning_Networks_Current_Status}

In this section, we will start by covering convolutional neural networks (CNNs) compression techniques as they have a significant impact on the implementation complexity of CNNs. CNNs compression techniques  target the minimization of the number of operations and the memory footprint with minimal impact  on accuracy. Then, we discuss hardware acceleration techniques for deep learning (DL) algorithms and CNNs based on both application specific integrated circuit (ASIC) and field programmable gate array (FPGA) implementations. In general, hardware accelerators focus on designing specific modules and architectures that ensure data reuse, enhance data locality, and accelerate convolutional (CONV) layer operations based on performing needed operations in parallel. 

\subsection{CNNs Compression}

In this subsection, we review techniques that target the compression of CNNs which 
results in significantly reducing their implementation complexity with minimal impact on accuracy.

Denton et al.~\cite{denton2014exploiting} proposed a technique to reduce the memory footprint for the network weights in object recognition systems. 
They used singular value decomposition (SVD)~\cite{van1996matrix} and filter clustering methods for this purpose. The results for convolutional model of 15 layers in~\cite{zeiler2014visualizing} show that the proposed technique speeds up the operations in convolutional layers by a factor of 2, compared to CPU Eigen3-based library implementation~\cite{guennebaud2015eigen}. In addition, it successfully achieved 13$\times$ memory footprint reduction for the  fully connected layers while preserving the recognition accuracy within 99\%.

In another work, Han et al.~\cite{han2015learning} employed network pruning techniques~\cite{lecun1990optimal, hanson1989comparing, hassibi1993second} to reduce the over-fitting and complexity of neural network models. Their results demonstrated that pruning redundant connections as well as less influential connections achieved 9$\times$ and 13$\times$ compression for AlexNet and VGG-16 models, respectively, while achieving zero accuracy loss for both.

In a subsequent  work, Han et al.~\cite{han2015deep} proposed a deep compression technique for more reduction of the storage requirements of CNNs through the enforcement of weights sharing. Deep compression basically consists of pruning, trained weights quantization, and Huffman coding pipeline stages. The experimental results show that the proposed compression technique successfully reduced the storage requirement of AlexNet and VGG-16 CNN models by 35$\times$ and 49$\times$, respectively,  without affecting their accuracy. This also improved the power efficiency (a.k.a., performance per Watt) by 3$\times$ to 7$\times$.

\subsection{ASIC-based Accelerators}

In this subsection, we present some recent work in the area of hardware-based accelerators (ASICs).

An ASIC-based hardware accelerator referred to as DianNao~\cite{chen2014diannao} was   designed for large-scale convolutional neural networks and deep neural networks. DianNao accelerates neural networks by 
minimizing memory transfers, 
which opened a new paradigm for hardware accelerators. Since the weights are repeatedly used in the computations of convolution layers, frequent memory access can significantly degrade the overall performance. Therefore, the authors exploited the 
locality properties 
of neural network layers to design custom storage structures that take advantages of these properties. In addition, they employed 
dedicated buffers and tiling techniques to reduce the overall external memory traffic through increasing  data locality. 

Chen et al.~\cite{chen2014diannao} also observed that using short fixed-point representation of feature maps (FMs) and weights can also significantly reduce computation resources and memory footprint. They found that the area and power of a 32-bit multiplier can be reduced by a factor of 0.164$\times$ and 0.136$\times$, respectively, using  16-bit multipliers. Consequently, DianNao has been implemented using 65nm fabrication technology with 16-bit fixed-point arithmetic units, 6 bits of which are used for the integer part  and  the remaining 10 for the fractional part.
The experimental results demonstrated that DianNao has an average performance of 452 GOPS with power consumption of 485 mW. The  results depicted that using 16-bit arithmetic units instead of 32-bit ones introduced only 0.26\% accuracy loss on MNIST dataset~\cite{lecun1998mnist}. On the other hand,  the scalability and efficiency of DianNao accelerator are severely limited by the bandwidth constraints of the memory system.

\Figure[t!](topskip=0pt, botskip=0pt, midskip=0pt)[width=0.52\textwidth]{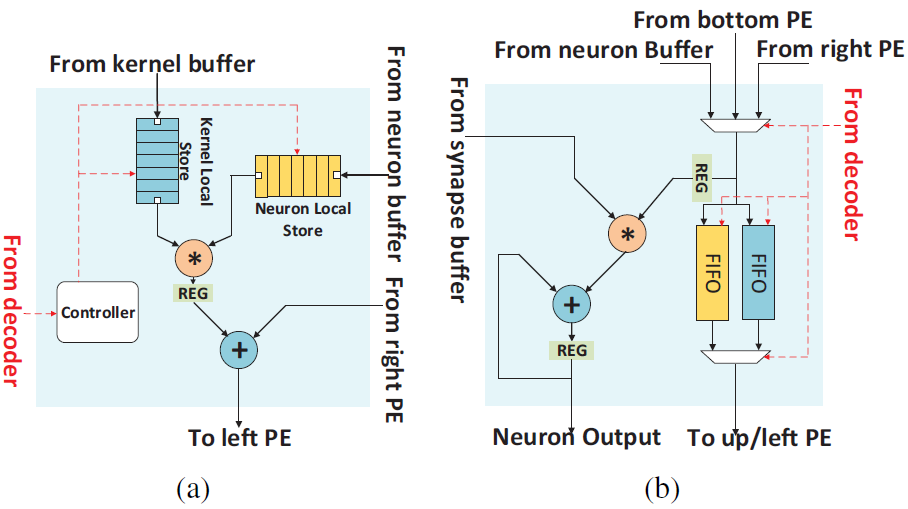}
{Processing Element (PE) Architecture in; (a) FlexFlow, (b) 2D-Mapping \cite{lu2017flexflow}.
\label{FlexFlow_PE_Arch}}

In a related research work, Chen et al.~\cite{chen2014dadiannao, luo2017dadiannao} proposed DaDianNao multi-chip supercomputer which offers sufficient memory capacity suitable for on-chip storage of all weights in CNNs. This  system is mainly important for today's large-scale deployments of sophisticated industry and consumers services. DaDianNao uses 16-bit fixed-point numbers in the inference process like DianNao, but it is implemented using 28nm technology. The results show that DaDianNao outperforms the performance of a single GPU architecture by up to 656.63$\times$ and reduces the average energy consumption by 184.05$\times$ with only 0.01\% accuracy error rate on MNIST dataset for a 64-chip system.

Another member of the DianNao family, called PuDianNao~\cite{liu2015pudiannao}, has been designed using TSMC 65nm process to support multiple techniques and scenarios of machine learning (ML). PuDianNao accelerates different ML techniques through extracting their 
critical locality properties and computational primitives 
with the use of 
on-chip storage 
as well as 7 novel functional units.  Experimental results show that PuDianNao is 1.20$\times$ and  128.41$\times$ faster and energy-efficient, respectively, than NVIDIA K20M GPU architecture. However, both of DaDianNao~\cite{chen2014dadiannao} and PuDianNao architectures have not been optimized to be used for embedded applications.

To improve the scalability and  energy efficiency of DianNao design discussed in~\cite{chen2014diannao},  ShiDianNao accelerator was proposed~\cite{du2015shidiannao}. ShiDianNao is designed especially for real-time object recognition applications such as self-driving cars, smartphones, and security using 65nm CMOS technology. The proposed accelerator directly connects with a CMOS/CCD sensor in the image processing chip. In addition, all the 
weights of CNN layers are stored in SRAM on-chip memory, as the target here is small CNN models. ShiDianNao is embedded inside the processing chip to eliminate off-chip DRAM memory accesses and minimize data movements between the SRAM holding the CNN model and the individual processing elements from the sensor. ShiDianNao has a power consumption of 320.10 mW with a peak performance of 194 GOPS under 1 GHz working frequency. Moreover, ShiDianNao has 1.87$\times$ speedup and is 60$\times$ more energy-efficient than DianNao~\cite{chen2014diannao}.

However, DianNao~\cite{chen2014diannao}, DaDianNao~\cite{chen2014dadiannao, luo2017dadiannao}, PuDianNao~\cite{liu2015pudiannao}, and ShiDianNao~\cite{du2015shidiannao} are not implemented using FPGA or any other reconfigurable hardware. Therefore, they cannot be efficiently adapted to different application demands (i.e., different neural network sizes). In addition, ASIC designs have a long development cycle and lack flexibility for handling varying DL network designs. Finally, CNN accelerators, which store all weights on-chip such as ShiDianNao~\cite{du2015shidiannao}, will not be able to support realistic large-scale CNN models. 

Similar approaches to the DianNao family of techniques are presented in~\cite{sze2017efficient} with similar limitations. ISAAC~\cite{shafiee2016isaac} and PRIME~\cite{chi2016prime} have explored in-memory processing to design an acceleration architecture for neural networks. The proposed ISAAC architecture has achieved better improvements of 14.8$\times$, 5.5$\times$, and 7.5$\times$ in throughput, energy, and computational density, respectively, than the state-of-the-art DaDianNao architecture. 

In CNN models, fine-grained parallelism appears at feature map level, in the neuron level,  and in  the synapse level. Lu et al.~\cite{lu2017flexflow} reviewed current accelerators that exploit the intrinsic parallelism and observed a mismatch between the parallel types supported by the computing engine and the dominant parallel types that appear in CNN workloads. They identified that most of the previous techniques proposed solutions that fall into one of the three representative architectures: (i) Systolic, (ii) 2D-mapping, and (iii) Tiling. 

Due to limitations of dataflow of each of the above three architectures, most existing accelerators support only  one specific parallelism. Systolic architectures exploit synapse parallelism (SP), 2D-mapping architectures exploit neuron parallelism (NP), and tiling architectures exploit  feature map parallelism (FP). However, in a practical CNN, the dominant parallel type depends on the number of input FMs, the number of output FMs, the size of the output FMs, and the size of the kernel. 

With three components (feature map, neuron, synapse) that can be either left serial, or parallelized, we get $2^3$ possible combinations. An example of processing style could be \textit{SFSNMS}, meaning, single feature map, single neuron, and multiple synapse. 

To address the above problem, and support all possible processing styles, Lu et al. \cite{lu2017flexflow} proposed a flexible dataflow architecture, called FlexFlow, with minimal controls. FlexFlow supports all types of data paths in each type of parallelism in different layers efficiently. 

As a first step, a modification to the processing element (PE) micro-architecture, and the interconnections between PEs, is proposed. PEs are arranged in rows where each row can complete one convolution and serve one output neuron. The adders in each PE row are connected to form the adder tree. Fig.~\ref{FlexFlow_PE_Arch} illustrates the proposed PE in FlexFlow and that in 2D-mapping architecture. By eliminating dependency between adjacent PEs, the proposed convolutional unit supports the comprehensive \textit{MFMNMS} parallelisms. To cater to different types of parallelisms, they also proposed a hierarchical dataflow with high data ``routability'' and low control overhead. The entire dataflow can be divided into three sub-flows: (i) distribution to local storage in each PE, (ii) fetching of data from local storage for operators (multiplier and adder), and, (iii) dataflow from neuron and kernel buffers to the distribution layer. 
They also presented a method to determine parallelization type and degree (i.e., the unrolling parameters) for each CONV layer.

FlexFlow architecture was evaluated for computing resource utilization, performance, power, energy, and area. Comparison was made with three typical architectures (i.e., systolic, 2D-mapping, and tiling) using six practical workloads, including AlexNet and VGG. They also examined the scalability of FlexFlow in terms of resource utilization, power, and area with growing scales of computing engine. 

From experimental results, it was found that computing resource utilization of each baseline was over 80\% across all workloads in contrast to other baselines that utilized less than 60\% (most of them less than 40\%).  In terms of performance, FlexFlow demonstrated over 420 GOPS performance with 1 GHz working frequency. It also outperformed others in terms of data reusability and power efficiency.  

\subsection{FPGA-based Accelerators} \label{FPGAs-based Accelerators}

In this subsection, we will review  recent  techniques employing FPGAs for the acceleration of deep learning networks.
For each reviewed technique, we will highlight the key features utilized to maximize performance and throughput
in the acceleration process.

FPGA implementations of CNNs appeared in the mid-1990's when Cloutier et al.~\cite{cloutier1996vip} designed the virtual image processor (VIP) on Altera EPF81500 FPGA. VIP is a single-instruction stream multiple-data streams (SIMD) multiprocessor architecture with a 2D torus connection topology of processing elements (PEs). VIP improves the performance through the use of low-accuracy arithmetic to avoid implementing full-fledged multipliers. Fortunately, recent digital signal processing (DSP)-oriented FPGAs include large numbers of multiply-and-accumulate (MAC) units which allow for extremely fast and low power CNN implementations.

Thereafter, FPGA implementations of deep learning networks have mainly focused on accelerating the computational engine through optimizing CONV layer operations. Several studies in the literature~\cite{wolf2001using, nichols2002feasibility, benkrid2002design, cardells2005area, girones2005fpga, zhang2007multiwindow, savich2007impact} have reported FPGA-based implementations of convolution operation.

Farabet et al.~\cite{farabet2009cnp} presented an FPGA implementation of CNN that uses one dedicated hardware convolver and a soft-processor for data processing and controlling, respectively. The proposed implementation is referred to as convolutional network processor (CNP). CNP exploits the parallelism of CONV layers to accelerate the computational engine of CNNs while fully utilizing the large number of DSPs, the MAC hardware units on FPGA. The proposed architecture consists of Virtex4 SX35 FPGA platform and external memory. The authors designed a dedicated hardware interface with the external memory to allow $8$ simultaneous read/write accesses transparently. In addition, they used first in first out (FIFO) buffers between the FPGA and the external memory chip in both directions to guarantee the steadiness of dataflow. 

The vector arithmetic and logic unit in CNP implements 2D CONV, pooling, and non-linear activation function operations of convolutional networks. The implementation of 2D CONV with kernel of size $3$ (i.e., $K = 3$) is shown in Fig.~\ref{CNP-Convolver}, where $x$ is the data from input feature map (FM), $y$ is the partial result to be combined with the current result, $z$ is the result to the output FM, $W_{ij}$ is the weight value in the convolution kernel, and $W$ is the width of the input image. It can be seen that the proposed convolutional module accomplishes $K^2$ MAC operations simultaneously in each clock cycle. CNP represents FMs and weights using 16-bit (Q8.8) fixed-point format. The proposed accelerator has been implemented for a face detection system with LeNet-5 architecture~\cite{lecun2015lenet}. It utilized 90\% and 28\% of the general logic and multipliers, respectively. In addition, CNP consumed less than 15 Watts of power.

\Figure[t](topskip=0pt, botskip=0pt, midskip=0pt)[width=0.45\textwidth, height = 6.4 cm]{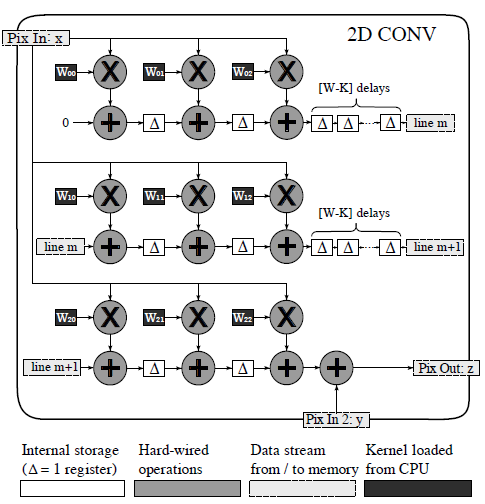}
{2D Convolution Module of $3 \times 3$ Kernel~\cite{farabet2009cnp}.
\label{CNP-Convolver}}

Sankaradas et al.~\cite{sankaradas2009massively} proposed a massively parallel coprocessor to accelerate CNNs using Virtex5 LX330T FPGA platform. The proposed accelerator mainly focused on optimizing computation engine by employing the parallelism within convolution kernel and FMs. The coprocessor can be considered as parallel clusters of vector processing elements (VPEs) where each cluster is designed using 2D convolvers, adders, sub-samplers, and look-up tables. Each VPE consists of multiplier-accumulator and programmable register units to hold kernel weights and FM data. To hold the massive intermediate data of CNNs, the authors employed a dedicated off-chip memory (4 DDR2 memory banks) with a large bandwidth on the coprocessor card. Moreover, the proposed accelerator uses a low precision data representation feature with memory packing to further improve the memory bandwidth as well as the throughput. 20-bit and 16-bit fixed-point representations were utilized for kernel weights and FMs, respectively.

The authors examined their architecture on CNN with 4 CONV layers and without any fully connected (FC) layer for a face recognition application. The results show that the proposed coprocessor is $6\times$ faster than a software implementation on a 2.2 GHz AMD Opteron processor with less than 11 Watts of power dissipation. However, the proposed accelerator cannot be used to accelerate full CNNs as it uses few CONV layers without any FC layer. A full CNN model consists of both CONV layers and FC layers. Thus, an efficient CNN accelerator for real-life applications is needed to consider both. Similar approaches to the work of Sankardas et al.~\cite{sankaradas2009massively} are presented in~\cite{graf2009massively, cadambi2009massively} to accelerate support vector machines (SVM).

\Figure[t](topskip=0pt, botskip=0pt, midskip=0pt)[width=0.48\textwidth, height = 5.2 cm]{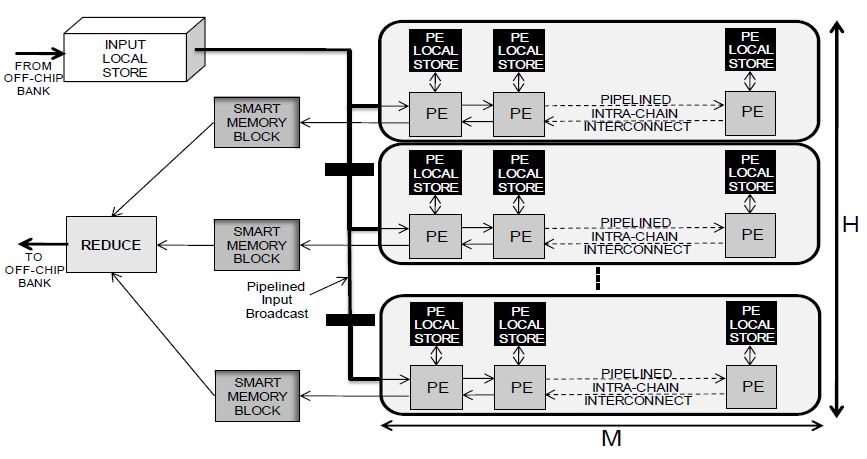}
{MAPLE Processing Core Architecture~\cite{cadambi2010programmable}.
\label{fig:cadambi2010programmable_first}}

MAPLE~\cite{cadambi2010programmable} is a programmable FPGA prototype system presented to accelerate both learning and classification tasks in applications with unstructured large amount of data. The authors analyzed five workload domains to help in designing MAPLE. These workloads are SVM~\cite{platt199912}, supervised semantic indexing (SSI)~\cite{bai2010learning}, K-means~\cite{macqueen1967some}, generalized learning vector quantization (GLVQ)~\cite{sato1996generalized}, and CNNs~\cite{lecun1998gradient}. They found that their computations can be structured as parallel streams of vector or matrix operations. Thus, they architected MAPLE as a 2D grid of vector processing elements as shown in Fig.~\ref{fig:cadambi2010programmable_first}. To efficiently perform matrix multiplication, they allocate a 
private local storage to each PE which is used to store a column, or part of it, from the multiplier matrix. In this  way, matrix multiplication is accomplished by streaming the multiplicand matrix rows through the PEs where each PE performs a MAC operation. The PEs are organized in clusters, where each group is  served by a separate memory bank of the banked off-chip memories, 
which create independent streams for processor-memory computation. 

Moreover, MAPLE uses on-chip smart memory blocks to process the large intermediate data on-the-fly using in-memory processing. Fig.~\ref{fig:cadambi2010programmable_second} shows the architecture of the smart memory block. To illustrate the idea of on-the-fly in-memory processing, lets consider finding the maximum $K$ elements. The filter compares the input data with the threshold value (VAL). If the input value is greater than VAL, it updates the list by replacing VAL at address ADDR with the input value. Then, the scanner (SCAN) searches for the new minimum value in the list and updates the threshold VAL and ADDR accordingly. 
It should be mentioned here that the employment of  in-memory processing reduced the off-chip memory traffic by $1.64\times$, $25.7\times$, and $76\times$ for SSI, K-means, and CNN workloads, respectively. MAPLE prototype has been implemented on Virtex5 SX240T platform running at 125MHz. The experimental results for face and digit recognition CNNs~\cite{lawrence1997face, chellapilla2006high, nasse2009face} show that MAPLE is 50\% faster than that for 1.3 GHz NVIDIA Tesla C870 GPU implementation.

\Figure[t](topskip=0pt, botskip=0pt, midskip=0pt)[width=0.48\textwidth, height = 5.2 cm]{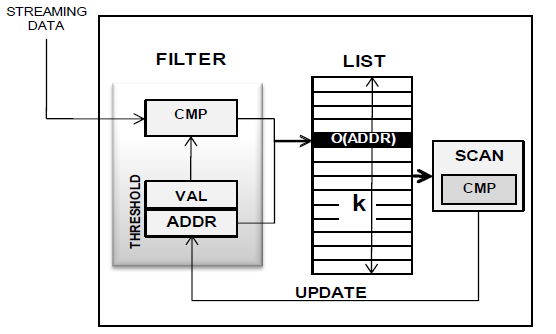}
{MAPLE Smart Memory Block~\cite{cadambi2010programmable}.
\label{fig:cadambi2010programmable_second}}

\Figure[b](topskip=0pt, botskip=0pt, midskip=0pt)[width=0.48\textwidth, height = 5.2 cm]{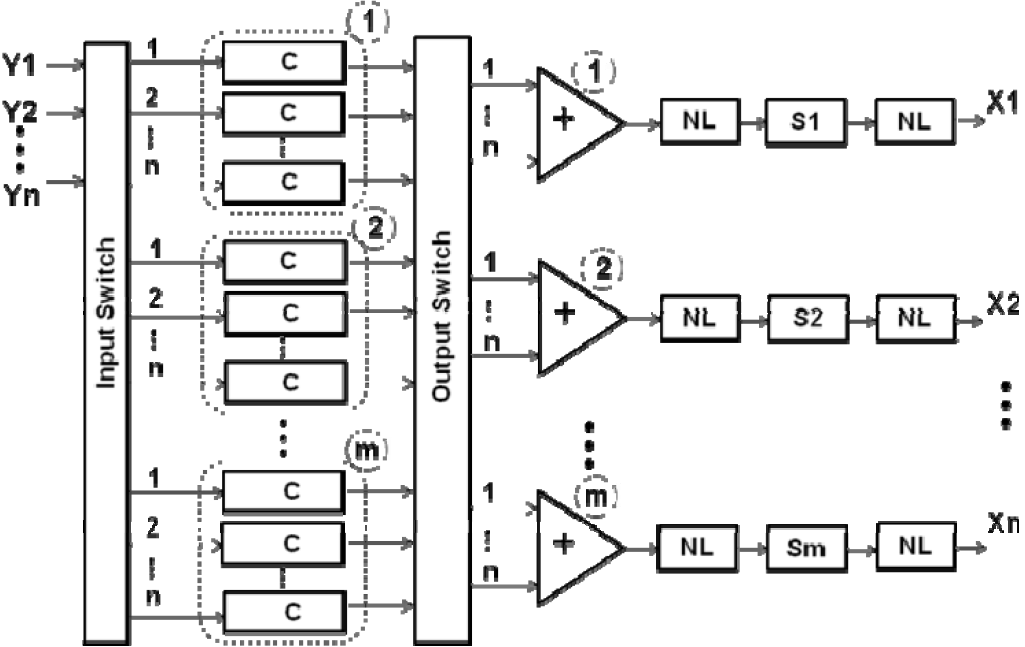}
{The Architecture of DC-CNN~\cite{chakradhar2010dynamically}.
\label{DC-CNN}}

Chakradhar et al.~\cite{chakradhar2010dynamically} proposed a dynamically configurable CNN architecture on FPGA. The proposed system consists of three main components; a coprocessor, a dynamically configurable CNN (DC-CNN) processing core, and 3-bank memory subsystem. The coprocessor is designed such that it  automatically configures the software and the hardware elements to fully exploit the parallelism at the workload level. DC-CNN is responsible for executing  CNN applications and its architecture is shown in Fig.~\ref{DC-CNN}. It consists of \textit{m} computational elements (each with \textit{n} 2D convolvers as well as sub-sampling (S) and non-linearity (NL) pipelined units), \textit{m} adders (each with \textit{n} inputs), and input/output switches. The internal structure of the switches vector encloses $m \times n$ selectors which are  used to help in exploring the entire design space and to provide the configurability function across different layers of CNN model. To determine the best (\textit{m}, \textit{n}) feasible combination for each layer, the system analyzes the workload using integer factorization techniques because it is considered fast for small numbers~\cite{dixon1981asymptotically, montgomery1994survey}. Dynamic programming is also used to quickly prune infeasible combinations.

The authors compared the proposed DC-CNN architecture, considering 20 2D convolvers as well as a memory subsystem of 128-bit port width, with a 1.35 GHz NVIDIA’s GPU implementation. The results show that DC-CNN achieved $4.0\times$, $4.4\times$, $5.4\times$, $6.0\times$, and $6.5\times$ speedup for face recognition~\cite{lawrence1997face}, face detection~\cite{nasse2009face}, mobile robot vision~\cite{farabet2009cnp}, video surveillance~\cite{chakradhar2010dynamically}, and automotive safety~\cite{chakradhar2010dynamically} workloads, respectively. It is worth mentioning  that DC-CNN is the first architecture that achieves a performance suitable for real-time processing for video streaming as it processes up to 30 frames per second. In addition, DC-CNN is more energy-efficient than the GPU implementation as it consumes 14 Watts, while more than 150 Watts are consumed by the GPU. On the other hand, the authors modeled their architecture on a CNN with 3 CONV layers only without any FC layer which makes it unsuitable for today's other real-life applications.

A second-generation of CNP~\cite{farabet2009cnp} architecture has been proposed in~\cite{farabet2010hardware} by designing a stream processor system. The proposed design replaces the  dedicated hardware convolver in CNP with multiple parallel vector processing units, named as ALUs, laid out in a 2D grid. Each ALU is composed of four local routers, one global router, and a streaming operator. The local routers are used to stream data to/from the neighbors. Streaming data to and from global data line is done through the global router. The streaming operators in the ALU are fully pipelined to produce a result per clock cycle as described in~\cite{farabet2009cnp} with the use of Q8.8 coding to represent FMs and weights. The proposed system also uses a multi-port direct memory access (DMA) streaming engine to allow individual streams of data to operate seamlessly within processing blocks. The results show that the proposed stream processor system can run small CNNs at up to 30 fps while consuming about 15 Watts.

An improved version of CNP architectures given in \cite{farabet2009cnp, farabet2010hardware} was presented in \cite{farabet2011neuflow} and referred to as neuFlow. Particularly, neuFlow has replaced the 2D grid of ALUs with a 2D grid of processing tiles (PTs). The proposed architecture contains a 2D grid of PTs, a control unit, and a smart DMA module, as shown in Fig.~\ref{CNP3_Arch}. Each PT consists of local operators and a routing multiplexer (MUX). The top three PTs have been implemented to perform MAC operation. Thus, they can be used to perform 2D convolution, simple dot-products, and spatial pooling. General-purpose operations, such as dividing and squaring, have been implemented at the middle three PTs. Therefore, the middle row of neuFlow can be used for normalization. Finally, neuFlow’s bottom PTs row implements non-linear operations.  Moreover, each operator employed  input and output FIFOs to stall its pipeline when required. On the other hand, PT’s MUX is used to connect its local operators with the neighboring PT’s streaming operators and off-chip memory instead of the used local routers and global router discussed in~\cite{farabet2010hardware}.

NeuFlow uses a dataflow compiler, named  luaFlow, to translate a high-level flow-graph representation of CNNs in Torch5~\cite{collobert2008torch} into HDL scripts with different levels of parallelism. In addition, luaFlow produces a binary code configuration file and holds it in the embedded control unit. Thereafter, the control unit configures the 2D grid of PTs (connections and streaming operator) and the DMA ports through run-time configuration buses. A smart memory module has been designed to support multiple asynchronous accesses of off-chip memory through its reconfigurable ports. By targeting the larger Xilinx Virtex6 VLX240T FPGA, neuFlow achieved 147 GOPS at 10 Watts for street scene parsing CNN in \cite{grangier2009deep} with the use of 16 bits to represent FMs and weights.

\Figure[t](topskip=0pt, botskip=0pt, midskip=0pt)[width=0.48\textwidth]{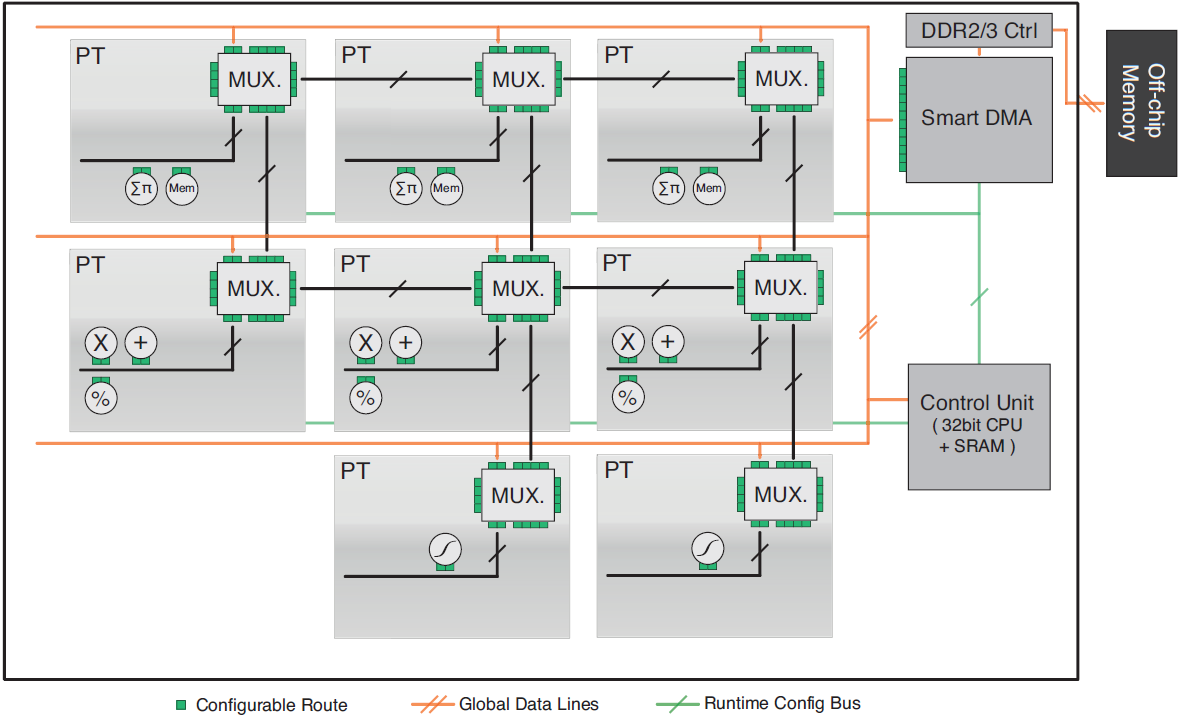}
{The Architecture of neuFlow~\cite{farabet2011neuflow}.
\label{CNP3_Arch}}

Peemen et al.~\cite{peemen2013memory} utilized the 
flexible off-chip memory hierarchy  method to design a configurable memory-centric accelerator template for a variety of models of CNNs. This  accelerator 
exploits data reuse in complex access patterns to reduce off-chip memory communication, which minimizes the bandwidth requirements. The memory-centric accelerator 
maximizes the efficiency of on-chip memories for better data locality using loop transformation (to optimize the tiling parameters) and block RAM (BRAM)-based multi-bank on-chip buffers~\cite{beric2008memory}. At the same time, it minimizes the size of FPGA on-chip memories to optimize energy and area usage, which are key requirements for embedded platforms.  

The memory-centric accelerator uses a SIMD cluster of MAC PEs with flexible reuse buffers to accelerate the CONV layer. The acceleration template has been implemented on Virtex6 FPGAs. In addition, a MicroBlaze processor has been utilized to configure and communicate with the accelerator via FIFO-based fast simplex link (FSL). The proposed accelerator has been analyzed for a CNN vision task of size 2.74 GMAC and the results show that the memory-centric accelerator is 11$\times$ faster than the standard implementation of similar FPGA resources.

Neural network next (nn-X)~\cite{gokhale2014240} is a real-time system-on-chip (SoC) computing system for deep learning networks on mobile devices. The architecture of nn-X consists of a host processor, a co-processor, and external memory. The co-processor accelerates the learning networks by 
parallelizing their operations throughout arrays of configurable processing elements referred to as {\it collections}. Each collection contains one convolution engine, one pooling module, and one non-linear
operator. The CONV engine 
accelerates the CONV operation by fully pipelining the incoming data with the use of cache memories. The collections are able to communicate with one  another using the collection 
route component to achieve cascaded pipelining, which results in reducing  accesses to  external memory. The data transfer between the collections and the external memory is accomplished throughout the co-processor full-duplex memory router, which provides independent data streams. The nn-X has been prototyped on Xilinx ZC706 which contains Zynq XC7Z045, two ARM Cortex-A9 processors, and 1 GB DDR3. Eight collections have been employed to achieve large parallelism. The results for face recognition model in~\cite{farabet2009fpga} show that nn-X is 115$\times$ faster than the two embedded ARM processors.

Zhang et al.~\cite{zhang2015optimizing} proposed a roofline-based model  to accelerate convolutional neural networks on FPGAs. 
The roofline model is an intuitive visual performance model used to relate the attainable performance to the peak performance that can be provided by the hardware platform and the off-chip memory traffic~\cite{williams2009roofline}.
The focus in their  work is primarily on accelerating the convolutional layers as it consumes more than 90\% of the computational time during the prediction process~\cite{cong2014minimizing}. In doing so, the authors optimized both the computation operations and the memory access operations in convolutional layers. They considered a CNN application composed of five convolutional layers that won the ImageNet competition in 2012~\cite{krizhevsky2012imagenet}. The proposed accelerator uses polyhedral-based data dependence analysis~\cite{pouchet2013polyhedral} to fully utilize all FPGA computational resources through loop unrolling, loop pipelining, and loop tile size enumeration. Note that loop unrolling maximizes the parallel computation of CONV MAC operations. On the other hand, local memory promotion and loop transformation are used to reduce redundant communication operations and to maximize the data sharing/reuse, respectively.

Subsequently, the roofline performance model is used to identify the optimal design from all possible solutions in the design space. Specifically, the authors model all possible legal designs delivered from the polyhedral analysis in roofline to find the optimal unrolling factor $\langle T_{m}, T_{n} \rangle$ for every convolutional layer, where $T_{m}$ and $T_{n}$ are the tile size for the output FMs and input FMs, respectively. However, designing a CNN accelerator with different unrolling factors to each convolutional layer is challenging. 
Therefore, the proposed architecture enumerates  all possible valid designs to find uniform cross-layer unrolling factors. Thereafter, the hardware accelerator is implemented based on the cross-layer optimal unrolling factors. 

\Figure[t](topskip=0pt, botskip=0pt, midskip=0pt)[width=0.48\textwidth]{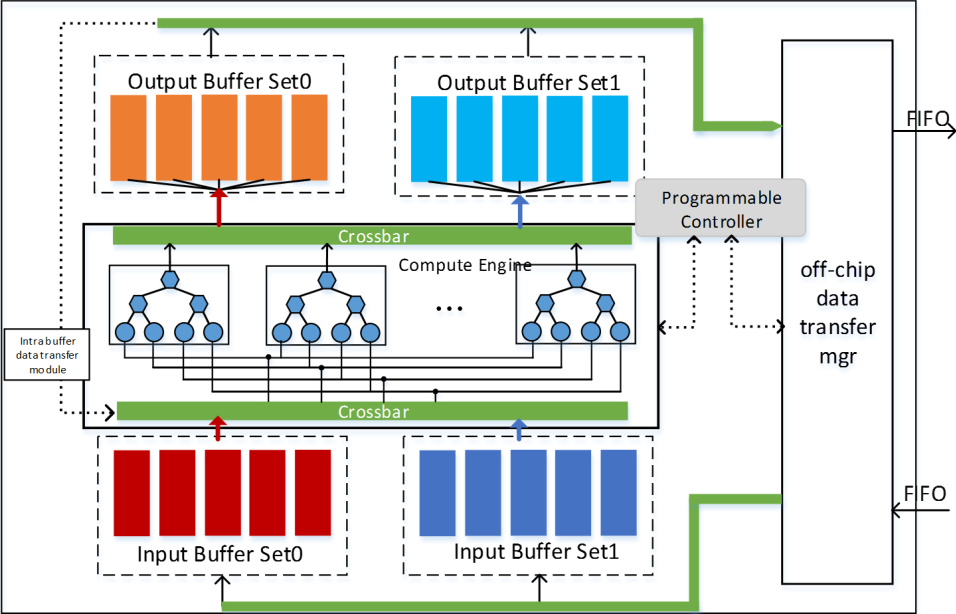}
{Zhang et al.~\cite{zhang2015optimizing} Accelerator Architecture.
\label{Optimizing_Accelerator_Arch}}

The proposed accelerator composed of a computational engine and memory sub-system  is depicted in Fig.~\ref{Optimizing_Accelerator_Arch}. The computation engine is designed as $T_{m}$ duplicated tree-shaped poly structures with $T_{n}$ inputs from the input FMs, $T_{n}$ inputs from the weights, and one input from the bias. On the other hand, the memory sub-system is implemented as four sets of on-chip buffers; two sets to store the input FMs and weights, each with $T_{n}$ buffer banks, and two buffer sets of $T_{m}$ independent banks for storing the output FMs. To overlap data transfer with computation, on-chip buffers are operated in a ping-pong manner. In addition, two independent channels are implemented for load and off-load operations to increase the bandwidth utilization. Moreover, MicroBlaze processor is used to send configuration parameters and commands for the accelerator over AXI4lite bus. The CNN accelerator communicates with external data transfer engines through FIFO interfaces, where the data transfer engines are used to access DDR3 DRAM memory through AXI4 bus.

The accelerator is designed using Vivado 2013.4 high level synthesis tool and implemented on Xilinx VC707 FPGA board clocked at 100 MHz. The experimental results depict that the proposed implementation achieves a peak performance of 61.62 GFLOPS as well as a 17.42$\times$ speedup over the software implementation on Intel Xeon CPU E5-2430 at 2.20 GHz with 15 MB cache and 16 threads. In addition to this, the results show that the proposed FPGA architecture is 24.6$\times$ more energy-efficient than the software implementation as the total power consumption is only 18.6 Watts. The proposed implementation has some limitations such as designing the accelerator with new cross-layer unrolling factors for different architectures of CNNs. Furthermore, using the CNN accelerator with uniform unrolling factors might be sub-optimal for some CONV layers, which affects the overall performance. 

In 2014, Microsoft research team of Catapult project integrated FPGA boards into data center applications to successfully achieve $2 \times$ speedup for Bing Ranking (the large-scale search engine)~\cite{putnam2014reconfigurable}. A year later, Ovtcharov et al.~\cite{ovtcharov2015accelerating} at Microsoft Research utilized Catapult hardware infrastructure, a dual-socket Xeon server equipped with Stratix-V GSMD5 FPGA, to design a specialized hardware for accelerating the forward propagation of deep CNNs in a power-constrained data center.

The top-level architecture of the proposed CNN accelerator is shown in~Fig.~\ref{Microsoft_CNN_Accelerator}. Multi-banked input buffer and kernel weight buffer are used to provide an efficient buffering scheme of FMs and weights, respectively. To minimize the off-chip memory traffic, a specialized network on-chip has been designed to re-distribute the output FMs on the multi-banked input buffer instead of transferring them to the external memory. The 3D convolution operations (such as the dot-product) and other CNN operations are independently performed using spatially distributed scalable vectors of PEs. The controller engine is responsible for streaming and data delivery of multi-banked input buffer and kernel weight buffer data to each of the PE vectors. In addition, it supports configuring multiple CNN layers at run-time. The results show that the proposed design is able to classify 134 images/sec, while consuming about 25 Watts, for AlexNet model on ImageNet-1K dataset~\cite{krizhevsky2012imagenet}, which is $3 \times$ better than the published throughput results for the Roofline-based FPGA Accelerator~\cite{zhang2015optimizing}. The authors mentioned that using top-of-the-line FPGAs such as Arria 10 GX 1150 improves the throughput to   around 233 images/sec.

\Figure[t](topskip=0pt, botskip=0pt, midskip=0pt)[width=0.43\textwidth]{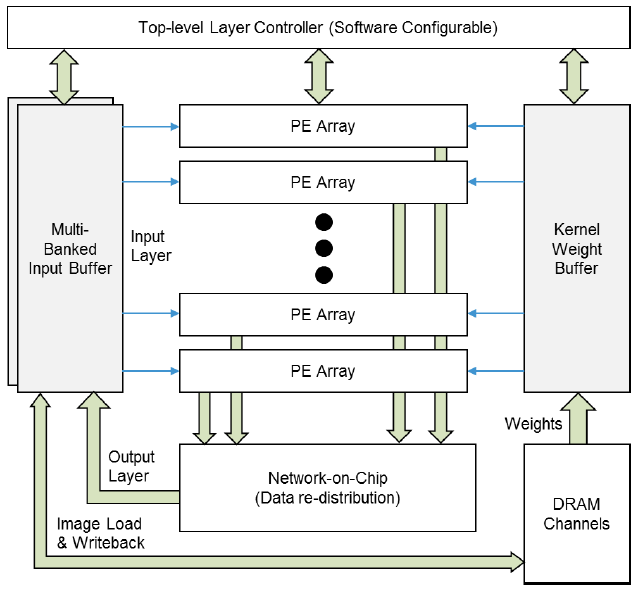}
{Top-Level Archeticture of Microsoft CNN Accelerator~\cite{ovtcharov2015accelerating}.
\label{Microsoft_CNN_Accelerator}}

Qiu et al.~\cite{qiu2016going} proposed  an FPGA design to accelerate CNNs for a large-scale image classification challenge on embedded systems. The  focus was  on accelerating both CONV and FC layers, since they are considered as the most computational-centric and the most memory-centric operations in CNNs, respectively. The proposed accelerator reduces the resource consumption using specific design of convolver hardware module. In addition, the authors applied singular value decomposition (SVD) to the weight matrix of FC layer in order to reduce memory footprint at this layer~\cite{van1996matrix}. To further reduce memory footprint and bandwidth requirement of CNN, they proposed a dynamic-precision data quantization flow component. This component is responsible for finding the optimal fractional length for weights in each layer as well as the optimal fractional length for FMs in adjacent layers, while achieving   minimal accuracy loss. Then, it converts the floating-point numbers representing weights and FMs into fixed-point numbers. 

\Figure[b](topskip=0pt, botskip=0pt, midskip=0pt)[width=0.48\textwidth]{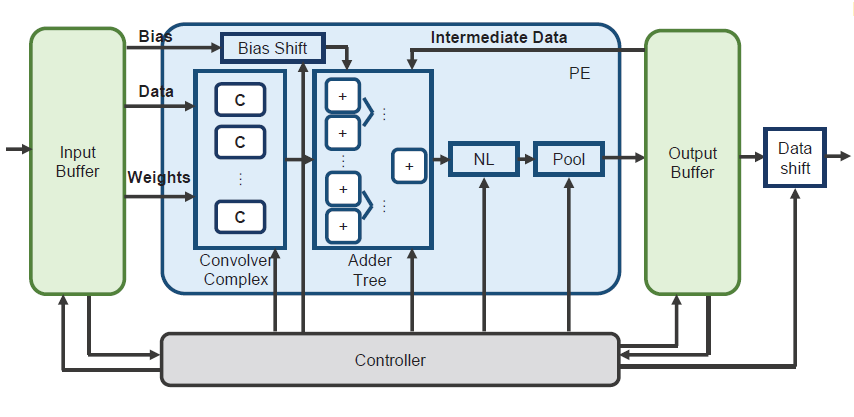}
{Processing Element Module of Qiu et al.~\cite{qiu2016going} Embedded Accelerator Architecture.
\label{fig:qiu2016going_first}}

In addition, the authors proposed a data arrangement scheme that maximizes the burst length of each transaction to the external memory to accelerate CONV and FC layers, as well as to avoid unnecessary access latency. Note that maximizing the DRAM burst length raises up the effective DRAM bandwidth~\cite{ zhang2015optimizing, zhang2018caffeine}.

\Figure[t](topskip=0pt, botskip=0pt, midskip=0pt)[width=0.45\textwidth]{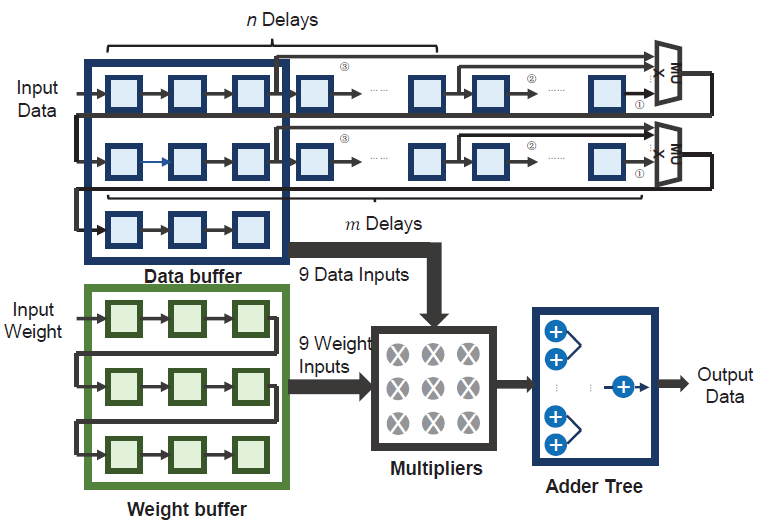}
{Convolver Complex Design of Qiu et al.~\cite{qiu2016going} Embedded Accelerator Architecture.
\label{fig:qiu2016going_second}}

The proposed architecture consists of a  processing system (CPU) and programmable logic (FPGA). CNN computations are performed through special design of processing element modules in FPGA. The main modules in the processing element are convolver complex, max-pooling, non-linearity, data shift, bias shift, and adder tree, 
as shown in Fig.~\ref{fig:qiu2016going_first}. The convolver complex is designed as a classical line buffer~\cite{bosi1999reconfigurable}, as shown in Fig.~\ref{fig:qiu2016going_second}, to achieve convolution operations as well as to compute FC layer multiplication of matrix-vector. The pooling layer implemented in the max-pooling module is used to output the maximum value in the input data stream with a window of size 2. The activation function of CNN is applied to the input data stream using the non-linearity module. 
The adder tree accumulates the partial sums generated from the convolvers. Finally, data shift and bias shift modules are responsible for accomplishing dynamic quantization. 

The proposed embedded FPGA platform has been implemented using VGG-16-SVD network with 16-bit fixed-point numbers on Zynq XC7Z045 platform. The results demonstrate that applying SVD technique reduces memory footprint of FC layer by 85.8\% with a compression rate of 7.04\% while introducing an accuracy loss of only 0.04\%. Finally, the overall performance of the proposed accelerator reported is 136.97 GOPS under 150 MHz working frequency with the top-5 accuracy of 86.66\% and a total power consumption of 9.63 Watts.

DeepBurning~\cite{wang2016deepburning} is an FPGA-based neural network (NN) design automation tool. It allows for building learning accelerators for specific NN with optimized performance and custom design parameters configuration using a pre-constructed register transfer level (RTL) module library. The RTL library holds the hardware descriptive scripts for NN reconfigurable components as well as their configuration scripts. In addition, it contains other RTL building blocks for logical and arithmetic operations such as the connection box (used to exchange data between NN layers as well as to approximate the division operation) and approximate look-up table (LUT) (used to simplify a function or operation to allow it to be mapped into hardware).

In order to design an optimized hardware, DeepBurning compresses the passed NN model to the greatest extent using temporal and spatial folding which helps also in satisfying the resource constraints and minimizing  the required hardware modules. DeepBurning not only generates the hardware description for neural network scripts, but also analyzes the complex access pattern and data locality using an integrated compiler to generate a run-time control flow which provides energy-efficient,  and,  better data reuse implementation. In addition, the DeepBurning compiler investigates the accelerator on-chip memory size and throughput to properly tile and partition the NN weights and feature data layouts. Moreover, DeepBurning uses the address flow component to automatically fetch and store off-chip memory and on-chip memory data. The authors compared the performance of DeepBurning with that in~\cite{zhang2015optimizing}, considering AlexNet CNN model, as they both operate at 100 MHz. They considered a high budget resources constrained DeepBurning on Zynq-7045 device. The results show that DeepBurning is 1.13$\times$ slower but 1.45$\times$ 
more energy-efficient.

An OpenCL-based optimization framework to accelerate large-scale convolutional neural network models was proposed by Suda et al.~\cite{suda2016throughput}. They found that the number of performed CONV MAC operations in parallel ($N_{CONV}$), SIMD vectorization factor ($S_{CONV}$), normalization layer loop unrolling factor ($N_{NORM}$), the number of parallel pooling outputs in one cycle ($N_{POOL}$), and the number of parallel FC MAC operations ($N_{FC}$) are the key variables that determine the parallelism of the design. Subsequently, they analytically and empirically modeled the execution time for each layer as a function of the above  mentioned variables. Then, genetic algorithm was used to explore the design space for finding the optimal combination of the key design variables considering the resources constraints.

The authors implemented the scalable CONV block in a similar fashion to that in~\cite{chellapilla2006high} as a matrix multiplication by flattening and on-the-fly rearrangement of  the feature data. The OpenCL software has been utilized in their work due to its parallel programming model as well as its ability to integrate the compiled RTL design with external memory interfacing IPs~\cite{khronos2011opencl}, which uses memory coalescing technique with complex load and store units. In addition, it has optimized matrix multiplication and CPU-FPGA communication libraries~\cite{abdelfattah2014gzip, Altera2018OpenCL}.

The framework is used on both VGG-16 and AlexNet CNN models which are implemented on P395-D8~\cite{P395D8} and DE5-Net~\cite{DE5Net} FPGA boards with fixed-point operations according to their precision study. They compared the proposed implementation with 3.3 GHz core i5-4590 CPU implementation that uses Caffe tool~\cite{jia2014caffe} with ATLAS~\cite{whaley1998automatically} optimized library for matrix/vector operations. The results show that the OpenCL optimized framework on P395-D8 achieved $5.5\times$ (117.8 GOPS) and $9.5\times$ (72.4 GOPS) speedups for VGG-16 and AlexNet models, respectively. On the other hand, DE5-Net FPGA achieved less throughput speedup than the P395-D8 ($2.2\times$ (47.5 GOPS) for VGG-16, and $4.2\times$ (31.8 GOPS) for AlexNet) as it has $7.67\times$ less DSPs than what is  available on P395-D8.

Zhang et al.~\cite{zhang2016caffeine, zhang2018caffeine}  analyzed the transformation of CONV and FC layers to regular matrix multiplication presented in prior work~\cite{qiu2016going}. For VGG-16 model, they found that such transformation necessitates up to 25$\times$ duplication of input FMs. To address this problem and improve the bandwidth utilization, they   designed a uniformed matrix multiplication kernel that uses either  input-major mapping (IMM) or weight-major mapping (WMM) techniques while computing FC layer. In IMM, the designed kernel batches a group of different input FMs together, and then performs the matrix multiplication. IMM technique improves the data reuse of FC weights. On the other hand, the designed kernel with WMM technique makes use of the fact that the FC layer is communication-bound in which the weight matrix is much larger than the input FM matrix.  In particular, it loads input FM matrix to a weight buffer and loads weight matrix to input FM buffer. Subsequently,  a regular matrix multiplication is performed on these matrices. As a result, WMM may allow for a higher data reuse than IMM, especially for input FMs that can be reused multiple times considering the limited hardware resources.

For the above, the roofline model was applied to identify the optimal mapping technique under different batch sizes and data precisions. The results demonstrate that WMM is better than IMM in term of data reuse and bandwidth utilization, especially in small batch sizes which is required for real-time inference. Hence, the same matrix multiplication kernel is utilized for the computation of both CONV and FC layers, but with the use of IMM in CONV layer and WMM in FC layer. Based on this, the authors proposed a software/hardware co-design library, which they named  Caffeine, to accelerate CNNs on FPGAs.

With an easy-to-use developed tool, Caffeine aids in automatically choosing the best hardware parameters, using the model files from Caffe and FPGA device specifications obtained from the user. Caffeine FPGA engine uses a high-level synthesis (HLS)-based systolic-like architecture to implement matrix multiplication kernel. It allows changing parameters such as number of PEs, precision, and FM size. Caffeine further maximizes the FPGA computing capability by optimizing multi-level data parallelism discussed in~\cite{zhang2015optimizing} and pipeline parallelism using polyhedral-based optimization framework given in~\cite{zuo2013improving}. Caffeine framework also handles the weights and biases reorganization in off-chip DRAM to maximize the underlying memory bandwidth utilization. In addition, the double-buffering technique is employed to prefetch the next
data tile for each PE. Caffeine has been evaluated by implementing AlexNet and VGG-16 CNNs on Ultrascale KU060 (20nm and 200 MHz) and on Virtex7 690T (28nm and 150 MHz) considering different precisions. The VGG-16 implementation with 16-bit fixed-point on Ultrascale KU060 and Virtex7 690T provided 43.5$\times$ and 65$\times$ overall throughput enhancement, respectively, compared to implementation on a two-socket server, each with a 6-core Intel CPU (E5-2609 at 1.9 GHz).

A special case of dataflow, referred to as synchronous dataflow (SDF) \cite{lee1987synchronous}, is a paradigm of computation that allows for representing a computing system as a streaming problem. In this way, SDF model can represent the hardware implementation of CNNs using linear algebra and directed SDF graph (SDFG). Each node of SDFG represents a hardware building block that can immediately start its computation as soon as the data are available through its input arcs. Such representation of CNN model offers a fast design space exploration.  Venieris and Bouganis \cite{venieris2016fpgaconvnet} employed SDF model to optimize the mapping of CNNs onto FPGAs based on HLS.

In particular, the proposed fpgaConvNet framework in \cite{venieris2016fpgaconvnet} takes as input a high-level script programmed by DL expert  describing the CNN model, along with specifications of the targeted FPGA platform. Thereafter, it parses the input script through a developed domain-specific language (DSL) processor to model the CNN in the form of a directed acyclic graph (DAG) where each node corresponds to a CNN layer. Then, the DAG-based CNN is transformed into an SDFG representation and modeled as a topology matrix. The topology matrix contains the number of incoming parallel streams, the width of each data stream, and the production or consumption rates at each node. In addition, the DSL processor extracts information about the platform-specific resource constraints.

\Figure[t](topskip=0pt, botskip=0pt, midskip=0pt)[width=0.45\textwidth]{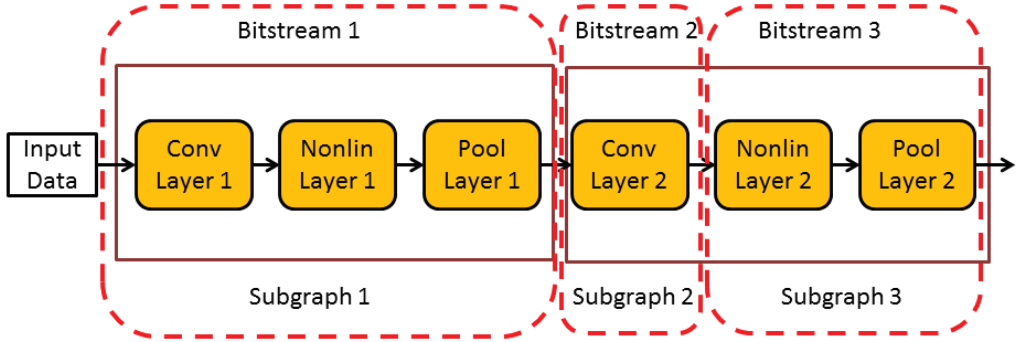}
{SDF Graph Partitioning~\cite{venieris2016fpgaconvnet}. 
\label{SDF_partitioning}}

Unlike other attempts, instead of exploring the design space for the optimal parameters of loop unrolling and tiling, fpgaConvNet explores the design space of the topology matrix components while considering the resource constraints. In doing so, fpgaConvNet performs graph partitioning, coarse-grained folding, and fine-grained folding. The graph partitioning splits the original SDFG into subgraphs and each subgraph is then mapped to a distinct bitstream as shown in Fig.~\ref{SDF_partitioning}. Note that the proposed multi-bitstream architecture might have multiple CONV layer processors (CLPs), as in the provided example. This away, on-chip RAM is used for intermediate results and data reuse within the subgraph, while accesss of  off-chip memory is minimized and limited for input and output streams of the subgraph. However, this scheme adds reconfiguration penalty due to the need for reconfiguring the FPGA when the data flows between adjacent subgraphs. To amortize this overhead, several input data streams are processed in a pipelined manner. 

Thereafter, each bitstream architecture is  optimized using coarse-grained folding and fine-grained folding. In coarse-grain folding, CONV, pooling, non-linear, and other major operations of each layer are unrolled to provide the highest possible throughput by having several parallel units of each operation. The fine-grain folding controls the unrolling and pipelining of the dot-product operations inside CONV and average pooling units. Instead of fully unrolling the implementation of dot-product which produces a 1 dot-product per cycle, with the use of a high number of multipliers and adders, fpgaConvNet uses a smaller number of MAC units and schedules the execution of different operations using time-multiplexing. A trade-off between the performance and the required hardware resources can be achieved by changing the unroll factor and the degree of multiplexing. Therefore, fpgaConvNet employed simulated annealing \cite{reeves1995modern} to find the optimal partitioning points and folding factors. Finally, fpgaConvNet uses optimal components to derive the configuration of PEs and buffers, and generates a synthesizable Vivado HLS hardware design.

fpgaConvNet framework has been evaluated by mapping LeNet-5 and scene labelling~\cite{cavigelli2015accelerating} small CNN models with Q8.8 fixed-point representation onto a Zynq-7000 XC7Z020 FPGA platform  working at 100 MHz. In mapping LeNet-5, fpgaConvNet achieves up to 1.62$\times$ the performance density of CNP~\cite{farabet2009cnp}. Compared to Tegra K1 GPU implementation of scene labelling CNN, fpgaConvNet surpasses Tegra K1’s power efficiency by 1.05$\times$.

Ma et al.~\cite{ma2016scalable} proposed a Python-based modularized RTL compiler to accelerate CNNs by employing loop unrolling optimization~\cite{bacon1994compiler, zhang2015optimizing} for CONV layer operations. A detailed review article of this work has been recently published and referred to as ALAMO~\cite{ma2018alamo}. The proposed compiler integrates both the RTL finer level optimization and the flexibility of HLS to generate  efficient Verilog parameterized RTL scripts for ASIC or FPGA platform under the available number of parallel computing resources (i.e., the number of multipliers ($N_{m}$)). If $N_{m}$ is greater than the number of input FMs ($N_{if}$), the proposed compiler fully unrolls \textit{Loop-3} ($N_{if}$, refer to subsection~\ref{Convolution (CONV)} for more details) while it partially unrolls \textit{Loop-4} ($N_{of}$) to exploit the data reuse of shared features among $N_{m}/N_{if}$ output FMs. Otherwise, it partially unrolls \textit{Loop-3} which results in $N_{if}/N_{m}$ repeated sliding of kernel window. On the other hand, \textit{Loop-2} ($X \times Y$) is serially computed after \textit{Loop-1} ($K$) to minimize the number of partial sums.

The overall modules of the proposed CNN accelerator are shown in Fig.~\ref{ALAMO-Accelerator}. The controller is responsible for directing and ensuring in-order computation of CNN modules for each layer. The data routers oversee the selection of data read and data write of two adjacent modules as well as the assignment of buffer outputs to shared or pool multipliers of the multiplier bank. The feature buffers hold the FMs using on-chip RAMs. The weight buffers are used to ensure the availability of CONV and FC layers' weights before their computation as well as to overlap the transfer of FC layer weights with its computation. The CONV module consists of control logic, groups of adder trees, and ReLU components. The control logic component parametrizes the loop unrolling factors based on the configuration of each layer ($N_{if}$, $N_{of}$, $X$, $Y$, and $K$). The CONV module contains $N_{m}/N_{if}$ adders to sum $N_{if}$ parallel multiplier results and accumulate them. Moreover, the adder trees can be shared by layers with identical $N_{if}$ to be as one single module. The ReLU component checks the input pixel sign bit to either output zero or the data pixel itself. The POOL module contains accumulators or comparators to perform average or maximum operation, respectively. The NORM module maintains the required components to perform the operations of local response normalization such as square, non-linear (using look-up table), and multiplication operations. Finally, the FC module shares the multiplier bank module with the CONV module to perform the matrix-vector multiplication (MVM).

\Figure[t](topskip=0pt, botskip=0pt, midskip=0pt)[width=0.45\textwidth, height = 5.2 cm]{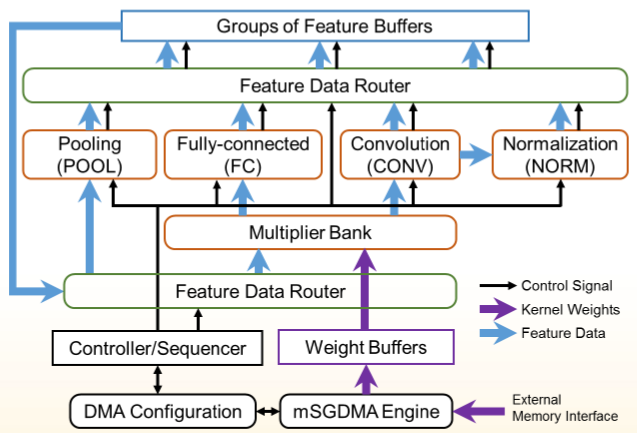}
{ALAMO Overall Acceleration Modules~\cite{ma2016scalable}.
\label{ALAMO-Accelerator}}

\Figure[t!](topskip=0pt, botskip=0pt, midskip=0pt)[width=0.84 \textwidth]{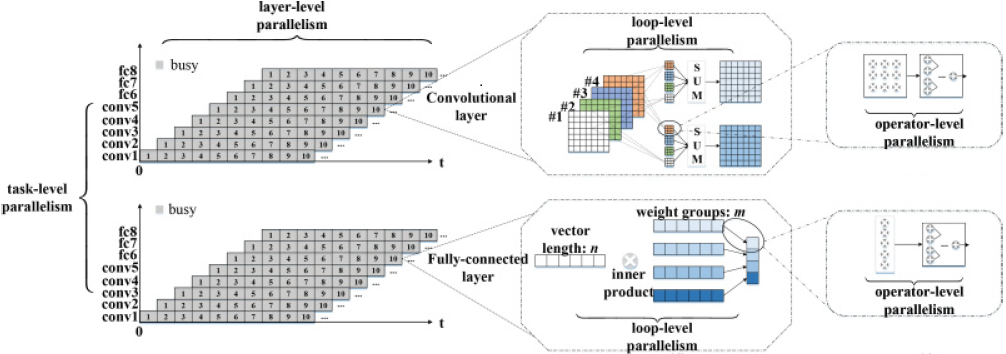}
{Parallel Framework Exploiting Four Levels of Parallelism~\cite{liu2017throughput}.
\label{ParallelFramework}}

ALAMO architecture permits the output pixels to be only stored in the feature buffers, which makes ALAMO suitable for CNNs with only small intermediate data volumes. The proposed RTL compiler has been tested by accelerating two CNN models; AlexNet and NiN~\cite{lin2013network}. The generated parameterized RTL scripts for AlexNet and NiN are synthesized using Altera Quartus synthesis tool and  implemented on DE5-Net FPGA board. The experimental results for AlexNet model are compared with the results for OpenCL-based design~\cite{suda2016throughput} as both use the same FPGA board with similar hardware resources for AlexNet. ALAMO  achieved $1.9 \times$ and $1.3 \times$ improvement for throughput and power consumption, respectively. Moreover, the overall throughput of NiN model is $1.03 \times$ better than that of AlexNet. This is because  NiN has more CONV layers and many of them have the same $N_{if}$.

Liu et al. \cite{liu2017throughput}   proposed a parallel framework for FPGA-based CNN accelerators that exploits
four levels of parallelism;   task level, layer level, loop level, and operator
level. Task-level parallelism involves executing multiple image 
prediction tasks simultaneously.
Layer-level parallelism exploits pipelining across layers to enable parallel execution
of all layers with different images. 
Loop-level parallelism  utilizes loop unrolling  in performing   convolutions and this can be achieved 
 either through intra-output or  inter-output parallelism. Finally, operator-level parallelism
is achieved by parallelising the $k \times k$ MACs operations needed for convolution operation in convolutional layers or the $ n$  MACs needed for inner-product computation in fully connected layers.
Fig.~\ref{ParallelFramework} shows the parallel framework exploiting these four levels of parallelism.

The authors  have  used 16-bit fixed-point format for representing  pixels in input feature maps and output feature maps.
However, they have used 32 bits for intermediate results which get truncated to 16 bits. In addition,
they have used 8 bits for representing kernels and weights.
They have presented  
a systematic
methodology for design space exploration to  find the optimal solution that maximizes
the throughput of an FPGA-based accelerator under given FPGA constraints such as on-chip
memory, computational resources, external memory bandwidth, and clock frequency.

The proposed technique has been evaluated by implementing  three CNN accelerators on the VC709 board for LeNet, AlexNet, and
VGG-S. It has achieved a throughput of 424.7 GOPS, 445.6 GOPS, and 473.4 GOPS for LeNet, AlexNet, and VGG-S accelerators, respectively.
In addition, the performance has been compared with MatConvNet
tool running the CNN models on Intel Core i7-4790K CPU (4.0 GHz) and NVIDIA GTX-770 GPU (1,536
CUDA cores, 2 GB GDDR5, 224.3 GB/s memory bandwidth). 
Compared to the CPU implementations, the accelerators for LeNet, AlexNet, and
VGG-S achieved $14.84\times$, $6.96\times$, and $4.79\times$ in performance, respectively, and $51.84\times$, $24.69\times$, and
$16.46\times$ in power efficiency, respectively.
Compared to the GPU implementations, the accelerators
achieved better performance in the small-scale network LeNet ($3.17\times$), comparable performance
in the medium-scale network AlexNet ($0.96\times$), and worse performance in the large-scale network VGG-S ($0.56\times$).
However, the accelerators achieved higher power efficiency than the
GPU implementations in all three networks with $28.3\times$ for LeNet, $8.7\times$ for AlexNet and $4.98\times$ for VGG-S.

FP-DNN \cite{guan2017fp} is an end-to-end framework that automatically generates optimized FPGA-based implementations of deep neural networks (DNNs) using an RTL-HLS hybrid library. FP-DNN compiler, programed using C++ and OpenCL, takes TensorFlow symbolic descriptions \cite{abadi2016tensorflow} of DNNs, and then performs model inference through the use of model mapper, software generator, and hardware generator modules. The model mapper extracts the topological structure and layers configurations of DNN model from the TensorFlow descriptions and generates an execution graph for the target model. The execution graph shows  layer-by-layer operations and read/write data transactions.

FP-DNN compiler allocates off-chip DRAM data buffers to store intermediate data, weights, and model parameters and configurations. The model mapper maximizes the storage resource reuse through minimizing the number of required physical buffers. Specifically, it formulates the data reuse problem as a graph coloring problem \cite{alsuwaiyel2016algorithms}, and then the left-edge algorithm is applied to generate kernel configuration and kernel schedule. Subsequently, the software generator uses the kernel schedule to generate a host C++ program which initializes the model, manages the data buffers, and schedules the kernel execution. On the other hand, the hardware generator uses the kernel configuration and the execution graph to generate the FPGA hardware codes by instantiating the corresponding optimized templates from an expandable RTL-HLS hybrid library. Each template is comprised of Verilog-based computational engine and OpenCL-based control logics engine.

The architecture of the proposed FPGA-based accelerator consists of matrix multiplication and data arranger modules. Matrix multiplication module is a hand-written Verilog code that is designed and optimized based on the hardware constraints of Altera Stratix-V GSMD5 FPGA. It applies tiling and ping-pong double buffers techniques to improve the throughput. On the other hand, data arranger is an OpenCL-based module that is responsible for mapping the computational part of a layer to matrix multiplication as well as performing data communication with off-chip memory and matrix multiplication module. Mapping DNNs computational operations to matrix multiplication has been widely applied in prior studies \cite{cadambi2010programmable, chetlur2014cudnn, suda2016throughput}. FP-DNN maps FC layer to matrix multiplication by batching input vectors together. Before model deployment, FMs and weights are rearranged in DRAM using the channel-major scheme to optimize the communication between the accelerator and off-chip DRAM. On the other hand, both floating-point and fixed-point representations have been supported for implementation, and they can be adjusted by the user.

The proposed RTL-HLS hybrid framework has been evaluated by accelerating VGG-19, LSTM-LM \cite{zaremba2014recurrent}, ResNet-152 DNNs on Stratix-V GSMD5 FPGA. Note that this is the first work that implements ResNet-152 on FPGA. The experimental results demonstrated that the speedup of FP-DNN for 16-bit fixed-point implementations are about 1.9$\times$ - 3.06$\times$ compared with the server that includes 2 processors each with 8-core Intel Xeon E5-2650v2 at 2.6 GHz.

In line with  the current trends towards compressed neural networks, with dramatically reduced weights and activations bit-width using 1-bit or 2-bit quantization~\cite{sung2015resiliency, rastegari2016xnor, kim2016bitwise, zhou2016dorefa, courbariaux2016binarynet}, Umuroglu et al.~\cite{umuroglu2017finn} conducted  a set of experiments to estimate the trade-off between the network size and precision using the roofline model. They found that binarized neural networks (BNNs)~\cite{courbariaux2016binarynet} require 2 to 11 times more operations and parameters than an 8-bit fixed-point CNN to achieve a comparable accuracy on MNIST~\cite{lecun1998gradient} dataset. However, the performance of BNN is found to be 16$\times$ faster than the fixed-point network.

Subsequently, the authors proposed a framework, referred to as FINN~\cite{umuroglu2017finn}, that maps a trained BNN onto FPGA. FINN generates a synthesizable C++ network description of a flexible heterogeneous streaming architecture. The architecture consists of pipelined compute engines that communicate via on-chip data streams. Each BNN layer has been implemented using dedicated compute engines with 1-bit values for weights and FMs; +1 and -1 are used to represent a \textit{set} bit and \textit{unset} bit, respectively.

The authors have optimized accumulation, batch normalization (batchnorm), activation, and pooling operations of BNNs. In particular, the accumulation of a binary dot-product has been implemented as a counter of \textit{set} bits (\textit{popcount} operation). The popcount-accumulate reduces the number of required look-up tables (LUTs) and flip-flops (FFs) by a half, compared to the implementation of signed-accumulation. BNN batchnorm and activation operations have been simplified and implemented together as unsigned comparison with a threshold ${\tau}_{k}$, +1 is produced when the input value is greater than or equals to ${\tau}_{k}$, and -1 otherwise. The value of ${\tau}_{k}$ is computed during  run-time. Such an implementation of batchnorm-activation operations requires much smaller number of LUTs, without the need for DSPs and FFs, compared to regular implementation of batchnorm-activation. Max-pooling, average-polling, and min-pooling have been effectively implemented with Boolean OR-operator, Boolean majority function, and Boolean AND-operator, respectively.

 The accelerator architecture is composed of building blocks from the FINN hardware library. The matrix-vector-threshold unit (MVTU) is the core computational building block as matrix-vector operations followed by thresholding form the majority of BNN operations. The design of MVTU consists of an input buffer, an array of $P$ parallel PEs each with $S$ SIMD lanes, and an output buffer. BNN weight matrix is distributed across the PEs and stored locally in on-chip memory. Subsequently, the input images are streamed through the MVTU and multiplied with the weight matrix. Particularly, the PE computes the dot-product between an input vector and a row of weight matrix, each of $S$-bits wide, using an XNOR gate, as shown in Fig.~\ref{MVTU_PE_Arch}. Then, it compares the number of \textit{set} bits to a threshold and produces a 1-bit output value as previously discussed.

\Figure[t](topskip=0pt, botskip=0pt, midskip=0pt)[width=0.35\textwidth]{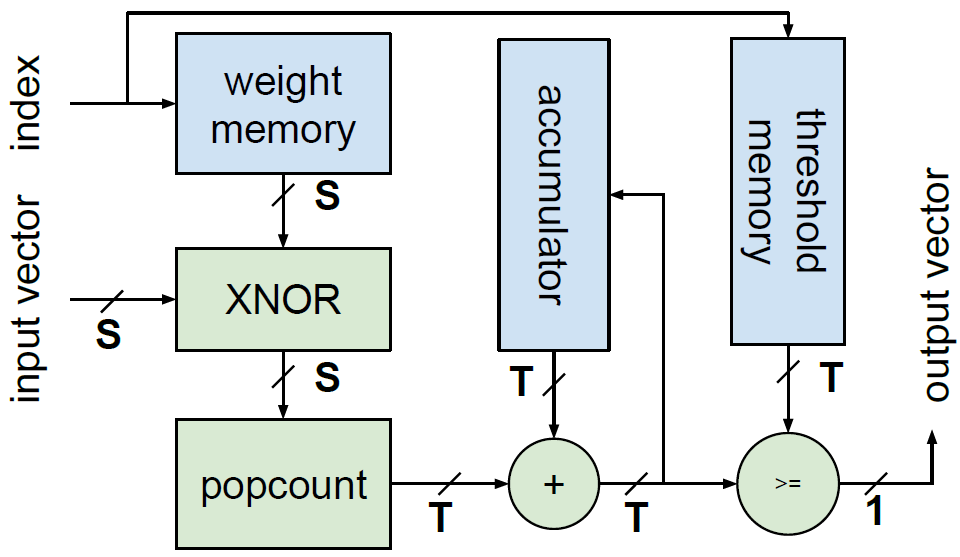}
{The Architecture of MVTU PE~\cite{umuroglu2017finn}.
\label{MVTU_PE_Arch}}

\Figure[b](topskip=0pt, botskip=0pt, midskip=0pt)[width=0.48\textwidth]{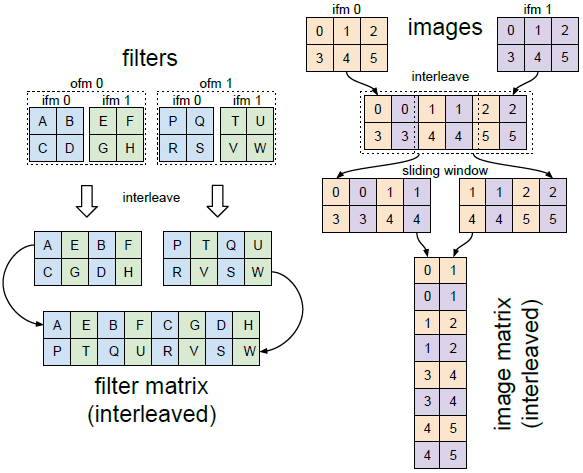}
{Transforming CONV to Matrix-Multiplication~\cite{umuroglu2017finn}, where, ifm and ofm are the input and output feature maps, respectively.
\label{MVTU_SWU_Arch}}

\Figure[t](topskip=0pt, botskip=0pt, midskip=0pt)[width=0.63\textwidth]{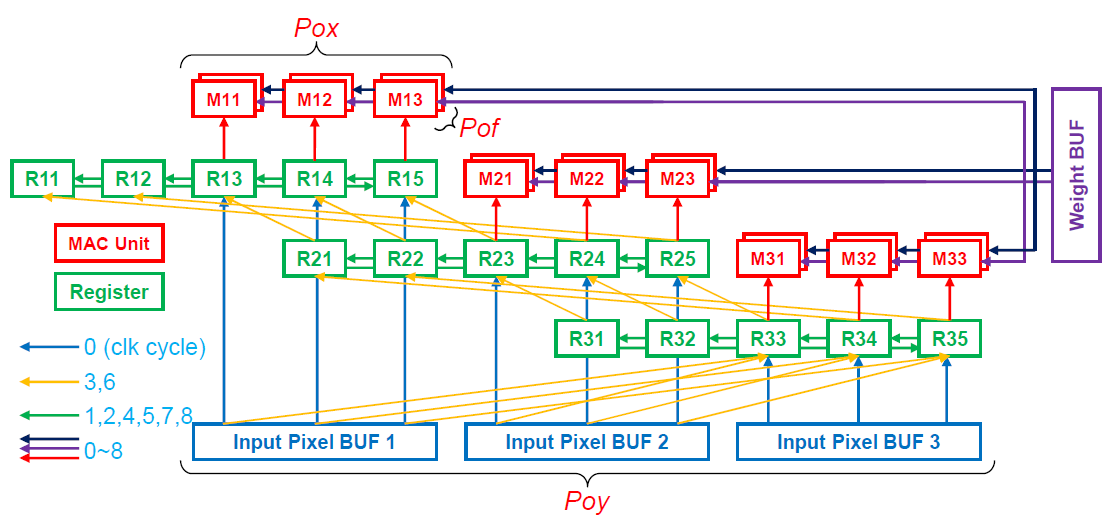}
{CONV Acceleration Architecture and Dataflow~\cite{ma2017optimizing}, where, $Pix=Pox=3$, $Piy=Poy=3$, and $Pof=3$.
\label{CONV_Loops_Arch}}

Umuroglu et al.~\cite{umuroglu2017finn} implemented the CONV layer using a sliding window unit (SWU) and an MVTU, where convolutional operation is transformed to matrix-multiplication of image matrix and filter matrix. SWU generates the image matrix to MVTU by moving the sliding window over the input FMs, while the filter matrix is generated by packing the weights from the convolution filters as shown in Fig.~\ref{MVTU_SWU_Arch}. In order to meet the user throughput requirement, MVTU is folded (time-multiplexed) by controlling the values of $P$ and $S$. Folding of MVM decides partitioning of the matrix across PEs. Every row of matrix tile is mapped to a distinct PE and every column of PE buffer is mapped to a distinct SIMD lane. In this away, the required number of cycles to compute one MVM (total fold) is obtained as $(X \times Y) / (P \times S)$, where $X$ and $Y$ are the dimensions of the matrix. The folding factors of BNN layers have been determined such that every BNN layer takes nearly the same number of cycles.

To evaluate FINN, the authors  implemented CNV topology on Xilinx Zynq-7000 board at 200 MHz to accelerate BNNs inference on CIFAR-10~\cite{krizhevsky2009learning}. CNV contains three repetitions of two $3 \times 3$ CONVs and $2 \times 2$ max-pooling layers. Its topology is inspired by VGG-16 and BinaryNet~\cite{courbariaux2016binarynet}. Although CNV accepts images with 24-bits/pixel as an input and produces a 10-element vector of 16-bit values, 2-bits are used for representing intermediate results while 1-bit is used for representing CONV and FC weights.  Experimental results demonstrated that the proposed design provides high performance (2.5 TOPS) while incurring low energy consumption (11.7 Watts). FINN outperforms the design by Ovtcharov et al. \cite{ovtcharov2015accelerating} by over 13.8$\times$ for throughput.

In~\cite{ma2017optimizing}, loop optimization techniques~\cite{bacon1994compiler, zhang2015optimizing} have been employed in FPGA to design a customized CNN accelerator through speeding up CONV layer operations. Firstly, an in-depth analysis is provided to numerically characterize loop unrolling, loop tiling, and loop interchange optimization techniques. In doing so, 8 CONV dimensions parameters ($N^*$), 8 loop unrolling design variables ($P^*$), and 8 loop tiling design variables ($T^*$) have been used with a constraint, as for a specific loop level, $1 \leq P^* \leq T^* \leq N^*$. Note that unrolling \textit{Loop-1} and \textit{Loop-3} requires $Pkx \times Pky$ and $Pif$ multipliers, respectively, an adder tree with fan-in of $Pkx \times Pky$ and $Pif$, respectively, and an accumulator. On the other hand, unrolling \textit{Loop-2} requires $Pix \times Piy$ parallel units of MAC to reuse the same weight for 
$Pix \times Piy$ times, while the input feature pixel can be reused by $Pof$ times when unrolling \textit{Loop-4} with the use of $Pof$ parallel MAC units. Thus, $Pkx \times Pky \times Pif \times Pix \times Piy \times Pof$ multipliers are required. Please refer to Fig.~\ref{CONV_Loops_Unrolling} for more details on CONV loops levels and their parameters. In loop tile optimization, the authors have numerically set the lower bound on the required size of the input pixel buffer, the weight buffer, and output pixel buffer that ensures reading each input feature pixel and weight from the off-chip memory only once. On the other hand, loop interchange technique has a great impact on the times of memory access as well as the number of partial sums since it determines the order of computing CONV loops.

Secondly, the authors have provided a quantitative analysis of the design variables to minimize each of computing latency, partial sum storage, on-chip buffer access, and off-chip DRAM access. Subsequently, MATLAB scripts are used to randomly sample a subset of the solution space to find the optimal design configurations. This is due to the large  solution space, more than $7.2 \times 10^{13}$ possible configurations for loop tiling variables of width ($Pox$) and height ($Poy$) output FM alone. According to the randomly sampling results for VGG-16 CNN model on Arria 10 GX 1150 FPGA, uniform unrolling factors for CONV layers are used with $Pix=Pox=Piy=Poy=14$ and $Pof=16$ for \textit{Loop-2} and \textit{Loop-4}, respectively, to reuse input feature pixels and weights. On the other hand, \textit{Loop-1} and \textit{Loop-3} are serially computed to prevent the movement of the partial sums between the MAC units and consume them ASAP since both \textit{Loop-1} and \textit{Loop-3} need to be finished in order to obtain one final output pixel. More importantly, the order of loops computation has been found to be as follows. \textit{Loop-1} is computed first, then comes \textit{Loop-3}, and finally \textit{Loop-2} and \textit{Loop-4} are computed in any order.

Finally, a customized convolution accelerator module with efficient dataflow has been designed based on the previous results and used for all VGG-16 CONV layers. The CONV accelerator consists of 3,136 ($Pix \times Piy \times Pof$) independent MAC units and 14 ($Pof$) input pixel buffers. Fig.~\ref{CONV_Loops_Arch} shows an example of the designed CONV accelerator when $Pix$, $Piy$, and $Pof$ are all equal to 3. The input pixels are shifted after fetching them out of the input pixel buffers. Subsequently, they can be reused among the input register arrays. Then, the input pixels are fed into the associated MAC units. The figure also shows that the input pixels and weights are shared by $Pof$ and $Pix \times Piy$ MAC units, respectively.

The overall CNN acceleration system mainly consists of two SDRAM banks that hold the input feature pixels and weights, two modular Scatter-Gather DMA (mSGDMA) engines to facilitate the simultaneous read/write from/to the SDRAMs, and a controller to govern the sequential computation of layers as well as the iterations of the four CONV loops. On the other hand, dual weight buffers have been used to increase the throughput of FC layer through overlapping the inner-product computation with off-chip communication. The acceleration system has been written as parametrized Verilog scripts. The experimental results show that the proposed accelerator has a throughput of 645.25 GOPS, which is more than $3.2\times$ enhancement compared to prior VGG-16 FPGA-based implementations~\cite{suda2016throughput, qiu2016going}.

Venieris and Bouganis~\cite{venieris2017latency} further extended fpgaConvNet framework~\cite{venieris2016fpgaconvnet} to allow for optimizing either throughput or latency depending on the size of the workload. For large workloads, weights reloading transformation has been introduced to efficiently design latency-critical CNNs on FPGA. In contrast with fpgaConvNet, where a distinct architecture is designed for each subgraph, the weights reloading transformation allows for generating a single flexible architecture, named as the reference architecture and derived using pattern matching, to execute the workloads of all subgraphs by transitioning to different modes. Upon the execution of a new subgraph, the subgraph’s weights are read into the on-chip memory and the multiplexers are configured to form the appropriate datapath. Fig.~\ref{Weights_reloading} demonstrates how weights reloading is applied. The authors have mentioned that the required time for transferring subgraph’s weights is much smaller than the average time for full FPGA reconfiguration, 272.7$\times$ less when loading 4.5 MB of weights for a VGG-16 layer on Zynq XC7Z045.

\Figure[!b](topskip=0pt, botskip=0pt, midskip=0pt)[width=0.45\textwidth]{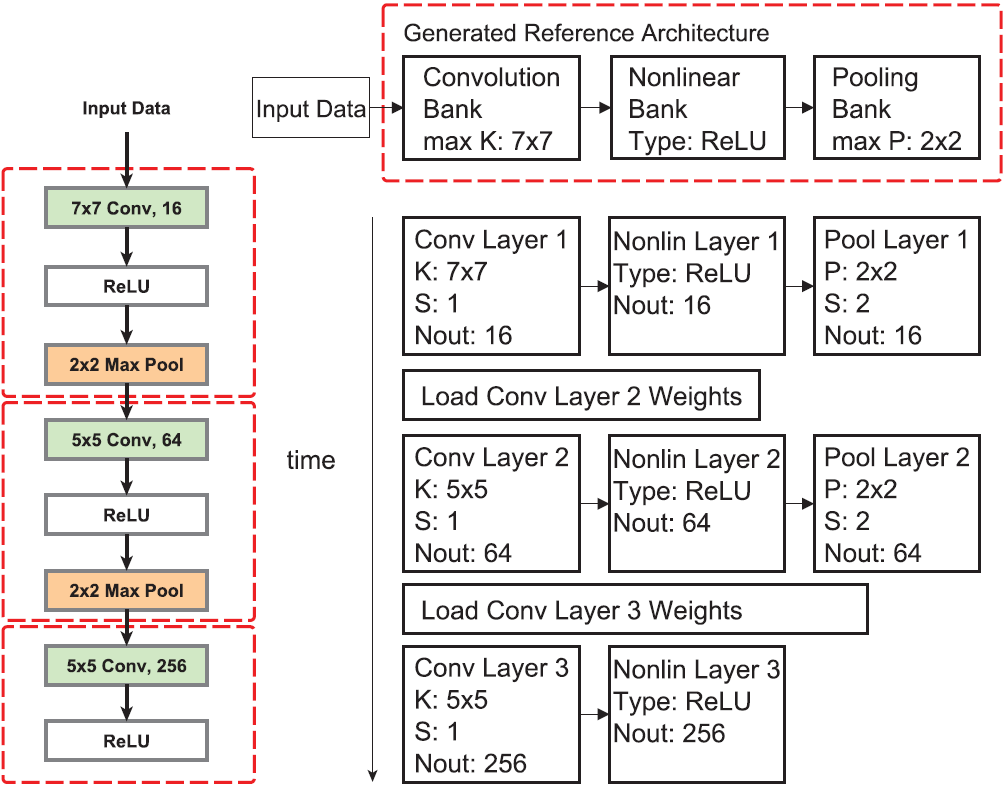}
{Weights Reloading~\cite{venieris2017latency}. 
\label{Weights_reloading}}

In the situation discussed above, due to limited on-chip memory capacity, it might not be possible to load all weights required for a single CONV layer. To handle this, the authors  introduced an input FMs folding factor ($f_{in}$) with each CONV layer. A CONV layer (${CONV}_{i}$) is  partitioned into ${f_{in}}_{i}$ subgraphs in which each subgraph executes a fraction of ${CONV}_{i}$ to produce a fraction of the output FMs. The proposed latency-driven methodology has been evaluated by implementing AlexNet and VGG-16 with 16-bit fixed-point precision for both on Zynq XC7Z045 at 125 MHz. The experimental results showed 1.49$\times$ and 0.65$\times$ higher CONV throughput than DeepBurning~\cite{wang2016deepburning} and the embedded FPGA accelerator in \cite{qiu2016going} for AlexNet and VGG-16 implementations, respectively.

Lavin and Gray~\cite{lavin2016fast} demonstrated that CNN algorithms with small filters can be efficiently derived using Winograd algorithm~\cite{winograd1980arithmetic} and fast Fourier transform (FFT) algorithm~\cite{van1992computational} due to their advantages in improving resource efficiency and reducing arithmetic complexity. Winograd computation involves a mix of element-wise (Eltwise) and general-purpose matrix multiplication, where some of the matrices need to be transformed. In particular, Winograd algorithm exploits the structure similarity among $n \times n$ tiled input FM pixels given a filter of size $r \times r$ to generate $m \times m$ tiled pixels of the output FM, where $m$ represents the stride between Winograd tiles ($m=n-r+1$),  while minimizing the number of required CONV multiplications from $m^{2}r^{2}$ for conventional CONV algorithm to $n^{2}$. In another work, Zhang et al.~\cite{zhang2017frequency}  implemented FFT algorithm for CNN on FPGA platform. However, their proposed implementation  shows little reduction of computation complexity with small filters such as $3 \times 3$.

\Figure[t](topskip=0pt, botskip=0pt, midskip=0pt)[width=0.48\textwidth ]{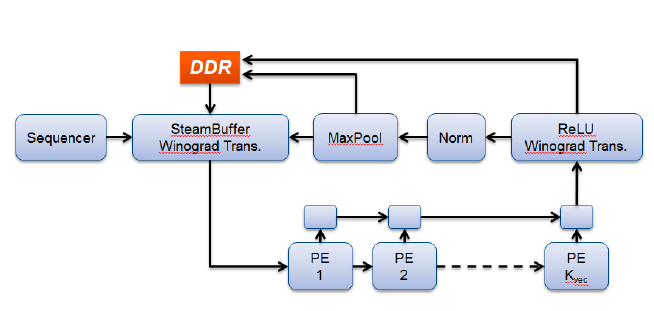}
{Overall DLA Architecture~\cite{aydonat2017opencl}.
\label{DLA}}

Aydonat et al. \cite{aydonat2017opencl} presented a deep learning architecture (DLA) based on OpenCL.
Their proposed architecture reduces the external memory bandwidth requirements by an order-of-magnitude for both the convolutional and fully connected layers. This is achieved by caching all intermediate feature maps on-chip in stream buffers. For fully connected layers, image batching is used where a batch of images are processed together through the fully connected layers. The approach utilizes the Winograd transformation to reduce the multiply-accumulate operations,  which could reduce the number of needed operations by about 50\%. In addition, it uses half-precision (FP16) 
floating-point operations with shared exponents, which significantly reduces the needed computational resources.  

The overall DLA architecture is shown in Fig.~\ref{DLA}. Each PE consists of dot-product units, accumulators, and caches, for performing dot-products for convolution and fully connected layers. Caches are used 
for storing filter weights. To avoid idle computation cycles, double-buffering is used such that 
filter weights for the next convolution layer are  prefetched onto the caches
while filter weights are loaded from the caches for a particular convolution layer.
Stream buffers store feature data and stream it to PEs. Each stream buffer is double-buffered similar
to filter caches. 
Images are loaded from the DDR and are stored in stream buffers before the first convolution layer starts execution.
During a convolution layer execution, while feature data
for a convolution layer is being streamed into the PEs, the
outputs of convolutions are simultaneously stored in the
buffers.
The StreamBuffer unit applies the
Winograd transformations to features, and streams the transformed features to the first PE which are forwarded through all the PEs via the daisy-chained input connections between them.
The ReLU unit receives the outputs of the PEs via daisy-chained output connections.
Then,  the normalization unit receives the outputs of the ReLU unit and applies the normalization formula across the feature maps. The pooling unit receives  the outputs of the normalization unit and computes the maximum value in a window. 
The output of the pooling unit is stored back in the stream buffer
for further processing, if more convolution layers are to follow. Otherwise, 
the outputs of the pooling unit are stored in external memory.
For the fully connected layers,  features data are stored on PEs caches while filter weights are stored in stream buffers. For the first fully connected layer, features data
are read back from external memory and loaded onto the PE caches. The ReLU output is sent
directly to DDR, without applying normalization or pooling. 
The sequencer generates the control signals to control the operation of the various blocks in DLA according to
the topology of the executed CNN. Executing a different CNN requires just changing the sequencer
configuration.

The DLA has been  evaluated by implementing AlexNet CNN
on Intel's Arria 10 dev kit which contains a A10-1150
device (20nm) using a 96 batch size for the fully connected layers.
It   achieved a performance of 
1020 images/s. In addition, it   achieved 8.4x more GFLOPS than the latest
Ultrascale (KU 20nm) result reported in \cite{zhang2016caffeine}, which uses a 32 batch size
for the fully connected layers, and  19$\times$ more GFLOPS
than the latest Stratix V result reported in \cite{suda2016throughput}.
Furthermore, it has achieved energy efficiency at 23 images/s/W,  which is similar to what is achieved with the 
best publicly known implementation of AlexNet on NVIDIA Titan X GPU.

Unlike DLA architecture~\cite{aydonat2017opencl} where a 1D Winograd algorithm was employed to reduce arithmetic complexity, Lu et al.~\cite{lu2017evaluating} implemented a novel FPGA architecture with a  two-dimensional Winograd algorithm~\cite{winograd1980arithmetic} to accelerate convolutional computation of CNNs. The overall architecture consists of line buffer structure and Winograd PE engine, as shown in Fig.~\ref{Winograd-based-CNN-Accelerator}. Particularly, $n+m$ input lines and $m$ output lines of on-chip buffers are used to effectively reuse FM data among different tiles. While Winograd PE engine reads the first $n$ input lines to perform Winograd computation, the next $m$ input lines load pixels from off-chip memory using FIFOs to overlap the data transfer and computation. Thereafter, the input lines are rotated in a circular fashion to make the next $n$ input lines ready. On the other hand, Winograd PE engine composed of 4 pipelined stages performs transformation, element-wise matrix multiplication, additional transformation, and accumulation of output tiles, respectively.

\Figure[t](topskip=0pt, botskip=0pt, midskip=0pt)[width=0.48\textwidth]{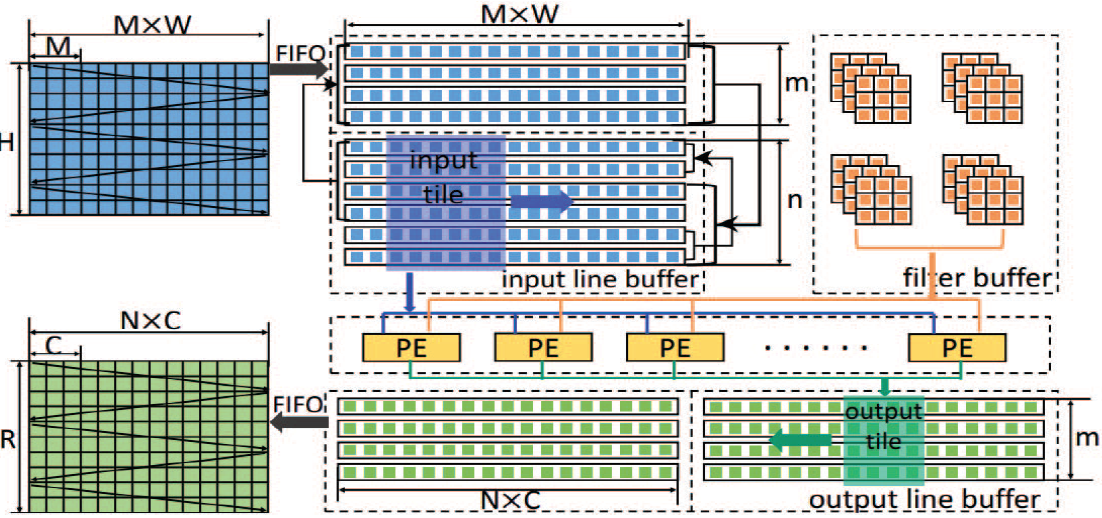}
{Winograd-based CNN Accelerator~\cite{lu2017evaluating}, where, $m$ is the size of the input FM tile, $n$ is the size of the output FM tile, $M$ is the number of input  channels, $N$ is the number of output channels, $W$ is the maximal width of all input FMs, $C$ is the width of the output FMs.
\label{Winograd-based-CNN-Accelerator}}

A vector of PEs is employed to achieve parallelism through unrolling $Loop-4$ ($Pof$) and $Loop-3$ ($Pif$) similar to that in~\cite{zhang2015optimizing}. To implement FC layer, the proposed accelerator uses the input line buffers to hold FC weights while input neurons are stored on the filter buffers. Then, Winograd PE engine is reused to implement FC operation but with bypassing the transformation stages. Moreover, a batch ($N_{batch}$) of input FMs are assembled and processed together in order to improve the memory bandwidth. An analytical model has been proposed for a fast design space exploration of optimal design parameters ($n$, $Pof$, $Pif$, $N_{batch}$) constrained by FPGA configuration with a 16-bit fixed-point representation for both FM data and filter.

The proposed accelerator has been evaluated by implementing AlexNet and VGG-16 on Xilinx ZCU102 FPGA. AlexNet CONV layers have 3 different filters. Conventional CONV algorithm has been applied to the first CONV layer as it has a filter of size $11 \times 11$ while a uniform filter of size $3 \times 3$ for Winograd algorithm has been used to implement the rest of the layers. The design parameters are found to be equal to (6, 4, 8, 128) and (6, 4, 16, 128) for AlexNet and VGG-16, respectively. The experimental results demonstrated that the proposed Winograd-based CNN accelerator has an average performance of 854.6 GOPS and 2940.7 GOPS for AlexNet and VGG-16, respectively, with power consumption of 23.6 Watts for both. The proposed accelerator has also been evaluated on Xilinx ZC706 platform where the design parameters are found to be as (6, 2, 8, 32) and (7, 4, 4, 32) for AlexNet and VGG-16, respectively. The experimental results demonstrated that Winograd-based CNN accelerator has an average performance of 201.4 GOPS and 679.6 GOPS for AlexNet and VGG-16, respectively, with power consumption of 23.6 Watts for both. Compared to the implementation of VGG-16 on NVIDIA Titan X with the latest CuDNN 5.1, Titan X gives better performance than Xilinx ZC706 but the implementation on Xilinx ZC706 achieves 1.73$\times$  higher energy  efficiency.

Zhang et al.~\cite{zhang2017improving} presented an OpenCL-based architecture for accelerating CNNs on FPGA. They also proposed an analytical performance model to identify the bottleneck in OpenCL-based acceleration of VGG-19 CCN model on modern FPGA platforms such as Altera Arria 10 GX 1150. Based on roofline mode analysis, it is shown that the bandwidth requirement of VGG-19 workload is higher than what is provided by the FPGA board. Thus, they  identified on-chip memory bandwidth as the key performance bottleneck. In addition, they  observed that exploited data-level parallelism in the existing Altera OpenCL library~\cite{czajkowski2012opencl2} leads to wasteful replication of on-chip memory (BRAM). This is due to connecting each PE with a dedicated BRAM port. 

\Figure[t](topskip=0pt, botskip=0pt, midskip=0pt)[width=0.30\textwidth, height = 4.7 cm]{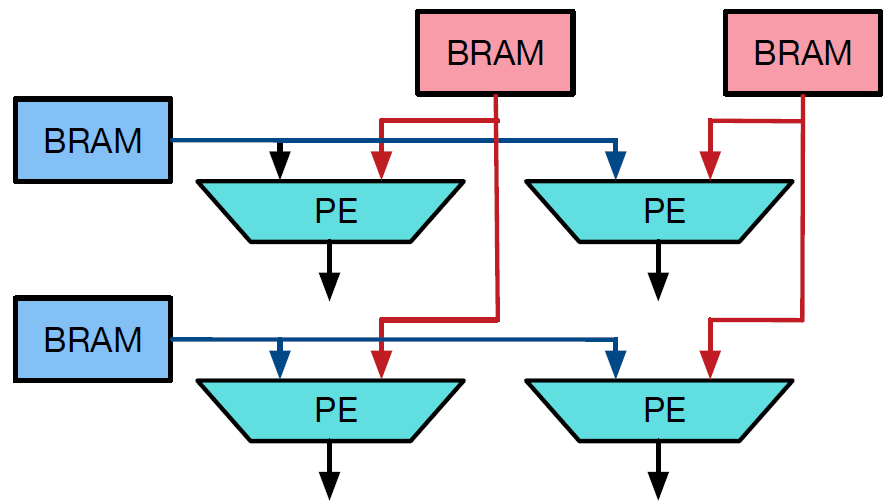}
{Compute Unit with a 2D BRAM-to-PE Interconnection~\cite{zhang2017improving}.
\label{fig:2-D_BRAM-to-PE_Interconnection}}

\Figure[b](topskip=0pt, botskip=0pt, midskip=0pt)[width=0.30\textwidth, height = 4.7 cm]{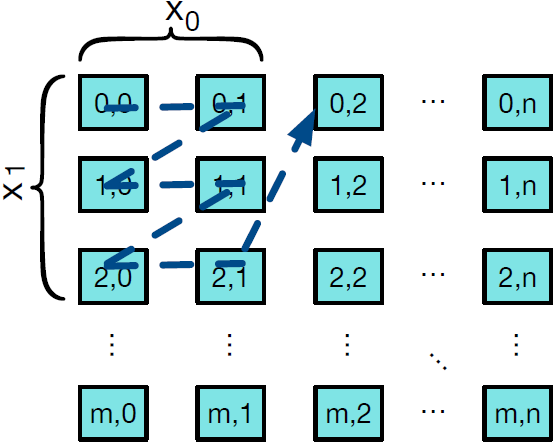}
{2D Dispatcher~\cite{zhang2017improving}, where, $X_{0}$ is the column size of kernel buffer as well as the row size of the input feature buffer, and $X_{1}$ is the row size of kernel buffer.
\label{fig:2-D_dispatcher}}

\Figure[t](topskip=0pt, botskip=0pt, midskip=0pt)[width=0.48\textwidth]{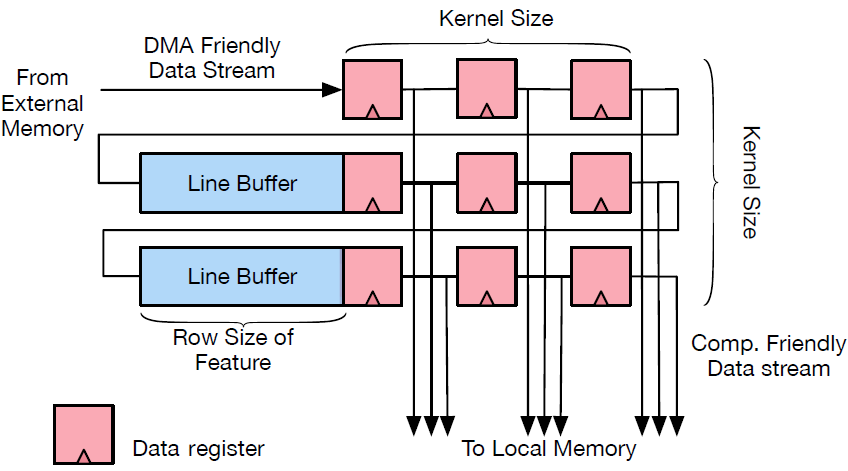}
{Line Buffer Design~\cite{zhang2017improving}.
\label{fig:Line_Buffer_Design}}

Therefore, a Verilog-based accelerator kernel has been designed and warped to an OpenCL IP in order to optimally balance on-chip memory bandwidth with workload computational throughput and off-chip memory accesses. In particular, the proposed kernel consists of a compute subsystem, a local memory subsystem, and a 2D dispatcher. The compute subsystem is organized hierarchically into compute units (CUs) and PEs. At PE level, the authors have designed a 2D multi-cast interconnection between BRAMs (32-bit data width) and PEs to improve the efficiency of on-chip BRAM usage by sharing the data of one BRAM port with several PEs as shown in Fig.~\ref{fig:2-D_BRAM-to-PE_Interconnection}. The CU has been designed as a 2D PE array of size $16 \times 16$ to match  the computational bandwidth with the maximum streaming bandwidth (512-bit data bus) provided by off-chip memory. The 2D dispatcher divides the work items into work groups each of size ($X_{0}$, $X_{1}$) as shown in Fig.~\ref{fig:2-D_dispatcher}. Thereafter, it adaptively schedules the work items within each work group to the CUs starting with the lowest dimension to balance the memory bandwidth with capacity. The 2D dispatcher is also responsible for host/device memory data transfers. In addition, the authors have limited the maximum fan-out for registers to 100 in order to guarantee a higher frequency.

The CONV layer has been implemented as a matrix multiplication by flattening and rearranging the data using line buffer~\cite{bosi1999reconfigurable}, as shown in~Fig.~\ref{fig:Line_Buffer_Design}, in a similar fashion to that in~\cite{suda2016throughput}. The line buffer converts continuous address stream from external memory into a stream conducive for CONV operation to substantially reduce the bandwidth requirement of off-chip memory. To implement FC layer, the proposed accelerator uses one column of PEs in the CU. The proposed implementation has achieved 866 GOPS and 1790 GOPS with the use of 32-bit floating-point and 16-bit fixed-point, respectively, under 370 MHz and 385 MHz working frequencies, respectively.

\Figure[t](topskip=0pt, botskip=0pt, midskip=0pt) [width=0.54\textwidth]{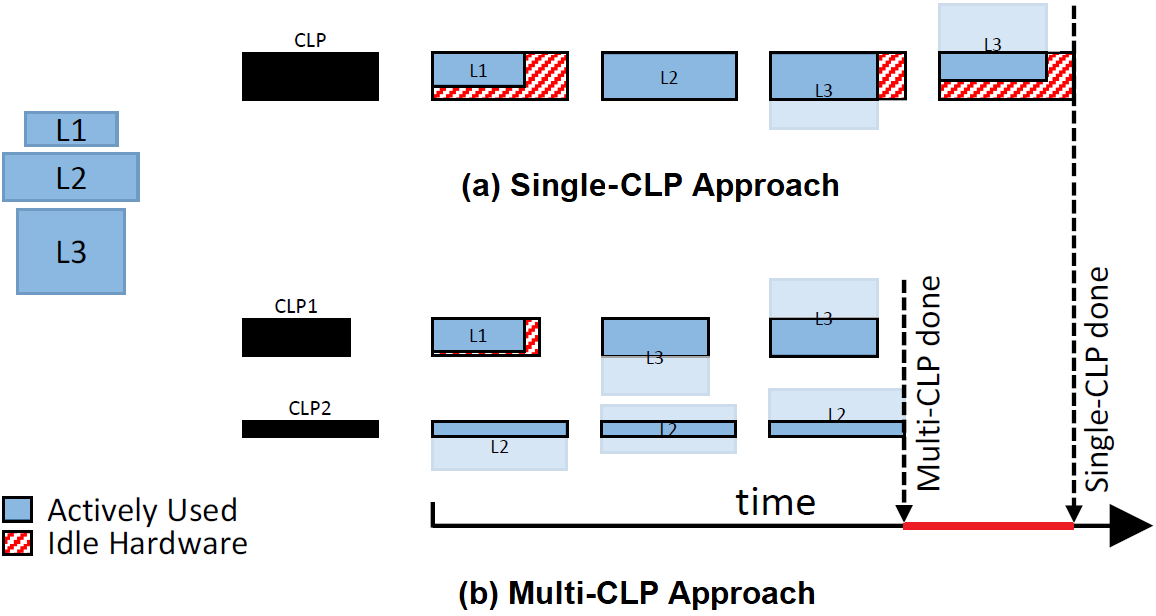}
{Operation of CONV Layer Processors (CLPs) on CNN with three CONV Layers~\cite{shen2017maximizing}.
\label{Multi-CLP_Arch}}

All previously discussed FPGA-based CNN accelerators, except the ones discussed in~\cite{venieris2016fpgaconvnet, liu2017throughput}, have employed a single CLP to maximize the aggregate throughput of performed consecutive convolutional operations. However, Shen et al.~\cite{shen2017maximizing} noted that using a single globally-optimized CLP design for the computation of CONV layers of radically different configurations and dimensions leads to sub-optimal performance and insufficient utilization of FPGA resources. Fig.~\ref{Multi-CLP_Arch}a demonstrates the use of a single CLP to iteratively process $L1$ , $L2$, and $L3$ CONV layers where the dimensions of the hardware ($CLP$, $CLP1$, and $CLP2$) and the layers are represented by the size and shape of the boxes. It is clear that computing $L1$ and  portions of $L3$ leaves FPGA resources unutilized as their dimensions are smaller than the dimension of the used CLP. Note that processing a CONV layer with a dimension bigger than the dimension of CLP, such as $L3$, requires the repeated use of CLP to process different portions of the layer.

The authors have also followed the methodology in~\cite{zhang2015optimizing} to derive an optimal single CLP, through finding the optimal unrolling factor $\langle T_{m}, T_{n} \rangle$, for implementing SqueezeNet~\cite{iandola2016squeezenet} and AlexNet on Virtex7 690T FPGA with a single precision floating-point and 16-bit fixed-point arithmetic units, respectively. They found that one quarter of DSP slices of SqueezeNet’s CLP remain unused.  Even more worse utilization has  been observed for AlexNet. The optimal single CLP has not utilized, on average, more than one quarter of the arithmetic unit resources.

On the other hand, they also noted that using one CLP for each stage of CONV layer in a fashion similar to that in~\cite{li2016high} is not efficient due to three reasons. First, it reduces the on-chip BRAM buffer size of each CLP which minimizes overall data locality. Second, such one-to-one mapping of CONV layers and CLPs requires orchestrating many off-chip memory accesses which incurs latency and bandwidth overheads. Third, the overall control overhead scales with the number of CLPs which leaves insufficient resources for the computation of CNN.

To address the above inefficiencies, Shen et al.~\cite{shen2017maximizing}  proposed a multi-CLP accelerator system for CNNs where the available PFGA hardware resources are partitioned across multiple smaller CLPs.  Each CLP is tailored with a dimension that closely matches the dimensions of a subset of CONV layers. Thereafter, these specialized CLPs are used to concurrently operate on a batch of images to achieve a higher overall throughput, as shown in Fig.~\ref{Multi-CLP_Arch}b, where the same hardware in Fig.~\ref{Multi-CLP_Arch}a is partitioned into two parallel CLPs; $CLP1$ and $CLP2$.

Shen et al.~\cite{shen2017maximizing} developed an optimization search algorithm that uses dynamic programming  to find  optimal designs. For given configurations of CNN model (i.e., CONV layers descriptions) and resource constraints of the targeted FPGA platform (i.e., number of DSP slices, BRAM-18Kb units, and off-chip memory bandwidth), it derives the optimal number of CLPs (along with their $\langle T_{m}, T_{n} \rangle$ dimensions) as well as the optimal mapping between CONV layers and CLPs that maximize the performance. The assignment of CNN layers to CLPs is static, where each CNN layer is mapped and bounded to a particular CLP. Subsequently, CNN layers are pipelined to their CLP, as shown in Fig.~\ref{Multi-CLP_Arch}b, where  $L1$ and $L3$ are pipelined to $CLP1$ while $L2$ is repeatedly processed on $CLP2$ with very little idle hardware which improves the performance compared to single CLP approach. Moreover, the optimization algorithm also finds the optimal partition of on-chip BRAM resources of each CLP that minimizes the overall off-chip memory accesses. Note that the optimal dimension of each CLP is found based on the work in~\cite{zhang2015optimizing}.

Subsequently, C++ (HLS) templates are parameterized to design CLPs and to form a complete implementation of CNN. A standard AXI crossbar is used to interconnect the independent CLPs. The ping-pong double-buffering technique is also used for input FMs, output FMs, and weights to allow for transferring data while computation is in progress. The experimental results of implementing AlexNet with a single precision floating-point using multi-CLP accelerator on Virtex7 485T and 690T FPGAs at 100 MHz demonstrate 1.31$\times$ and 1.54$\times$ higher throughput than the state-of-the-art single CLP design in~\cite{zhang2015optimizing}, respectively. For the more recent SqueezeNet network, the proposed multi-CLP accelerator results in speedup of 1.9$\times$ and 2.3$\times$ on Virtex7 485T and 690T FPGAs at 170 MHz with 16-bit fixed-point, respectively.

Wei et al. \cite{wei2017automated} presented a systolic architecture for automatically implementing 
a given CNN on FPGA based on OpenCL description, maximizing clock frequency and resource utilization. 
The proposed systolic architecture is shown in Fig.~\ref{Systolic}.
Each PE shifts the data of the weights (W) and inputs (IN) horizontally and vertically to the neighboring PEs 
in each cycle. The 2D structure of PEs is designed to match the FPGA 2D layout structure 
to reduce routing complexity and achieve timing constraints.

\Figure[t](topskip=0pt, botskip=0pt, midskip=0pt)[width=0.4\textwidth ]{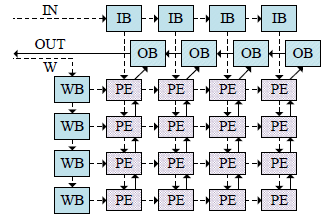}
{Systolic Array Architecture for CNN~\cite{wei2017automated}.
\label{Systolic}}

The technique first finds a feasible mapping 
for the given CNN to the systolic array to guarantee that  proper data is available
at specific locations in the PE array at every cycle. Then, the size of PE array (dimensions) is determined
which has an impact on the required number of DSPs, the clock frequency, and the DSPs efficiency.
Finally, the data reuse strategy is determined by choosing proper tiling sizes. 
The proposed technique has been evaluated using AlexNet and VGG16 on  Intel’s Arria 10 GT 1150 board.
The technique has explored the use of  both 32-bit floating-point and fixed-point using 8-bits for weights and 16-bits for data.
Evaluation results show that,  for the VGG16 CNN, the technique  achieves
up to 1,171 GOPS on Intel’s Arria 10 device with a clock frequency of 231.85 MHZ and (8-16)-bit fixed-point 
representation.

In another recent research work, Ma et al.~\cite{ma2017end} generalized the previously proposed accelerator in~\cite{ma2017optimizing} to efficiently accelerate ResNet-50 and ResNet-152 on Arria 10 GX 1150 FPGA. In doing so, they designed flexible and scalable CONV, ReLU, BatchNorm, scale, pooling, FC, and Eltwise primitives. In addition, local control logic and registers have been used with each primitive to control their computation order and to hold their configurations, respectively. By doing so, ResNets primitives can be efficiently reused for different parameters of each layer.

For ResNets scalable CONV primitive, there are four (kernel, stride) size configurations; ($3 \times 3$, 1), ($1 \times 1$, 1), ($1 \times 1$, 2), and ($7 \times 7$, 2). Therefore, a similar architecture and dataflow to that shown in Fig.~\ref{CONV_Loops_Arch} has been used for CONV but with the use of two sets of register arrays; with shifting between the registers (which is shown in Fig.~\ref{CONV_Loops_Arch}, \textit{Set-1}), and without shifting between the registers (\textit{Set-2}). The CONV primitive with $3 \times 3$ kernel and stride of 1 uses \textit{Set-1} register array, while \textit{Set-2} is used with ($1 \times 1$, 1), ($1 \times 1$, 2), and ($7 \times 7$, 2) configurations. In CONV primitive with \textit{Set-2}, the input pixels are fed from the input pixel buffers into the corresponding registers without shifting, and then to MAC units. The skipped input pixels in ($1 \times 1$, 2) configuration are not stored to the input pixel buffers. On the other hand, the ($7 \times 7$, 2) configuration of the kernel and stride sizes is retained as the ($1 \times 1$, 1) case while transferring repeated input pixels into the input pixel buffers and rearranging their storage patterns. The CONV primitive  also takes care of zero-paddings for different (kernel, stride) size configurations.

The loop unrolling and tiling techniques in~\cite{ma2017optimizing} have also been employed to accelerate CONV primitive with a uniform mapping of PEs to all ResNets CONV layers. However, designing of efficient CNN modules is not enough, as the memory accesses and data movements between these modules must  also be minimized. Therefore, the authors have designed a layer-by-layer computation flow. The global control logic is responsible for governing the sequential operations of primitives and their dataflow through predefined and preloaded layered-based execution flowchart, as shown in Fig.~\ref{ResNet_Execution_Flowchart}. In addition, it has been modeled to reconfigure ResNet primitives according to the parameters of each layer during  runtime. For instance, it maps a particular number of PEs to CONV layer based on loop unrolling parameters as well as it controls the selection of register array type (\textit{Set-1} or \textit{Set-2}) based on CONV (kernel, stride) parameters.

\Figure[b](topskip=0pt, botskip=0pt, midskip=0pt)[width=0.48\textwidth, height = 5.8 cm]{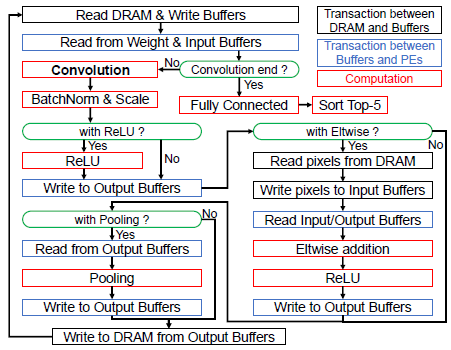}
{Execution Flowchart of ResNets Layers~\cite{ma2017end}.
\label{ResNet_Execution_Flowchart}}

On the other hand, a custom DMA manager has been designed to control the operations of DMA. Note that the DMA is responsible for transferring the input FM pixels, weights, and output FM pixels between off-chip memory and on-chip buffers. Unlike ALAMO architecture~\cite{ma2018alamo} where the output pixels are only stored in on-chip buffers, this work as well as the work discussed in~\cite{ma2017optimizing} store the output pixels in off-chip memory with the use of loop tiling technique in order to have a flexible architecture that can process large-scale CNNs. The dual weight buffers technique has not been used in this work due to the current trend in CNNs where either the size of FC weights has been significantly reduced (2 M in ResNet compared with 123.6 M in VGG) or the FC layers are completely removed such as in NiN. The experimental results demonstrated that the achieved throughput for ResNet-50 and ResNet-152 are 285.1 GOPS and 315.5 GOPS, respectively. Finally, the authors mentioned that higher throughput can be achieved using batch computing~\cite{li2016high}.

Wang et al.~\cite{wang2017dlau} proposed a scalable design on FPGA for accelerating deep learning algorithms. In order to provide a  scalable architecture and support various deep learning applications, the proposed architecture utilizes the tiling technique in which the large-scale input data is partitioned into small subsets. The size of the tile is configured to leverage the trade-off between the hardware cost and the speedup. Moreover, the authors explored  hot spots profiling to determine the computational parts that need to be accelerated to improve the performance. The experimental results illustrated that matrix multiplication and activation functions are the key operations in deep learning algorithms as they consume about 98.6\% and 1.1\% of the overall execution time, respectively. Thus, the proposed accelerator is responsible for speeding up both matrix multiplication and activation function computations.

\Figure[t](topskip=0pt, botskip=0pt, midskip=0pt)[width=0.44\textwidth, height = 5 cm]{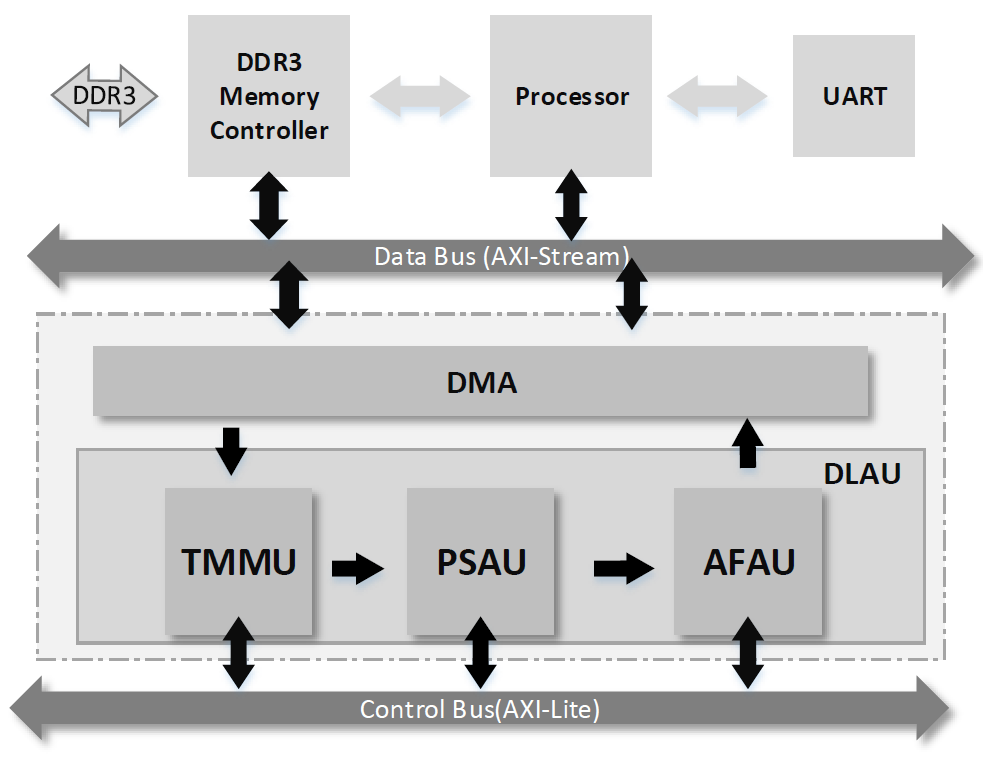}
{DLAU Accelerator Architecture~\cite{wang2017dlau}.
\label{DLAU_Arch}}

The main components of the proposed architecture are the embedded processor, the DDR3 memory controller, the DMA module, and the deep learning acceleration unit (DLAU), as shown in Fig.~\ref{DLAU_Arch}. The embedded processor utilizes the JTAG-UART to communicate with the acceleration unit~\cite{jtag2018altera}. The  DLAU unit accesses the DDR3 memory to read the tiled input data and to write the results back through the DMA module during the execution. The DLAU utilizes three fully pipelined processing units to improve the throughput, while minimizing the memory transfer operations. These units are tiled matrix multiplication unit (TMMU), partial sum accumulation unit (PSAU), and activation function acceleration unit (AFAU). TMMU is responsible for multiplication and generating the partial sums. To optimize the performance, TMMU is structured as a pipelined binary adder tree. Moreover, it uses two sets of registers alternately to overlap the computation with the communication, one group is used for the computation, while in parallel, the other group is loaded with the next node data every clock cycle. On the other hand, PSAU is responsible for accumulating the partial sums generated from TMMU. Finally, AFAU implements the sigmoid function using piecewise linear interpolation to speedup the computation with negligible accuracy loss. Since the processing units in DLAU might have inconsistent throughput rates, each unit has input FIFO buffer and output FIFO buffer to prevent  data loss.

The authors implemented the proposed architecture on Xilinx Zynq Zedboard with ARM Cortex-A9 processors clocked at 667 MHz. In addition, they used the MNIST dataset as a benchmark considering the network size as 64$\times$64, 128$\times$128, and 256$\times$256. The experimental results demonstrated that the speedup of the DLAU accelerator is up to 36.1$\times$ compared with the Intel Core2 processors at 256$\times$256 network size. In addition, the results depict that the proposed architecture is quite energy-efficient as the total power consumption was only 234 mW.

In~\cite{ma2017automatic}, a generalized end-to-end acceleration system of the previously proposed accelerators in~\cite{ma2016scalable, ma2018alamo, ma2017optimizing, ma2017end} has been developed to support diverse CNN models. In doing so, a user-friendly 
interface and an RTL-level compiler have been proposed to automatically generate customized FPGA designs. The authors have developed an expandable optimized RTL-based library containing the most commonly used CNN operations. These operations have been coded in Verilog and designed based on the quantitative analysis and optimization strategies discussed in~\cite{ma2017optimizing}. The compiler generates a DAG-based structure for the used CNN model and then compiles it with RTL modules in the library. The proposed compiler allows the user to input the high-level information of the used CNN model (previously designed on Caffe framework~\cite{jia2014caffe}) as well as the design variables (i.e., loop unrolling and loop tiling variables) with the resource constrains of the targeted FPGA platform. Such utility facilitates the exploration of the best trade-off between the resource usage and the performance.

\Figure[b](topskip=0pt, botskip=0pt, midskip=0pt)[width=0.45\textwidth, height = 5.2 cm]{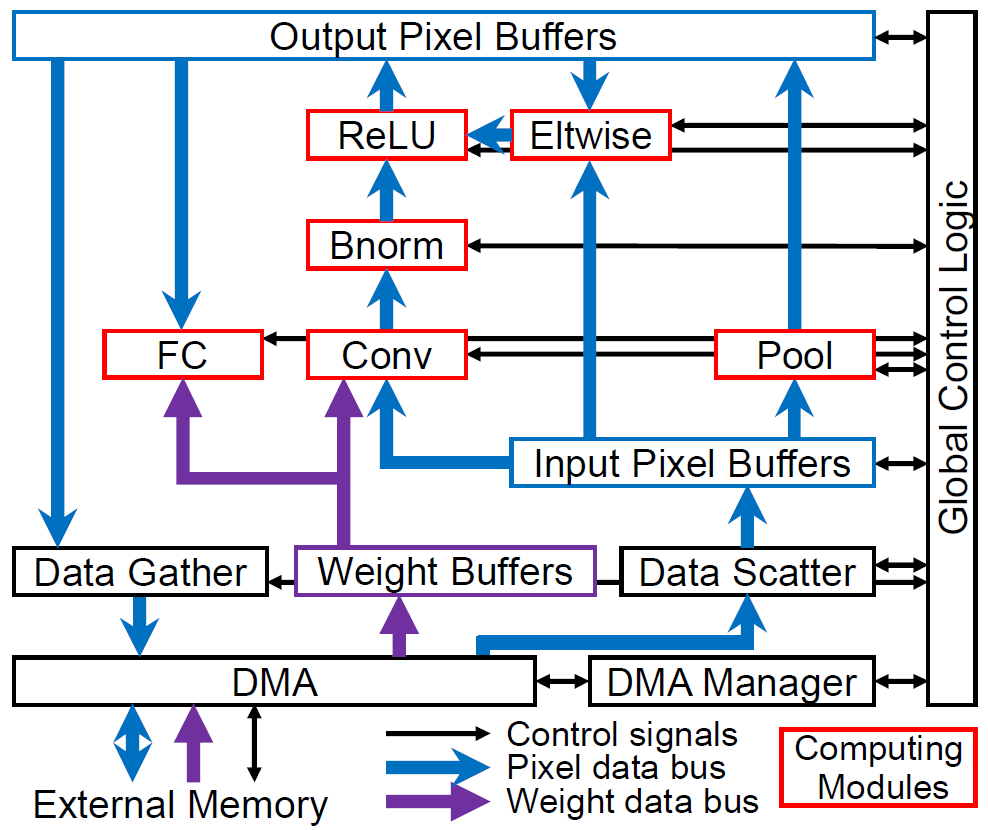}
{Overall Architecture and Dataflow~\cite{ma2017automatic}. 
\label{fig:Ma2017automatic_Arch}}

Unlike the architecture in~\cite{ma2018alamo} where individual CONV module is assigned to each CONV layer, the scalable RTL computing module proposed in this work is reused by all CNN layers of the same type for different CNNs as shown in~Fig.~\ref{fig:Ma2017automatic_Arch}. Note that it is not necessary to have all these modules in the architecture. For instance, the RTL compiler will not compile or synthesize Eltwise and combined batch normalization with scale (Bnorm) modules for VGG-16 model which greatly saves the hardware resources.

On the other hand, the authors categorized CNN layers into key layers (i.e., CONV, pool, and FC layers) and affiliated layers (e.g., ReLU, Bnorm, Eltwise, and all other layers). They have also defined layer combos, where each combo is composed of a key layer and several affiliated layers. Layer combos are sequentially executed according to their order in the DAG. Moreover, the layer combo is also divided into several sequential tiles. The computation of each combo tile starts by reading its input pixels from off-chip DRAM and ends by writing back its output pixels to off-chip DRAM. The global control logic, inter-tile control logic, and intra-tile control logic are responsible for governing the sequential operations of layer combos and reconfiguring the modules, combo tiles, and tile layers (key and affiliated layers), respectively, through predefined flexible execution schedule similar to that in~\cite{ma2017end}.

\Figure[t](topskip=0pt, botskip=0pt, midskip=0pt)[width=0.45\textwidth, height = 5.2 cm]{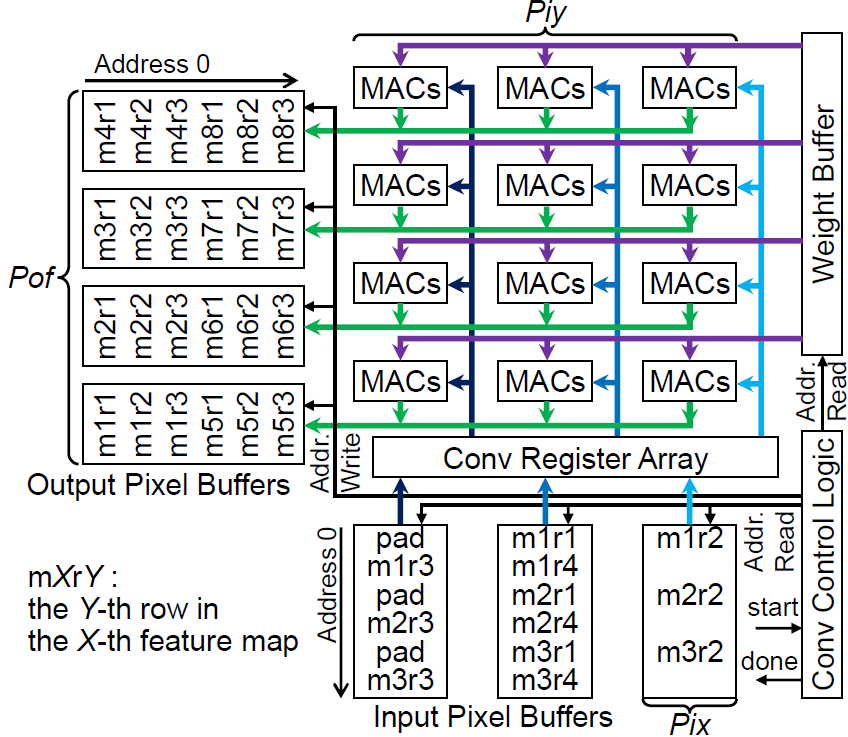}
{CONV Reconfigurable Computing Module~\cite{ma2017automatic}.
\label{fig:Ma2017automatic_CONV}}

The authors have also employed special storage pattern of both input pixels and weights on off-chip memory before the acceleration process to maximize data reuse and minimize of data communication. The architecture of CONV module is designed based on the acceleration strategies in~\cite{ma2017optimizing, ma2017end} but with a different organization of MAC units as shown in Fig.~\ref{fig:Ma2017automatic_CONV}. The MAC units of CONV module have been organized into $Piy \times Pof$ independent MAC blocks, with each MAC block containing $Pix$ MAC units to further minimize the buffer read operations and the partial sums movements. Moreover, such organization enables to handle varying (kernel, stride) sizes configurations through generating different variants of CONV register arrays during the compilation.

Experimental results demonstrated that the achieved throughput on Intel Stratix V GXA7 for NiN, VGG-16, ResNet-50, and ResNet-152 are 282.67 GOPS, 352.24 GOPS, 250.75 GOPS, and 278.67 GOPS, respectively. On the other hand, the achieved throughput on Intel Arria 10 GX 1150 was 587.63 GOPS, 720.15 GOPS, 619.13 GOPS, and 710.30 GOPS for NiN, VGG-16, ResNet-50, and ResNet-152, respectively. More than 2$\times$ throughput improvements have been achieved on Intel Arria 10 GX 1150 since it has 1.8$\times$ and 5.9$\times$ more logic elements and DSPs than the Intel Stratix V GXA7, respectively, which allows for larger loop unrolling variables.

Recently, the programmable solutions group at Intel has developed an FPGA software-programmable and run-time reconfigurable overlay for deep learning inference~\cite{abdelfattah2018dla}. The developed overlay is referred to as the deep learning accelerator (DLA). For the hardware side of Intel's DLA, the team has partitioned the configurable parameters into run-time and compile-time parameters. The run-time parameters allow for easy and quick use of different neural network frameworks, while the compile-time parameters provide a tunable architecture for performance. Intel's DLA uses a lightweight very long instruction word (VLIW)  network, an 8-bit unidirectional ring network, to support the control and reprogramming logic. Compared with typical overlays, Intel's DLA comes with only $\sim$1\% overhead while other  typical overlays tend to always come with larger overheads~\cite{jain2015efficient}. The reprogramming of Intel's DLA overlay allows for consecutive runs of multiple NNs in a single application run~\cite{liu2016ssd} without the need for reconfiguring and recompiling the FPGA.

\Figure[b](topskip=0pt, botskip=0pt, midskip=0pt)[width=0.48\textwidth]{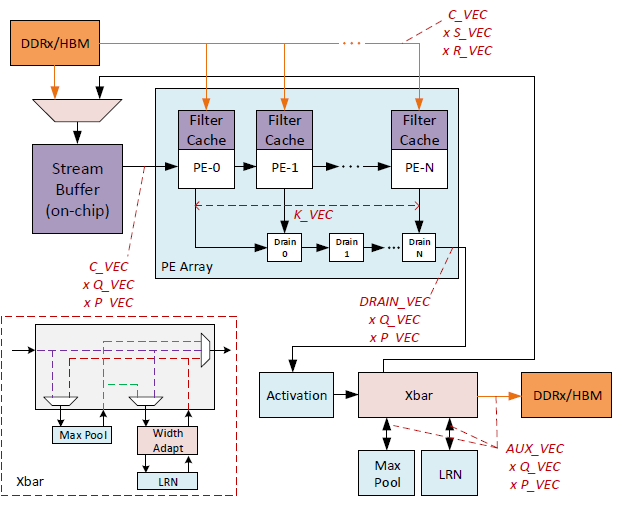}
{Intel's DLA: Neural Network Inference Accelerator~\cite{abdelfattah2018dla}.
\label{DLA-Accelerator}}

Fig.~\ref{DLA-Accelerator} shows that a 1D array of PEs is used to perform convolution, multiplication, or any other matrix operations. Each PE contains a double-buffered filter cache allowing for pre-loading of next filters while computing. The stream buffer employed the double-buffering mechanism as well to store the inputs and the intermediate data on-chip. To have flexible NN architecture, Intel's DLA employs an Xbar interconnect that connects all the core functions required. Thus, deep learning functions can be easily added to the overlay through the Xbar by picking them from a suite of pre-optimized functions of the select frameworks that Intel's DLA uses. The width adaptation module has been used to control the throughput of the function. In addition, Intel's DLA supports vectorization across the input width (Q\_VEC), input height (P\_VEC), input depth (C\_VEC), output depth (K\_VEC), filter width (S\_VEC), and other dimensions as depicted in Fig.~\ref{DLA-Accelerator}. The authors mention that vectorization depends on the layers' dimensions of the considered framework. However, they did not provide a systematic way for finding the optimal balance for the number of used PEs and the size of the caches. For efficient use of resources, Intel's DLA maps AVG pooling and FC primitives to convolutions in order to avoid having under-utilized (over time) dedicated auxiliary functions.

For the software side of Intel's DLA, the proposed accelerator uses a graph compiler to map a NN architecture to the overlay for maximizing the hardware efficiency through slicing, allocation, and scheduling. In the slicing pass, the graph compiler breaks down the architecture into subgraph (a chain of functions) in such a way that they fit within the computing and storage resources of the overlay. A single CONV layer followed by a pooling layer is an example of CNN subgraph. The graph compiler optimizes the external memory spill-points by group slicing technique. The group slicing allows several sequential convolutions, for instance, of a single slice to be computed before moving onto the next slice while using the whole stream buffer. During the allocation pass, the graph compiler optimizes the use of a custom developed filter caches and stream buffer by managing the read and write from the stream buffer for each slice. Moreover, it assigns an external memory address when the stream buffer is not big enough to hold the slice data.

\Figure[b](topskip=0pt, botskip=0pt, midskip=0pt)[height=0.43\textwidth ]{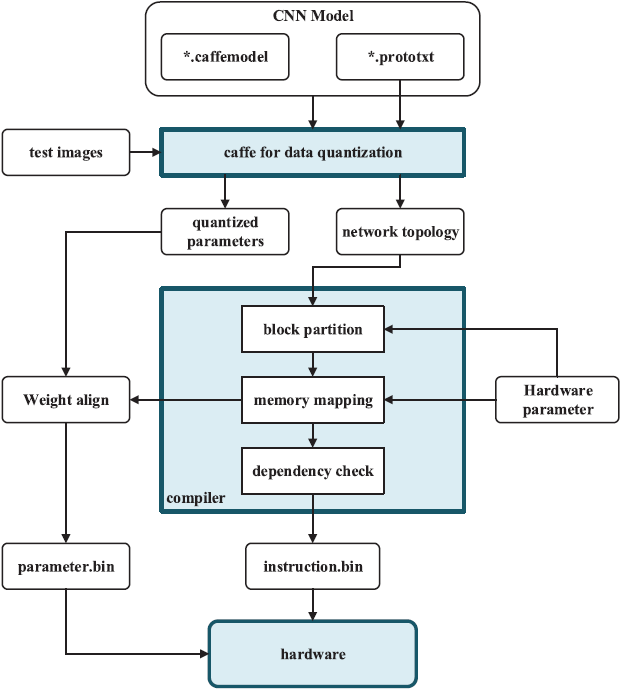}
{Design Flow from CNN Model to Hardware Acceleration~\cite{guo2018angel}.
\label{fig:Angel-Eye-Flow}}

Finally, Intel's DLA compiler schedules the execution of subgraphs using cost-based (the ratio of the output size to the effective input size) priority queue. The authors utilized the software-programmable and run-time reconfigurable overlay to optimize the software and hardware implementation of GoogleNet~\cite{szegedy2015going} and ResNet~\cite{he2016deep} CNNs. The benchmark results on an Arria 10 GX 1150 FPGA demonstrated that Intel's DLA has a throughput of 900 fps on GoogLeNet. The team pointed that multi-FPGA deployment~\cite{chung2017accelerating} might be used to further improve the throughput of Intel's DLA.

Kaiyuan et al. \cite{guo2018angel} proposed a complete design flow, referred to as Angel-Eye, for mapping CNNs onto FPGA.
It includes a data quantization strategy, a parameterizable and run-time configurable hardware architecture to support various CNNs,  FPGA platforms, and a compiler to map a given CNN onto the hardware architecture. It adopts the approach of using a flexible
hardware architecture and maps different CNNs onto it by changing the software.
The proposed design flow from CNN model to hardware acceleration is shown in Fig.~\ref{fig:Angel-Eye-Flow}.

\Figure[t](topskip=0pt, botskip=0pt, midskip=0pt)[width=0.45\textwidth]{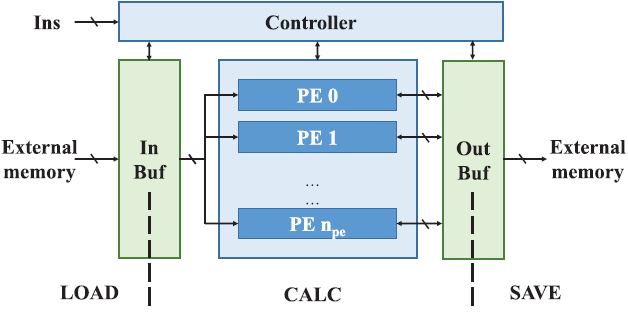}
{Overall Architecture of Angel-Eye~\cite{guo2018angel}.
\label{fig:Angel-Eye-Arch}}

\Figure[b](topskip=0pt, botskip=0pt, midskip=0pt)[width=0.45\textwidth]{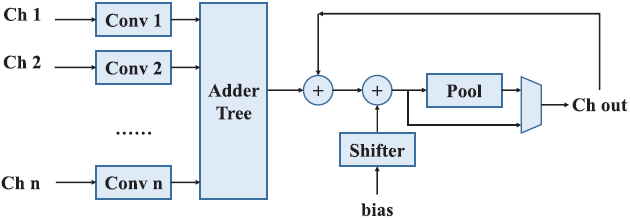}
{Structure of a Single PE~\cite{guo2018angel}.
\label{fig:Angel-Eye-PE}}

Due to the large dynamic range of data across different layers, the best radix point is found 
for each layer for a given bit width. They demonstrated that their strategy can simplify state-of-the-art CNNs to 8-bit fixed-point format with negligible accuracy loss. Although 8-bits are used for representing data, 
24 bits are used for representing intermediate data in layers,  which is then aligned and quantized to 8 bits.
Fig.~\ref{fig:Angel-Eye-Arch} and Fig.~\ref{fig:Angel-Eye-PE} show the overall architecture of Angel-Eye and the structure of a single PE, respectively. The architecture is designed for supporting an instruction interface that supports three types of 
instructions; LOAD, SAVE, and CALC.

The overall architecture is divided into four main components; PE array, on-chip buffer, external memory, and controller. The PE array implements the convolution operations in CNN and supports
kernel level parallelism, and input and output channel parallelisms. It uses a $3\times 3$ convolution
kernel,  as this size is   most popular in state-of-the-art CNN models. However, larger kernel sizes are supported based
on the use of multiple  $3 \times 3$ kernels. The use of on-chip buffers allows data I/O and convolution calculation
to be done in parallel. The controller is responsible for receiving the instructions and issuing them 
to the other components. 
The compiler maps the CNN descriptor to the set of instructions that will be executed by the hardware.
It follows basic scheduling rules to fully utilize  data localization
in CNN and reduce data I/O.

The block partition step partitions the calculation of one layer to fit
each block into the hardware.
The memory mapping step allocates memory for communication between the host CPU and the CNN accelerator. 
Based on the block partition result, on-chip memory is  allocated for the input and output feature
map blocks and  for the convolution kernels and bias values. The dependency check step checks data dependency among instructions and sets appropriate instruction flags to maximize parallelism
between convolution calculation and data I/O.
Based on experimental results, it is shown that  the 8-bit
implementation of Angel-Eye on XC7Z020 achieves up to 16$\times$ higher energy efficiency
than NVIDIA TK1 and 10$\times$ higher than NVIDIA TX1.
In addition, the 16-bit implementation of Angel-Eye on
XC7Z045 is 6$\times$ faster and 5$\times$ higher in power efficiency than
peer FPGA implementation on the same platform \cite{gokhale2014240}. 

\Figure[b](topskip=0pt, botskip=0pt, midskip=0pt)[width=0.48\textwidth, height = 5.4 cm]{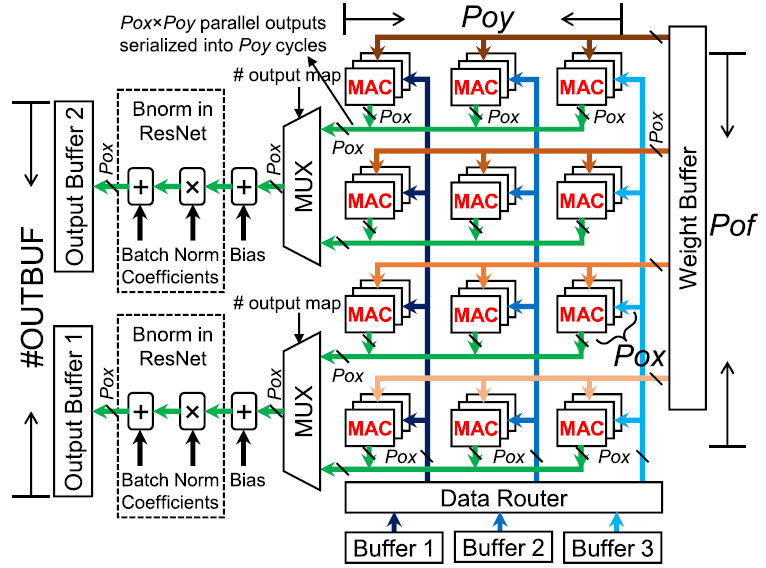}
{CONV Acceleration Architecture and Dataflow using Data Router~\cite{ma2018optimizing}, where, $Pix=Pox$, and $Piy=Poy$.
\label{Optimizing_CONV_Arch}}

In \cite{ma2017optimizing} and \cite{ma2017automatic}, a special register array architecture has been designed to rearrange buffers data and direct them into PEs for the purpose of implementing CONV module that supports specific stride and zero-padding settings. Although the designed CONV module is not generalized for any (kernel, stride) size configurations, it is composed of complex wire routing and control logic as shown in Fig.~\ref{CONV_Loops_Arch}. To have flexibility in directing the dataflow of CONV pixels, Ma et al.~\cite{ma2018optimizing} replaced the register array architecture in \cite{ma2017automatic} with a data router as shown in Fig.~\ref{Optimizing_CONV_Arch}.

The data router is a scalable set of data bus from buffer to PE (BUF2PE). The BUF2PE data bus consists of simple register arrays with FIFOs in between to form a line buffer similar to that in~\cite{bosi1999reconfigurable}. The register array uses the FIFO to pass its input pixels to the adjacent registers. Each BUF2PE data bus has different data movements within its register arrays to implement specific stride and kernel size settings. Unlike the register array architecture in \cite{ma2017optimizing} where the west zero-paddings are handled by changing the storage pattern within the input pixel buffer, the BUF2PE handles such kind of paddings by shifting the connection between the register arrays and the input pixel buffers to simplify the data transferring from off-chip memory to on-chip buffers. However, there is still a need for adjusting the storage pattern within the input buffers in order to handle other zero-paddings.

The global control logic is responsible for selecting the suitable BUF2PE data bus from the data router as well as the suitable storage pattern within the input buffers based on the (kernel, stride) size configuration of CONV layer. The CONV module has also been optimized by reducing the required number of parallel adders that add the partial sums with biases as well as the number of parallel multipliers and adders needed to perform Bnorm operation by serializing the parallel outputs using multipliers. In addition, 16-bit fixed-point has been used to represent both weights and pixels, while dynamically adjusting the decimal point in different layers to fully utilize the existing data width~\cite{guo2018angel}. The proposed compiler in~\cite{ma2017automatic} has been used to configure the parameterized Verilog scripts of the overall CNN acceleration system.   Experimental results show throughput degradation on both Intel Arria 10 GX 1150 and Intel Stratix V GXA7 in comparison to   the results in~\cite{ma2017automatic}.

In Table~\ref{tab_comparison_1} and Table~\ref{tab_comparison_2}, we summarize  the reviewed FPGA-based deep learning networks acceleration techniques. For each technique, we list the year the technique was introduced, the key features employed for acceleration, the used deep learning model, the number of needed operations per image, the FPGA platform used to implement the technique, the precision used for the FMs and weights, the clock frequency used, the design entry for describing the modeled deep learning network, the type of LUT for the used platform, the number of resources available by the used platform in terms of BRAMs, LUTs, FFs, and DSPs, the percentage of each resource utilization, the performance in GOPS, the speedup in comparison to a given baseline model, and finally the power efficiency (GOPS/W).

\begin{sidewaystable*}[ph!]
	\renewcommand{\arraystretch}{1}
	\centering
	\captionsetup{justification=centering}
	\caption{Implementation and Performance Summary of FPGA-based Accelerators.}
	\resizebox{\textwidth}{!}{%
	\begin{tabular}{|l||c|c?{0.8mm}c|c?{0.8mm}c|c|c|c|c|c|c|c|c?{0.8mm}c|c|c|c|c|c|c|c|}
		\hline
		\multirow{3}{*}{\textbf{Technique}} & 
		\multirow{3}{*}{\textbf{Year}} & 
		\multicolumn{1}{c|}{\multirow{3}{*}{\textbf{Key Features}}} & 
		\multirow{3}{*}{\textbf{DL Model}} & 
		\multicolumn{1}{c|}{\multirow{3}{*}{\textbf{\makecell{Image Operations\\ (GOP)}}}} & 
		\multirow{3}{*}{\textbf{Platform}} & 
		\multirow{3}{*}{\textbf{Precision}} & 
		\multirow{3}{*}{\textbf{\makecell{Frequency \\ (MHz)}}} & 
		\multirow{3}{*}{\textbf{LUT Type}} & 
		\multirow{3}{*}{\textbf{Design Entry}} & 
		\multicolumn{4}{c|}{\textbf{Resources}} &
		\multicolumn{4}{c|}{\textbf{Resources Utilization}} & 
		\multirow{3}{*}{\textbf{\makecell{Performance \\ (GOPS)}}} & 
		\multirow{3}{*}{\textbf{Speedup}} & 
		\multirow{3}{*}{\textbf{Baseline}} & 
		\multirow{3}{*}{\textbf{\makecell {Power\\Efficiency\\(GOPS/W)}}}\\
		\cline{11-18}
		&  & \multicolumn{1}{c|}{} &  & \multicolumn{1}{c|}{} &  &  &  &  & & \makecell{BRAMs/\\M20K} & \makecell{LUTs/\\ALMs} & FFs & \multicolumn{1}{c|}{DSPs} & \makecell{BRAMs/\\M20K} & \makecell{LUTs/\\ALMs} & FFs & DSPs & & & &\\
		\hline
		\hline
		\textbf{CNP \cite{farabet2009cnp}} & 2009 & \makecell{2D CONV modules,\\Memory interface with\\ 8 simultaneous accesses} & LeNet-5 & 0.52 & \makecell{Virtex4 \\ SX35} & \makecell{16-bit \\ fixed-point} & 200 & 4-input LUTs & \makecell{Lush} & 192 & 30,720 & 30,720 & 192 & N/A & 90\% & 90\% & 28\% & \textbf{5.25} & \multicolumn{2}{c|}{N/A} & \textbf{0.35}\\ 
		\hline
		\textbf{\makecell[l]{CONV Coprocessor\\ Accelerator \cite{sankaradas2009massively}}} & 2009 & \makecell{Parallel clusters of\\ 2D convolver units,\\ data quantization,\\ off-chip memory banks} & 4 CONV layers & 1.06 & \makecell{Virtex5 \\ LX330T} & \makecell{16-bit \\ fixed-point} & 115 & 6-input LUTs & \makecell{C} & 324 & 207,360 & 207,360 & 192 & 0.93\% & 17\% & 19.05\% & 55.73\% & \textbf{6.74} & \textbf{6$\times$} & \makecell{2.2 GHz\\ AMD Opteron} & \textbf{0.61}\\
        \hline
        \textbf{MAPLE \cite{cadambi2010programmable}} & 2010 & \makecell{In-memory processing,\\ banked off-chip memories,\\ 2D array of VPEs} & 4 CONV layers & 1.06 & \makecell{Virtex5 \\ SX240T} & \makecell{fixed-point} & 125 & 6-input LUTs & \makecell{C++} & 516 & 149,760 & 149,760 & 1,056 & \multicolumn{4}{c|}{N/A} & \textbf{7} & \textbf{0.5$\times$} & \makecell{1.3 GHz \\ C870 GPU} & N/A\\
		\hline
		\textbf{DC-CNN \cite{chakradhar2010dynamically}} & 2010 & \makecell{Integer factorization\\ to determine the best \\ config for each layer,\\ input and output switches} & 3 CONV layers & 0.52 & \makecell{Virtex5 \\ SX240T} & \makecell{48-bit \\ fixed-point} & 120 & 6-input LUTs & RTL & 516 & 149,760 & 149,760 & 1,056 & \multicolumn{4}{c|}{N/A} & \textbf{16} & \textbf{4.0$\times$ - 6.5$\times$} & \makecell{1.35 GHz \\ C870 GPU} & \textbf{\makecell{1.14}}\\
		\hline
		\textbf{\makecell[l]{NeuFlow~\cite{farabet2011neuflow}}} & 2011 & \makecell{Multiple full-custom\\processing tiles (PTs),\\Pipelining, Fast streaming\\memory interface} & \makecell{4 CONV layers\\in~\cite{grangier2009deep}} & N/A & \makecell{Virtex6\\VLX240T} & \makecell{16-bit\\fixed-point} & 200 & 6-input LUTs & \makecell{HDL} & 416 & 150,720 & 301,440 & 768 & \multicolumn{4}{c|}{N/A} & \textbf{147} & \textbf{133.6$\times$} & \makecell{2.66 GHz\\Core 2 Duo CPU} & \textbf{14.7} \\
		\hline
		\textbf{\makecell[l]{Memory-Centric\\ Accelerator \cite{peemen2013memory}}} & 2013 & \makecell{Flexible off-chip memory\\ hierarchy, Data reuse,\\ Loop transformation} & 4 CONV layers & 5.48 & \makecell{Virtex6 \\ VLX240T} & \makecell{fixed-point} & 150 & 6-input LUTs & \makecell{C} & 416 & 150,720 & 301,440 & 768 & 45.5\% & 1.1\% & N/A & 6\% & \textbf{17} & \textbf{11$\times$} & \makecell{Standard Virtex6 \\ Implementation} & N/A\\
		\hline
		\textbf{nn-X \cite{gokhale2014240}} & 2014 & \makecell{Cascaded pipelining,\\ Multiple stream\\ processing} & 2 CONV layers & 0.55 & \makecell{Zynq \\ XC7Z045} & \makecell{16-bit \\ fixed-point} & 142 & 4-input LUTs & \makecell{Lua} & 545 & 218,600 & 437,200 & 900 & \multicolumn{4}{c|}{N/A} & \textbf{23.18} & \textbf{115$\times$} & \makecell{800 MHz\\ Embedded ARM\\ Cortex-A9 Processors} & \textbf{2.9}\\
		\hline
		\textbf{\makecell[l]{Roofline-based FPGA \\ Accelerator \cite{zhang2015optimizing}}} & 2015 & \makecell{Roofline-based model,\\ Loop transformation,\\ Loop tiling/unrolling} & AlexNet & 1.33 & \makecell{Virtex7 \\ VX485T} & \makecell{32-bit \\ float-point} & 100 & 4-input LUTs & \makecell{C} & 2,060 & 303,600 & 607,200 & 2,800 & 50\% & 61.3\% & 33.87\% & 80\% & \textbf{61.62} & \textbf{17.42$\times$} & \makecell{2.2 GHz\\ Intel Xeon} & \textbf{3.31}\\
		\hline
		\textbf{\makecell[l]{Microsoft Specialized\\ CNN Accelerator \cite{ovtcharov2015accelerating}}} & 2015 & \makecell{Multi-banked buffers,\\ Network on-chip for re-\\distribution of output data,\\ Software configurable} & AlexNet & 1.33 & \makecell{Stratix-V \\ GSMD5} & \makecell{32-bit \\ float-point} & 100 & 4-input LUTs & \makecell{C} & 2,014 & 172,600 & 690,000 & 1,590 & \multicolumn{4}{c|}{N/A} & \textbf{178.63} & \textbf{3$\times$} & \makecell{Roofline-based FPGA \\ Accelerator \cite{zhang2015optimizing}} & \textbf{7.15}\\
		\hline
		\textbf{\makecell[l]{Embedded FPGA \\ Accelerator \cite{qiu2016going}}} & 2015 & \makecell{Data quantization\\ and arrangment,\\ SVD} & VGG-16 & 30.76 & \makecell{Zynq \\ XC7Z045} & \makecell{16-bit \\ fixed-point} & 150 & 4-input LUTs & RTL & 545 & 218,600 & 437,200 & 900 & 86.7\% & 83.5\% & 29.2\% & 89.2\% & \textbf{136.97} & \textbf{1.4$\times$} & \makecell{2.9 GHz\\ Intel Xeon} & \textbf{14.22}\\
		\hline
		\textbf{DeepBurning \cite{wang2016deepburning}} & 2016 & \makecell{NN model compression,\\ Compiler-based library,\\ Automatic partitioning/\\tiling, Function Approx.} & AlexNet & 1.46 & \makecell{Zynq\\ XC7Z045} & \makecell{16-bit\\fixed-point} & 100 & 4-input LUTs & RTL & 545 & 218,600 & 437,200 & 900 & N/A & 17.3\% & 7.65\% & 16\% & \textbf{73} & \textbf{15.88$\times$} & \makecell{2.4 GHz\\ Intel Xeon} & \textbf{\makecell{6.43}}\\
		\hline
		\multirow{4}{*}{\textbf{\makecell[l]{OpenCL-based FPGA \\ Accelerator\cite{suda2016throughput}}}} & \multirow{4}{*}{2016} & \multirow{4}{*}{\makecell{Design space exploration \\ for all CNN layers, \\ Genetic algorithm, \\ Altera OpenCL  SDK}} & \multirow{2}{*}{AlexNet$^{(\rm a)}$} & \multirow{2}{*}{1.46} & \multirow{2}{*}{\makecell{{Stratix-V}$^{(\rm 1)}$ \\ GXA7}} & \multirow{4}{*}{\makecell{(8-16)-bit \\ fixed-point}} & \multirow{4}{*}{120} & \multirow{4}{*}{7-input LUTs} & \multirow{4}{*}{\makecell{OpenCL}} & \multirow{2}{*}{2,560} & \multirow{2}{*}{234,720} & \multirow{2}{*}{938,880} & \multirow{2}{*}{256} & 37\% & 47\% & N/A & 100\% & \textbf{31.8}$^{(\rm a1)}$ & \textbf{4.2$\times$} & \multirow{4}{*}{\makecell{3.3 GHz \\ Intel i5-4590}} & \textbf{1.23}\\
		\cdashline{15-20}
		\cdashline{22-22}
		&  &  &  &  & & & & & & & & & & \multicolumn{4}{c|}{N/A} & \textbf{47.5}$^{(\rm b1)}$ & \textbf{2.2$\times$} &  & N/A\\
		\cdashline{4-6}
		\cdashline{11-20}
		\cdashline{22-22}
		&  &  & \multirow{2}{*}{VGG-16$^{(\rm b)}$} & \multirow{2}{*}{30.9} & \multirow{2}{*}{\makecell{{Stratix-V}$^{(\rm 2)}$ \\ GSD8}} & & & & &\multirow{2}{*}{2,567} & \multirow{2}{*}{262,400} & \multirow{2}{*}{1,050,000} & \multirow{2}{*}{1,963} & 52\% & 46\% & N/A & 37\% & \textbf{72.4}$^{(\rm a2)}$ & \textbf{9.5$\times$} &  & \textbf{3.79}\\
		\cdashline{15-20}
		\cdashline{22-22}
		&  &  &  &  & & & & & & & & & & \multicolumn{4}{c|}{N/A} & \textbf{117.8}$^{(\rm b2)}$ & \textbf{5.5$\times$} &  & N/A\\
		\hline
		\multirow{4}{*}{\textbf{Caffeine~\cite{zhang2016caffeine, zhang2018caffeine}}} & \multirow{4}{*}{2016} & \multirow{4}{*}{\makecell{Systolic array architecture,\\Loop unrolling, Double\\buffering, Pipelining,\\Roofline model}} & 
		\multirow{2}{*}{AlexNet$^{(\rm a)}$} & \multirow{2}{*}{1.46} & \multirow{2}{*}{\makecell{{Virtex7}$^{(\rm 1)}$\\VX690T}} & \multirow{2}{*}{\makecell{{16-bit}$^{(\rm a)}$\\ fixed-point}} & \multirow{2}{*}{150} & \multirow{4}{*}{6-input LUTs} & \multirow{4}{*}{C++} & \multirow{2}{*}{2,940} & \multirow{2}{*}{433,200} & \multirow{2}{*}{866,400} & \multirow{2}{*}{3,600}& 42\% & 81\% & 36\% & 78\% & \textbf{354}$^{(\rm b1a)}$ & \textbf{9.7$\times$} & \multirow{4}{*}{\makecell{Two-socket server\\each with a 6-core\\Intel CPU\\E5-2609 at 1.9GHz}} & \textbf{\makecell{13.62}}\\
		\cdashline{15-20}
		\cdashline{22-22}
		&  &  &  &  &  &  & & & &  &  &  &  & 36\% & 31\% & 11\% & 38\% & \textbf{266}$^{(\rm b2a)}$ & \textbf{7.3$\times$} & &\textbf{\makecell{10.64}}\\
		\cdashline{4-8}
		\cdashline{11-20}
		\cdashline{22-22}
		&  &  & \multirow{2}{*}{\makecell{VGG-16$^{(\rm b)}$}} & \multirow{2}{*}{31.1}  & \multirow{2}{*}{\makecell{{Xilinx}$^{(\rm 2)}$\\KU060}} & \multirow{2}{*}{\makecell{{8-bit}$^{(\rm b)}$\\fixed-point}} & \multirow{2}{*}{200} &  & & \multirow{2}{*}{2,160} & \multirow{2}{*}{331,680} & \multirow{2}{*}{663,360} & \multirow{2}{*}{2,760} & 36\% & 60\% & 20\% & 4\% & \textbf{1,171.7}$^{(\rm b2b)}$ & \textbf{29$\times$} &  & \textbf{\makecell{45.07}}\\
		\cdashline{15-20}
		\cdashline{22-22}
		&  &  &  &  &  &  & & & &  &  &  &  & \multicolumn{4}{c|}{N/A} & \textbf{165}$^{(\rm a2a)}$ & \textbf{4.2$\times$} & & N/A\\
		\hline
        \multirow{4}{*}{\textbf{fpgaConvNet~\cite{venieris2016fpgaconvnet}}} & \multirow{4}{*}{2016} & \multirow{4}{*}{\makecell{Datapath optimization,\\Synchronous dataflow,\\Partitioning and folding,\\Design space exploration}} & \multirow{2}{*}{LeNet-5} & \multirow{2}{*}{0.0038} & \multirow{4}{*}{\makecell{Zynq\\XC7Z020}} & \multirow{4}{*}{\makecell{16-bit \\ fixed-point}} & \multirow{4}{*}{100} & \multirow{4}{*}{4-input LUTs} & \multirow{4}{*}{HLS} & \multirow{4}{*}{140} & \multirow{4}{*}{53,200} & \multirow{4}{*}{106,400} & \multirow{4}{*}{200} & \multirow{2}{*}{4.4\%} & \multirow{2}{*}{18.2\%} & \multirow{2}{*}{13\%} & \multirow{2}{*}{1.2\%} & \multirow{2}{*}{\textbf{0.48}} & \multirow{2}{*}{\textbf{0.09$\times$}} & \multirow{2}{*}{CNP~\cite{farabet2009cnp}}  & \multirow{2}{*}{N/A}\\
		&  &  &  &  &  &  & & & &  &  &  &  &  &  &  &  &  &  & &\\
		\cdashline{4-5}
		\cdashline{15-22}
		&  &  & \multirow{2}{*}{\makecell{Scene\\Labelling~\cite{cavigelli2015accelerating}}} & \multirow{2}{*}{7.6528} &  & & & &  &  &  & & & \multirow{2}{*}{6.5\%} & \multirow{2}{*}{61\%} & \multirow{2}{*}{36.6\%} & \multirow{2}{*}{71.8\%} & \multirow{2}{*}{\textbf{12.73}} & \multirow{2}{*}{\textbf{0.17$\times$}} & \multirow{2}{*}{Tegra K1 GPU} & \multirow{2}{*}{\textbf{7.27}}\\
        &  &  &  &  &  &  & & & &  &  &  &  &  &  &  &  &  &  & &\\
		\hline
\end{tabular}}
\label{tab_comparison_1}
\end{sidewaystable*}

\begin{sidewaystable*}[ph!]
	\renewcommand{\arraystretch}{1}
	\centering
	\captionsetup{justification=centering}
	\caption{Implementation and Performance Summary of FPGA-based Accelerators.}
	\resizebox{\textwidth}{!}{%
	\begin{tabular}{|l||c|c?{0.8mm}c|c?{0.8mm}c|c|c|c|c|c|c|c|c?{0.8mm}c|c|c|c|c|c|c|c|}
		\hline
		\multirow{3}{*}{\textbf{Technique}} & 
		\multirow{3}{*}{\textbf{Year}} & 
		\multicolumn{1}{c|}{\multirow{3}{*}{\textbf{Key Features}}} & 
		\multirow{3}{*}{\textbf{DL Model}} & 
		\multicolumn{1}{c|}{\multirow{3}{*}{\textbf{\makecell{Image Operations\\ (GOP)}}}} & 
		\multirow{3}{*}{\textbf{Platform}} & 
		\multirow{3}{*}{\textbf{Precision}} & 
		\multirow{3}{*}{\textbf{\makecell{Frequency \\ (MHz)}}} & 
		\multirow{3}{*}{\textbf{LUT Type}} & 
		\multirow{3}{*}{\textbf{Design Entry}} & 
		\multicolumn{4}{c|}{\textbf{Resources}} &
		\multicolumn{4}{c|}{\textbf{Resources Utilization}} & 
		\multirow{3}{*}{\textbf{\makecell{Performance \\ (GOPS)}}} & 
		\multirow{3}{*}{\textbf{Speedup}} & 
		\multirow{3}{*}{\textbf{Baseline}} & 
		\multirow{3}{*}{\textbf{\makecell {Power\\Efficiency\\(GOPS/W)}}}\\
		\cline{11-18}
		&  & \multicolumn{1}{c|}{} &  & \multicolumn{1}{c|}{} &  &  &  &  & & \makecell{BRAMs/\\M20K} & \makecell{LUTs/\\ALMs} & FFs & \multicolumn{1}{c|}{DSPs} & \makecell{BRAMs/\\M20K} & \makecell{LUTs/\\ALMs} & FFs & DSPs & & & &\\
		\hline
		\hline
		\multirow{3}{*}{\textbf{\makecell[l]{Throughput-Optimized\\FPGA Accelerator \cite{liu2017throughput}}}} & \multirow{3}{*}{2017} & \multirow{3}{*}{\makecell{Four-levels of parallelism,\\ Memory-cache mechanism,\\Design space exploration}} & LeNet & 0.04 & \multirow{3}{*}{\makecell{Virtex7\\VX690T}} & \multirow{3}{*}{\makecell{(8-16)-bit \\ fixed-point}} & \multirow{3}{*}{100} & \multirow{3}{*}{6-input LUTs} & \multirow{3}{*}{\makecell{N/A}} & \multirow{3}{*}{1,470} & \multirow{3}{*}{433,200} & \multirow{3}{*}{866,400} & \multirow{3}{*}{3,600} & 32.4\% & 53.8\% & 35.5\% & 80.8\% & \textbf{424.7} & \textbf{14.84$\times$} & \multirow{3}{*}{\makecell{4.0 GHz\\Intel Core\\i7-4790K CPU}} & \textbf{\makecell{16.85}}\\
		\cdashline{4-5}
		\cdashline{15-20}
		\cdashline{22-22}
		&  &  & AlexNet & 1.46 & & & & & & & & & & 69.5\% & 47.7\% & 37.3\% & 79.8\% & \textbf{445.6} & \textbf{6.96$\times$} &  & \textbf{\makecell{17.97}}\\
		\cdashline{4-5}
		\cdashline{15-20}
		\cdashline{22-22}
		&  &  & VGG-S & 5.29 & & & & & & & & & & 88.8\% & 51.7\% & 34.5\% & 81.9\% & \textbf{473.4} & \textbf{4.79$\times$} &  & \textbf{\makecell{18.49}}\\
		\hline
		\multirow{4}{*}{\textbf{FP-DNN~\cite{guan2017fp}}} & \multirow{4}{*}{2017} & \multirow{4}{*}{\makecell{RTL-HLS hybrid compiler,\\Hand-written matrix\\multiply, Quantization,\\Tiling and double buffers}} & \multirow{2}{*}{VGG-19} & \multirow{2}{*}{39.26} & \multirow{4}{*}{\makecell{Stratix-V \\ GSMD5}} & \multirow{4}{*}{\makecell{16-bit \\ fixed-point}} & \multirow{4}{*}{150} & \multirow{4}{*}{4-input LUTs} & \multirow{4}{*}{\makecell{C++\\and\\OpenCL}} & \multirow{4}{*}{2,014} & \multirow{4}{*}{172,600} & \multirow{4}{*}{690,000} & \multirow{4}{*}{1,590} & \multirow{2}{*}{48\%} & \multirow{2}{*}{27\%} & \multirow{2}{*}{N/A} & \multirow{2}{*}{66\%} & \multirow{2}{*}{\textbf{364.36}} & \multirow{2}{*}{\textbf{3.06$\times$}} & \multirow{4}{*}{\makecell{2 Processors each\\2.6 GHz\\Intel Xeon\\8-core E5-2650v2}} & \multirow{2}{*}{\textbf{\makecell{14.57}}}\\
		&  &  &  &  & & & & & & & & & &  &  &  &  &  &  &  & \\
		\cdashline{4-5}
		\cdashline{15-20}
		\cdashline{22-22}
		&  &  & \multirow{2}{*}{ResNet-152} & \multirow{2}{*}{22.62} & & & & & & & & & & \multirow{2}{*}{48\%} & \multirow{2}{*}{27\%} & \multirow{2}{*}{N/A} & \multirow{2}{*}{66\%} & \multirow{2}{*}{\textbf{226.47}} & \multirow{2}{*}{\textbf{1.9$\times$}} &  & \multirow{2}{*}{\textbf{\makecell{9.06}}}\\
		&  &  &  &  & & & & & & & & & &  &  &  &  &  &  &  & \\
		\hline
		\textbf{FINN \cite{umuroglu2017finn}} & 2017 & \makecell{Binarized CNN,\\Pipelining, Automatic\\partitioning and tiling,\\Approximate arithmetic} & CNV & 0.113 & \makecell{Zynq\\ XC7Z045} & \makecell{(1-2)-bit\\precision} & 200 & 4-input LUTs & C++ & 545 & 218,600 & 437,200 & 900 & 34.1\% & 21.2\% & N/A & N/A & \textbf{2,465.5} & \textbf{13.8$\times$} & \makecell{Microsoft Specialized\\ CNN Accelerator \cite{ovtcharov2015accelerating}} & \textbf{\makecell{210.7}}\\
		\hline
		\multirow{4}{*}{\textbf{\makecell[l]{Customized CONV \\ Loops Accelerator \cite{ma2017optimizing}}}} & \multirow{4}{*}{2017} & \multirow{4}{*}{\makecell{Numerical analysis of\\ CONV loops opt.,\\Solution space exploration,\\ Dataflow opt.}} & \multirow{4}{*}{VGG-16} & \multirow{4}{*}{30.95} & \multirow{4}{*}{\makecell{Arria 10\\GX 1150}} & \multirow{4}{*}{\makecell{(8-16)-bit \\ float-point}} & \multirow{4}{*}{150} & \multirow{4}{*}{8-input LUTs} & \multirow{4}{*}{Verilog} & \multirow{4}{*}{2,713} & \multirow{4}{*}{427,200} & \multirow{4}{*}{1,708,800} & \multirow{4}{*}{1,518} & \multirow{4}{*}{70\%} & \multirow{4}{*}{38\%} & \multirow{4}{*}{N/A} & \multirow{4}{*}{100\%} & \multirow{4}{*}{\textbf{645.25}} & \multirow{2}{*}{\textbf{3.2$\times$}} & \multirow{2}{*}{\makecell{Energy-efficient\\CNN~\cite{zhang2016energy}}} & \multirow{4}{*}{\textbf{30.44}}\\
		&  &  &  &  & & & & & & & & & &  &  &  &  &  &  &  & \\
		\cdashline{20-21}
		&  &  &  &  & & & & & & & & & &  &  &  &  &  &  \multirow{2}{*}{\textbf{5.5$\times$}} & \multirow{2}{*}{\makecell{OpenCL-based FPGA\\ {Accelerator \cite{suda2016throughput}}$^{(\rm b2)}$}}  & \\
		&  &  &  &  & & & & & & & & & &  &  &  &  &  &  &  & \\
		\hline
        \multirow{4}{*}{\textbf{\makecell[l]{Latency-Driven Design\\for FPGA-based\\CNNs~\cite{venieris2017latency}}}} & \multirow{4}{*}{2017} & \multirow{4}{*}{\makecell{Synchronous dataflow,\\Weights reloading\\and SDF transformations,\\Design space exploration}} & \multirow{2}{*}{AlexNet} & \multirow{2}{*}{1.3315} & \multirow{4}{*}{\makecell{ Zynq\\XC7Z045}} & \multirow{4}{*}{\makecell{16-bit\\fixed-point}} & \multirow{4}{*}{125} & \multirow{4}{*}{4-input LUTs} & \multirow{4}{*}{HLS} & \multirow{4}{*}{545} & \multirow{4}{*}{218,600} & \multirow{4}{*}{437,200} & \multirow{4}{*}{900} & \multicolumn{4}{c|}{\multirow{4}{*}{N/A}} & \multirow{2}{*}{\textbf{161.98}} & \multirow{2}{*}{\textbf{1.49$\times$}} & \multirow{2}{*}{DeepBurning~\cite{wang2016deepburning}}  & \multirow{4}{*}{N/A}\\
		&  &  &  &  &  &  & & & &  &  &  &  &  \multicolumn{4}{c|}{} &  &  & &\\
		\cdashline{4-5}
		\cdashline{19-21}
		&  &  & \multirow{2}{*}{VGG-16} & \multirow{2}{*}{30.72} &  & & & &  &  &  & & &  \multicolumn{4}{c|}{} & \multirow{2}{*}{\textbf{123.12}} & \multirow{2}{*}{\textbf{0.65$\times$}} & \multirow{2}{*}{\makecell{Embedded FPGA \\ Accelerator \cite{qiu2016going}}} & \\
        &  &  &  &  &  &  & & & &  &  &  &  &  \multicolumn{4}{c|}{} &  &  & &\\
		\hline
		\multirow{4}{*}{\textbf{DLA~\cite{aydonat2017opencl}}} & \multirow{4}{*}{2017} & \multirow{4}{*}{\makecell{Winograd transformations,\\Double stream buffers,\\PEs double cache buffers,\\Daisy-chained PEs}} & \multirow{4}{*}{AlexNet} & \multirow{4}{*}{1.46} & \multirow{4}{*}{\makecell{Arria 10\\GX 1150}} & \multirow{4}{*}{\makecell{Half-precision\\FP16 with\\shared\\exponent}} & \multirow{4}{*}{303} & \multirow{4}{*}{8-input LUTs} & \multirow{4}{*}{\makecell{OpenCL}} & \multirow{4}{*}{2,713} & \multirow{4}{*}{427,200} & \multirow{4}{*}{1,708,800} & \multirow{4}{*}{1,518} & \multirow{4}{*}{92\%} & \multirow{4}{*}{58\%} & \multirow{4}{*}{40\%} & \multirow{4}{*}{97\%} & \multirow{4}{*}{\textbf{1,382}} & \multirow{2}{*}{\textbf{8.4$\times$}} & \multirow{2}{*}{Caffeine \cite{zhang2016caffeine}$^{(\rm a2a)}$} & \multirow{4}{*}{\textbf{\makecell{30.71}}}\\
		&  &  &  &  & & & & & & & & & &  &  &  &  &  &  &  & \\
		\cdashline{20-21}
		&  &  &  & & & & & & & & & & & &  &  &  &  & \multirow{2}{*}{\textbf{19.1$\times$}} & \multirow{2}{*}{\makecell{OpenCL-based FPGA\\ {Accelerator \cite{suda2016throughput}}$^{(\rm a2)}$}} & \\
		&  &  &  &  & & & & & & & & & &  &  &  &  &  &  &  & \\
		\hline
		\multirow{4}{*}{\textbf{\makecell[l]{Winograd-based\\CNN Accelerator~\cite{lu2017evaluating}}}} & \multirow{4}{*}{2017} & \multirow{4}{*}{\makecell{Winograd algorithm, Loop\\unrolling, Double buffers,\\Batching, Line buffers,\\Design space exploration}} & \multirow{2}{*}{{AlexNet}$^{(\rm a)}$} & \multirow{2}{*}{1.33} & \multirow{2}{*}{\makecell{{Zynq}$^{(\rm 1)}$\\ZCU102}} & \multirow{4}{*}{\makecell{16-bit \\ fixed-point}} & \multirow{2}{*}{200} & \multirow{2}{*}{6-input LUTs} & \multirow{4}{*}{\makecell{C}} & \multirow{2}{*}{912} & \multirow{2}{*}{274,080} & \multirow{2}{*}{548,160} & \multirow{2}{*}{2,520} & \multicolumn{4}{c|}{\multirow{4}{*}{N/A}} & \textbf{854.6}$^{(\rm a1)}$ & \textbf{11.8$\times$} & \cite{suda2016throughput}$^{(\rm a2)}$ & \textbf{36.2}\\
		\cdashline{19-22}
		&  &  &  &  & & & & & & & & & & \multicolumn{4}{c|}{} & \textbf{2,940.7}$^{(\rm b1)}$ & \textbf{8.31$\times$} & Caffeine \cite{zhang2016caffeine}$^{(\rm b1a)}$ & \textbf{124.6}\\
		\cdashline{4-6}
		\cdashline{8-9}
		\cdashline{11-14}
		\cdashline{19-22}
		&  &  & \multirow{2}{*}{{VGG-16}$^{(\rm b)}$} & \multirow{2}{*}{30.76} & \multirow{2}{*}{\makecell{{Xilinx}$^{(\rm 2)}$\\ZC706}} & & \multirow{2}{*}{167} & \multirow{2}{*}{4-input LUTs} &  & \multirow{2}{*}{545} & \multirow{2}{*}{218,600} & \multirow{2}{*}{437,200} & \multirow{2}{*}{900} & \multicolumn{4}{c|}{} & \textbf{201.4}$^{(\rm a2)}$ & \textbf{2.78$\times$} & \cite{suda2016throughput}$^{(\rm a2)}$ & \textbf{\makecell{21.4}}\\
		\cdashline{19-22}
		&  &  &  &  & & & & & & & & & & \multicolumn{4}{c|}{} & \textbf{679.6}$^{(\rm b2)}$ & \textbf{0.13$\times$} & Titan X (CuDNN 5.1) & \textbf{72.3}\\
		\hline
		\multirow{4}{*}{\textbf{\makecell[l]{OpenCL-based\\Architecture for\\Accelerating\\CNNs~\cite{zhang2017improving}}}} & \multirow{4}{*}{2017} & \multirow{4}{*}{\makecell{2D BRAM-to-PE\\interconnection, 2D\\dispatcher, Roofline\\model, OpenCL}} & 
		\multirow{4}{*}{\makecell{VGG-16}} & \multirow{4}{*}{30.76} & \multirow{4}{*}{\makecell{Arria 10\\GX 1150}} & \multirow{2}{*}{\makecell{32-bit\\float-point}} & \multirow{2}{*}{370} & \multirow{4}{*}{8-input LUTs} & \multirow{4}{*}{OpenCL} & \multirow{2}{*}{2,713} & \multirow{2}{*}{427,200} & \multirow{2}{*}{1,708,800} & \multirow{2}{*}{1,518} & \multirow{2}{*}{46.1\%} & \multirow{2}{*}{N/A} & \multirow{2}{*}{N/A} & \multirow{2}{*}{87\%} & \multirow{2}{*}{\textbf{866}} & \multirow{2}{*}{\textbf{4.41$\times$}} & \multirow{2}{*}{\makecell{Altera OpenCL \cite{czajkowski2012opencl2}\\on Arria 10 Platform}} & \multirow{2}{*}{\textbf{\makecell{20.75}}}\\
		&  &  &  &  &  &  & & & &  &  &  &  &  &  &  &  &  &  & &\\
		\cdashline{7-8}
		\cdashline{11-22}
		&  &  & &   &  & \multirow{2}{*}{\makecell{16-bit\\fixed-point}} & \multirow{2}{*}{385} &  & & \multirow{2}{*}{2,713} & \multirow{2}{*}{427,200} & \multirow{2}{*}{1,708,800} & \multirow{2}{*}{3,036} & \multirow{2}{*}{53.4\%} & \multirow{2}{*}{N/A} & \multirow{2}{*}{N/A} & \multirow{2}{*}{90.8\%} & \multirow{2}{*}{\textbf{1,790}} & \multicolumn{2}{c|}{\multirow{2}{*}{N/A}} & \multirow{2}{*}{\textbf{\makecell{47.78}}}\\
		&  &  &  &  &  &  & & & &  &  &  &  &  &  &  &  &  & \multicolumn{2}{c|}{} &\\
		\hline
		\multirow{4}{*}{\textbf{\makecell[l]{Multi-CLP\\Accelerator for\\CNNs~\cite{shen2017maximizing}}}} & \multirow{4}{*}{2017} & \multirow{4}{*}{\makecell{Multiple CONV\\processors, Pipelining,\\Dynamic programming,\\Double-buffering}} & \multirow{2}{*}{{AlexNet}$^{(\rm a)}$} & \multirow{2}{*}{1.33} & \multirow{2}{*}{\makecell{{Virtex7}$^{(\rm 1)}$\\VX485T}} & \multirow{2}{*}{\makecell{32-bit \\ float-point}} & \multirow{2}{*}{100} & \multirow{2}{*}{4-input LUTs} & \multirow{4}{*}{\makecell{C++}} & \multirow{2}{*}{2,060} & \multirow{2}{*}{303,600} & \multirow{2}{*}{607,200} & \multirow{2}{*}{2,800} & 39.4\% & 58.3\% & 44.6\% & 87.3\% & \textbf{85.2}$^{(\rm a1)}$ & \textbf{1.31$\times$} & \multirow{4}{*}{\makecell{Single CLP\\Design Based\\in~\cite{zhang2015optimizing}}} & \textbf{11.2}\\
		\cdashline{15-20}
		\cdashline{22-22}
		&  &  &  &  & & & & & & & & & & 48.8\% & 84.7\% & 40.2\% & 88.3\% & \textbf{113.92}$^{(\rm a2)}$ & \textbf{1.54$\times$} &  & \textbf{11.17}\\
		\cdashline{4-7}
		\cdashline{8-9}
		\cdashline{11-20}
		\cdashline{22-22}
		&  &  & \multirow{2}{*}{{SqueezeNet}$^{(\rm b)}$} & \multirow{2}{*}{0.77} & \multirow{2}{*}{\makecell{{Virtex7}$^{(\rm 2)}$\\VX690T}} & \multirow{2}{*}{\makecell{16-bit \\ fixed-point}} & \multirow{2}{*}{170} & \multirow{2}{*}{6-input LUTs} &  & \multirow{2}{*}{2,940} & \multirow{2}{*}{433,200} & \multirow{2}{*}{866,400} & \multirow{2}{*}{3,600} & \multicolumn{4}{c|}{N/A} & \textbf{708.3}$^{(\rm b1)}$ & \textbf{1.9$\times$} &  & N/A\\
		\cdashline{15-20}
		\cdashline{22-22}
		&  &  &  &  & & & & & & & & & & 37.7\% & 30.9\% & 18.6\% & 97.1\% & \textbf{909.7}$^{(\rm b2)}$ & \textbf{2.3$\times$} &  & \textbf{126.3}\\
		\hline
		\multirow{4}{*}{\textbf{\makecell[l]{Automated Systolic\\Array Architecture\\for CNN~\cite{wei2017automated}}}} & \multirow{4}{*}{2017} & \multirow{4}{*}{\makecell{2D systolic array\\architecture, Roofline\\model, Automation flow\\Design space exploration}} & 
		\multirow{2}{*}{\makecell{{AlexNet}$^{(\rm a)}$}} & \multirow{2}{*}{1.4} & \multirow{4}{*}{\makecell{Arria 10\\GX 1150}} & \multirow{2}{*}{\makecell{{32-bit}$^{(\rm 1)}$\\float-point}} & \multirow{2}{*}{{239.62}$^{(\rm a1)}$} & \multirow{4}{*}{8-input LUTs} & \multirow{4}{*}{OpenCL} & \multirow{4}{*}{2,713} & \multirow{4}{*}{427,200} & \multirow{4}{*}{1,708,800} & \multirow{4}{*}{1,518} & \multirow{2}{*}{87\%}  & \multirow{2}{*}{82\%}  & \multirow{2}{*}{N/A}  & \multirow{2}{*}{85\%}  & \multirow{2}{*}{\textbf{360.4}$^{(\rm a1)}$} &  \multicolumn{2}{c|}{\multirow{4}{*}{N/A}} & \multirow{2}{*}{\textbf{\makecell{20.75}}}\\
		&  &  &  &  &  &  & & & &  &  &  &  &  &  &  &  &  &  \multicolumn{2}{c|}{} & \multirow{4}{*}{N/A} \\
		\cdashline{4-5}
		\cdashline{7-8}
		\cdashline{15-19}
		\cdashline{22-22}
		&  &  & \multirow{2}{*}{\makecell{{VGG-16}$^{(\rm b)}$}} & \multirow{2}{*}{31.1} &  & \multirow{2}{*}{\makecell{{(8-16)-bit}$^{(\rm 2)}$\\fixed-point}} & {221.65}$^{(\rm b1)}$ & & & &  &  &  & 90.5\%  & 83\%  & N/A  & 88.3\%  & \textbf{460.5}$^{(\rm b1)}$ &  \multicolumn{2}{c|}{} & \\
		\cdashline{8-8}
		\cdashline{15-19}
		&  &  & &   &  &  & {231.85}$^{(\rm b2)}$ &  & &  &  &  &  & 61.5\% & 73\%  & N/A  & 98.8\%  & \textbf{1,171.3}$^{(\rm b2)}$ & \multicolumn{2}{c|}{} & \\
		\hline
		\multirow{6}{*}{\textbf{\makecell[l]{End-to-End Scalable \\ FPGA Accelerator \cite{ma2017end}}}} & \multirow{6}{*}{2017} & \multirow{6}{*}{\makecell{Flexible and scalable\\ResNet modules, CONV\\loops opt., Dataflow opt.,\\Controlled execution\\flow of ResNets layers}} & \multirow{3}{*}{ResNet-50} & \multirow{3}{*}{7.74} & \multirow{6}{*}{\makecell{Arria 10\\GX 1150}} & \multirow{6}{*}{\makecell{16-bit \\ fixed-point}} & \multirow{6}{*}{150} & \multirow{6}{*}{8-input LUTs} & \multirow{6}{*}{Verilog} & \multirow{6}{*}{2,713} & \multirow{6}{*}{427,200} & \multirow{6}{*}{1,708,800} & \multirow{6}{*}{1,518} & \multirow{3}{*}{80\%} & \multirow{3}{*}{30\%} & \multirow{3}{*}{N/A} & \multirow{3}{*}{69\%} & \multirow{3}{*}{\textbf{285.07}} & \multicolumn{2}{c|}{\multirow{6}{*}{N/A}} & \multirow{6}{*}{N/A}\\
		&  &  &  &  & & & & & & & & & &  &  &  &  &  &  \multicolumn{2}{c|}{} & \\
		&  &  &  &  & & & & & & & & & &  &  &  &  &  &  \multicolumn{2}{c|}{} & \\
		\cdashline{4-5}
		\cdashline{15-19}
		&  &  & \multirow{3}{*}{ResNet-152} & \multirow{3}{*}{22.62} & & & & & & & & & & \multirow{3}{*}{93\%} & \multirow{3}{*}{33\%} & \multirow{3}{*}{N/A} & \multirow{3}{*}{69\%} & \multirow{3}{*}{\textbf{315.48}} &  \multicolumn{2}{c|}{}  & \\
		&  &  &  &  & & & & & & & & & &  &  &  &  &  &  \multicolumn{2}{c|}{}  & \\
		&  &  &  &  & & & & & & & & & &  &  &  &  &  &  \multicolumn{2}{c|}{} & \\
		\hline
		\textbf{DLAU \cite{wang2017dlau}} & 2017 & \makecell{Pipelined processing\\ units, Tiling,\\ FIFO buffers} & DNN & N/A & \makecell{Zynq \\ XC7Z020} & \makecell{48-bit \\ float-point} & 200 & 4-input LUTs & N/A & 280 & 53,200 & 106,400 & 220 & 12.5\% & 68.4\% & 26.6\% & 75.9\% & N/A & \textbf{36.1$\times$} & \makecell{2.3 GHz\\ Intel Core2} & N/A\\
		\hline
		\multirow{8}{*}{\textbf{\makecell[l]{An Automatic RTL\\Compiler for High-\\Throughput Deep\\CNNs\cite{ma2017automatic}}}} & 
		\multirow{8}{*}{2017} & 
		\multirow{8}{*}{\makecell{Library-based RTL\\compiler, Flexible and\\scalable CNN modules,\\Layer combo computation,\\CONV loops and dataflow\\opt., Controlled execution\\flow of CNN layers}} &
		\multirow{2}{*}{{NiN}$^{(\rm a)}$} &
		\multirow{2}{*}{2.2} &
		\multirow{4}{*}{\makecell{{Stratix-V}$^{(\rm 1)}$ \\ GXA7}} & 
		\multirow{8}{*}{\makecell{16-bit \\ fixed-point}} & 
		\multirow{4}{*}{150} & 
		\multirow{4}{*}{7-input LUTs} & 
		\multirow{8}{*}{Verilog} &
		\multirow{4}{*}{2,560} & \multirow{4}{*}{234,720} & \multirow{4}{*}{938,880} & \multirow{4}{*}{256} & 
		59\% & 96\% & N/A & 100\% & 
		\textbf{282.67}$^{(\rm a1)}$ &
		\textbf{$>$6.4$\times$} &  
		DeepBurning \cite{wang2016deepburning} & 
		\multirow{8}{*}{N/A}\\
		\cdashline{15-21}
		&  &  &  &  &  &  & & & & &  &  & &  86\% & 90\% & N/A & 100\% & \textbf{352.24}$^{(\rm b1)}$ & \multicolumn{2}{c|}{N/A} &\\
		\cdashline{4-5}
		\cdashline{15-21}
		&  &  & \multirow{2}{*}{{VGG-16}$^{(\rm b)}$} & \multirow{2}{*}{30.95} &  &  & & & & &  & &  & 76\% & 74\% & N/A & 100\% & \textbf{250.75}$^{(\rm c1)}$ & \multicolumn{2}{c|}{N/A} &\\
		\cdashline{15-21}
		&  &  &  &  &  &  & & & &  &  &  & &  93\% & 77\% & N/A & 100\% & \textbf{278.67}$^{(\rm d1)}$ &	\textbf{1.23$\times$} & FP-DNN \cite{guan2017fp} &\\
		\cdashline{4-6}
		\cdashline{8-9}
		\cdashline{11-21}
		&  &  & \multirow{2}{*}{{ResNet-50}$^{(\rm c)}$} & \multirow{2}{*}{7.74} & \multirow{4}{*}{\makecell{{Arria 10}$^{(\rm 2)}$\\GX 1150}} &  & \multirow{4}{*}{200} & \multirow{4}{*}{8-input LUTs} & & \multirow{4}{*}{2,713} & \multirow{4}{*}{427,200} & \multirow{4}{*}{1,708,800} & \multirow{4}{*}{1,518} & 56\% & 37\% & N/A & 100\% & \textbf{587.63}$^{(\rm a2)}$ & \multicolumn{2}{c|}{N/A} & \\
		\cdashline{15-21}
		&  &  &  &  &  &  & & & & &  &  & &  82\% & 30\% & N/A & 100\% & \textbf{720.15}$^{(\rm b2)}$ & \textbf{2.0$\times$} & Caffeine \cite{zhang2016caffeine}$^{(\rm b1a)}$ & \\
		\cdashline{4-5}
		\cdashline{15-21}
		&  &  & \multirow{2}{*}{{ResNet-152}$^{(\rm d)}$} & \multirow{2}{*}{22.62} &  &  &  & & & & & &  & 71\% & 51\% & N/A & 100\% & \textbf{619.13}$^{(\rm c2)}$ & \multicolumn{2}{c|}{N/A} & \\
		\cdashline{15-21}
		&  &  &  &  &  &  & & & &  &  &  &  & 87\% & 54\% & N/A & 100\% & \textbf{710.30}$^{(\rm d2)}$ & \multicolumn{2}{c|}{N/A} &\\
		\hline
		\multirow{4}{*}{\textbf{ALAMO \cite{ma2016scalable, ma2018alamo}}} & \multirow{4}{*}{2018} & \multirow{4}{*}{\makecell{Modularized RTL\\ compiler, Loop\\ unrolling, Loop\\ tiling}} & \multirow{2}{*}{AlexNet} & \multirow{2}{*}{1.46} & \multirow{4}{*}{ \makecell{Stratix-V \\ GXA7}} & \multirow{4}{*}{\makecell{(8-16)-bit \\ fixed-point}} & \multirow{4}{*}{100} & \multirow{4}{*}{7-input LUTs} & \multirow{4}{*}{Verilog} & \multirow{4}{*}{2,560} & \multirow{4}{*}{234,720} & \multirow{4}{*}{938,880} & \multirow{4}{*}{256} & \multirow{2}{*}{61\%} & \multirow{2}{*}{52\%} & \multirow{2}{*}{N/A} & \multirow{2}{*}{100\%} & \multirow{2}{*}{\textbf{114.5}} & \multirow{2}{*}{\textbf{1.9$\times$}} & \multirow{2}{*}{\makecell{Roofline-based FPGA \\ Accelerator \cite{zhang2015optimizing}}} & \multirow{2}{*}{\textbf{\makecell{5.87}}}\\
		&  &  &  &  &  &  & & & &  &  &  &  &  &  &  &  &  &  & &\\
		\cdashline{4-5}
		\cdashline{15-22}
		&  &  & \multirow{2}{*}{NiN} & \multirow{2}{*}{2.2} &  & & & &  &  &  & & & \multirow{2}{*}{91\%} & \multirow{2}{*}{48\%} & \multirow{2}{*}{N/A} & \multirow{2}{*}{100\%} & \multirow{2}{*}{\textbf{117.3}} & \multicolumn{2}{c|}{\multirow{2}{*}{N/A}} & \multirow{2}{*}{\textbf{6.14}}\\
		&  &  &  &  &  &  & & & &  &  &  &  &  &  &  &  &  &  \multicolumn{2}{c|}{} &\\
		\hline
		\multirow{4}{*}{\textbf{Angel-Eye \cite{lu2017evaluating}}} & \multirow{4}{*}{2018} & \multirow{4}{*}{\makecell{Quantization,\\Parallel PEs,\\Compiler,\\On-chip buffers}} & 
		\multirow{4}{*}{\makecell{VGG-16}} & \multirow{4}{*}{30.76} & \multirow{2}{*}{\makecell{Zynq\\XC7Z020}} & \multirow{2}{*}{\makecell{8-bit\\fixed-point}} & \multirow{2}{*}{214} & \multirow{4}{*}{4-input LUTs} & \multirow{4}{*}{N/A} & \multirow{2}{*}{140} & \multirow{2}{*}{53,200} & \multirow{2}{*}{106,400} & \multirow{2}{*}{220} & \multirow{2}{*}{61\%} & \multirow{2}{*}{56\%} & \multirow{2}{*}{33\%} & \multirow{2}{*}{86.4\%} & \multirow{2}{*}{\textbf{84.3}} & \multirow{2}{*}{\textbf{3.8$\times$}} & \multirow{4}{*}{\makecell{nn-X \cite{gokhale2014240}}} & \multirow{2}{*}{\textbf{\makecell{24.1}}}\\
		&  &  &  &  &  &  & & & &  &  &  &  &  &  &  &  &  &  & &\\
		\cdashline{6-8}
		\cdashline{11-20}
		\cdashline{22-22}
		&  &  & &   & \multirow{2}{*}{\makecell{Zynq\\XC7Z045}} & \multirow{2}{*}{\makecell{16-bit\\fixed-point}} & \multirow{2}{*}{150} &  & & \multirow{2}{*}{545} & \multirow{2}{*}{218,600} & \multirow{2}{*}{437,200} & \multirow{2}{*}{900} & \multirow{2}{*}{89\%} & \multirow{2}{*}{84\%} & \multirow{2}{*}{29\%} & \multirow{2}{*}{87\%} & \multirow{2}{*}{\textbf{137}} & \multirow{2}{*}{\textbf{6$\times$}} &  & \multirow{2}{*}{\textbf{\makecell{14.2}}}\\
		&  &  &  &  &  &  & & & &  &  &  &  &  &  &  &  &  &  & &\\
		\hline
		\multirow{8}{*}{\textbf{\makecell[l]{Optimizing the CONV\\Operation to Accelerate\\DNNs on FPGA~\cite{ma2018optimizing}}}} & 
		\multirow{8}{*}{2018} & 
		\multirow{8}{*}{\makecell{Scalable set of data buses\\(BUF2PE), Optimized\\CONV module for\\different (kernel, stride)\\size  configurations,\\Flexible and scalable CNN\\modules, CONV loops\\and dataflow opt.}} &
		\multirow{2}{*}{{NiN}$^{(\rm a)}$} &
		\multirow{2}{*}{2.2} &
		\multirow{4}{*}{\makecell{{Stratix-V}$^{(\rm 1)}$ \\ GXA7}} & 
		\multirow{8}{*}{\makecell{16-bit \\ fixed-point}} & 
		\multirow{4}{*}{150} & 
		\multirow{4}{*}{7-input LUTs} & 
		\multirow{8}{*}{Verilog} &
		\multirow{4}{*}{2,560} & \multirow{4}{*}{234,720} & \multirow{4}{*}{938,880} & \multirow{4}{*}{256} & 
		59\% & 97\% & N/A & 100\% & 
		\textbf{278.2}$^{(\rm a1)}$ &
		\multicolumn{2}{c|}{N/A} &
		\multirow{8}{*}{N/A}\\
		\cdashline{15-21}
		&  &  &  &  &  &  & & & & &  &  & &  86\% & 93\% & N/A & 100\% & \textbf{348.8}$^{(\rm b1)}$ & \multicolumn{2}{c|}{N/A} &\\
		\cdashline{4-5}
		\cdashline{15-21}
		&  &  & \multirow{2}{*}{{VGG-16}$^{(\rm b)}$} & \multirow{2}{*}{30.95} &  &  & & & & &  & &  & 76\% & 75\% & N/A & 100\% & \textbf{243.3}$^{(\rm c1)}$ & \multicolumn{2}{c|}{N/A} &\\
		\cdashline{15-21}
		&  &  &  &  &  &  & & & &  &  &  & &  93\% & 78\% & N/A & 100\% & \textbf{276.6}$^{(\rm d1)}$ &	\textbf{1.2$\times$} & FP-DNN \cite{guan2017fp} &\\
		\cdashline{4-6}
		\cdashline{8-9}
		\cdashline{11-21}
		&  &  & \multirow{2}{*}{{ResNet-50}$^{(\rm c)}$} & \multirow{2}{*}{7.74} & \multirow{4}{*}{\makecell{{Arria 10}$^{(\rm 2)}$\\GX 1150}} &  & \multirow{4}{*}{200} & \multirow{4}{*}{8-input LUTs} & & \multirow{4}{*}{2,713} & \multirow{4}{*}{427,200} & \multirow{4}{*}{1,708,800} & \multirow{4}{*}{1,518} & 56\% & 38\% & N/A & 100\% & \textbf{584.8}$^{(\rm a2)}$ & \multicolumn{2}{c|}{N/A} & \\
		\cdashline{15-21}
		&  &  &  &  &  &  & & & & &  &  & &  82\% & 32\% & N/A & 100\% & \textbf{715.9}$^{(\rm b2)}$ & \multicolumn{2}{c|}{N/A} & \\
		\cdashline{4-5}
		\cdashline{15-21}
		&  &  & \multirow{2}{*}{{ResNet-152}$^{(\rm d)}$} & \multirow{2}{*}{22.62} &  &  &  & & & & & &  & 71\% & 52\% & N/A & 100\% & \textbf{611.4}$^{(\rm c2)}$ & \multicolumn{2}{c|}{N/A} & \\
		\cdashline{15-21}
		&  &  &  &  &  &  & & & &  &  &  &  & 87\% & 55\% & N/A & 100\% & \textbf{707.2}$^{(\rm d2)}$ & \multicolumn{2}{c|}{N/A} &\\
		\hline
\end{tabular}}
\label{tab_comparison_2}
\end{sidewaystable*}

\section{Metaheuristics in the Design  of Convolutional Neural Networks} \label{sec:Metaheuristics_in_the_Design_of_Convolutional_Neural_Networks}

Currently, convolutional neural network (CNN) structures are designed  based on human expertise. For a given application, this consists of  determining the number of convolution layers,
number of fully connected layers, sizes of feature maps in each layer, along with other operators. 
Recent research has demonstrated that a large number of weights in fully connected layers could be eliminated with
minimal impact on accuracy. In addition, although the suggested CNN structures by experts perform well for various
applications, the question arises whether the suggested structures could be optimized for performance 
with minimal impact on accuracy. 
Since the designed CNN has a significant impact on the complexity of its implementation, we review in this section
some  approaches attempting to optimize the design of CNNs using metaheuristics.

NP-hard combinatorial optimization problems~\cite{sait1999iterative} appear in  the design of CNNs.  Some examples of areas include design of CNN structures, selection of weights and bias values to improve accuracy, and determination of optimal values of variables to reduce run-time. Below, we briefly touch upon some existing literature in these areas. 

\subsection{CNN Structure Optimization}
In the design of CNNs, the number of possible network structures increases exponentially with the number of layers. Xie and Yuille   used   genetic algorithm in learning deep network structures~\cite{xie2017genetic}.   The objective was  to find the best CNN structure that would minimize the error rate. The cost function was  the CNN accuracy.
They proposed an elegant  encoding of chromosome  using a fixed length binary string to represent each network structure. A CNN string represents only the convolution layers.

In each generation, using standard genetic operations new individuals are generated and weak ones eliminated. The quality of an individual was assessed by its {\it recognition accuracy} which is obtained via the time consuming operation of training the network, and evaluating it on a validation set. Two small data sets were used (MNIST and CIFAR-10) to run the genetic implementation via which they demonstrated the discovery of new structures. 

\subsection{CNN Weights and Bias Values Optimization}
An attempt to train CNNs using metaheuristics (that is, determine weights and bias values) is presented in~\cite{rere2016metaheuristic}. The objective again was to improve accuracy and minimize the estimated error.
The authors experiment with three metaheuristic algorithms, namely; simulated annealing, differential evolution, and harmony search. The algorihtms compute the values of weights and bias in the last layer. These  values are used as the solution vector denoted by $x$ which is to be optimized. The move comprised adding a small value of $\Delta x$ to perturb the state. The cost function  $y $  is modeled as

\begin{equation}
 y =\frac{1}{2} \bigg(\frac{\sum_{i=n}^{N}(o-u)^2 }{N }\bigg)^{0.5}
\end{equation}
\\
where, \textit{o} is the expected output, \textit{u} is the real output, and \textit{N} is the number of used samples. The stopping criterion is when the  iteration count is reached  or 
when  the cost function goes below a pre-specified value.  

\subsection{CNN Design Variables Optimization} 
Suda et al.~\cite{suda2016throughput} presented  a systematic methodology for design space exploration with the objective of maximizing the throughput of an OpenCL-based FPGA accelerator for a given CNN model (please see subsection~\ref{FPGAs-based Accelerators}). FPGA resource constraints such as on-chip memory, registers, computational resources and external memory bandwidth are considered. 
The optimization problem comprises finding the best combination of $N_{CONV}$, $S_{CONV}$, $N_{NORM}$, $N_{POOL}$, and $N_{FC}$ variables, where 

\begin{itemize}
    \item 

$N_{CONV}$ is  size of the filter (or neuron or kernel); 

 \item
$S_{CONV}$ is the factor by which computational resources are vectorized to execute in a single-instruction stream multiple-data streams (SIMD) fashion; 
\item $N_{NORM}$ represents the number of normalization operations performed in a
single cycle;
\item $N_{POOL}$ is the number of parallel outputs of the pooling layer in a single cycle to achieve acceleration; and, 

\item $N_{FC}$ is the number of parallel multiply and accumulate (MAC) operations preformed in a single work-item within the fully connected layer. 
\end{itemize}

The objective function to be minimized is the run-time ($RT$), and is given by 

\begin{equation}
\linebreak \sum_{i=0}^{TL} RT_i [N_{CONV}, S_{CONV}, N_{NORM},N_{POOL},N_{FC}]  \linebreak
\end{equation}
\\
subject to digital signal processing (DSP) slices, logic, and memory  constraints, where $TL$ represents the total number of CNN layers  including the
repeated layers.
The convolution layer run-time (${RT}_{CONV}$) is analytically modeled as a function of design variables as  

\begin{equation}
 {RT}_{{CONV}_i} = \frac {\#~of~Convolution~Ops_i}{N_{CONV}\times S_{CONV}\times Frequency} 
\end{equation}

As for the other layers, that is,  normalization, pooling, and fully connected, the following general model is proposed

\begin{equation}
 {RT}_{{Layer}_i} = \frac{\#~of~Layer~Ops_i}{Unroll~factor\times Frequency} 
\end{equation} 

The above  analytical models are later validated by performing full synthesis at selective points and running them on the FPGA accelerator. 

Clearly, in order to determine the best values of the discussed design variables, exhaustive search, especially if the number of variables and or FPGA resources is large,  is infeasible. We have to resort to iterative non-deterministic heuristics~\cite{sait1999iterative}  such as simulated annealing, simulated evolution, tabu search, genetic algorithm, particle swarm optimization, cuckoo search, etc.,  or any of the modern metaheuristics, to efficiently traverse the search space to find acceptable solutions. 

 The proposed methodology employing genetic algorithm was demonstrated by optimizing the implementation of two representative  CNNs, AlexNet and VGG,  on two Altera Stratix-V FPGA platforms, DE5-Net and P395-D8 boards, both of which have different hardware resources. Peak performance is achieved for both, for the  convolution operations, and for the entire CNN network.

One major issue related to use of  non-deterministic iterative heuristics in the design of neural networks and CNNs is the large amount of memory  required to store the state of solution and the amount of time taken to determine the cost  of the solution, be it accuracy/error estimation, run-time, or any other objective. Reasonable estimation techniques and analytical formulations  are required to efficiently traverse the design space in search of efficient  solutions.

\section{Summary and Recommendations} \label{sec:Summary_and_Recommendations}

In this section, we   highlight the key features discussed in the acceleration of convolutional neural networks (CNNs) implemented on FPGAs, and  provide recommendations to enhance the effectiveness of employing
FPGAs in the acceleration of CNNs.

All reviewed techniques are centered around accelerating the convolution (CONV) operation as it consumes around
90\% of the computational time. This is achieved by utilizing parallel multiply-accumulate operations bounded
by resource limitations. In addition, careful design of data access patterns are targeted to minimize
the memory bandwidth requirements utilizing internal memory structures and maximizing data reuse. This is crucial
in the acceleration process due to the large memory data that needs to be accessed including feature maps (FMs) and weights.
To minimize the memory footprint and to achieve effective utilization of resources, some techniques optimize the
number of bits used to represent the feature maps and weights with minimal impact on accuracy. This is combined with
the optimized selection of the number of fraction bits used for each layer.  Other techniques
optimize the number of used weights in the fully connected (FC) layers as they are memory-intensive.
Coprocessors are also employed to automatically configure both the software and the hardware elements to fully exploit  parallelism~\cite{chakradhar2010dynamically}.

To optimize parallelization  of convolution operations, several approaches have been attempted.
Work load analysis has been tried to determine   computations that can be structured as parallel streams~\cite{cadambi2010programmable}.  The roofline model based accelerator    uses  polyhedral-based data dependence analysis to find the optimal unrolling factor for every convolutional layer~\cite{williams2009roofline}, and to fully utilize all FPGA computational resources through loop pipelining.
To optimize performance, tiled matrix multiplication is structured as a pipelined binary adder tree for performing multiplication and generating partial sums~\cite{jtag2018altera}.
An optimization framework has been proposed by Suda et al.  who identified the key variables  of the design~\cite{suda2016throughput} and optimize them to maximize parallelism. 

To reduce computational complexity of CONV layers and improve resource efficiency, a number of 
approaches such as~\cite{lavin2016fast, aydonat2017opencl, lu2017evaluating} utilized  Winograd transformation in performing CONV operations as this  reduces the computational complexity by around $50\%$.

To maximize throughput, several techniques such as \cite{venieris2016fpgaconvnet, liu2017throughput, shen2017maximizing} have used  multiple CONV layer processors (CLPs) instead of using a single CLP
that is optimized for all CONV layers. This pipelines the operation of the multiple CLPs achieving layer-level parallelism 
which maximizes resource utilization and enhances performance in comparison to using a single CLP.

Since the computational requirement of FC layers is significantly less than that of CONV layers, to improve performance, and
maximize resource utilization, a number of techniques such as~\cite{zhang2016caffeine, zhang2018caffeine, aydonat2017opencl, lu2017evaluating} create batches by grouping different input FMs and processing them together in FC layers.

Complex access patterns and data locality are  used in DeepBurning tool~\cite{wang2016deepburning} for better data reuse. In~\cite{wang2017dlau}, the authors explored hot spots profiling to determine the computational parts that need to be accelerated to improve the performance. Acceleration is accomplished by reducing the memory bandwidth requirements. Techniques proposed exploit data reuse to reduce off-chip memory communications. Loop transformations have also been used by reducing tiling parameters to improve data locality, and to reduce redundant communication operations to maximize the data sharing/reuse. 

Efficient buffering, where the weight buffers are used to ensure the availability of CONV and FC layers’ weights before their computation, as well as to overlap the transfer of FC layer weights with its computation, helps in improving performance~\cite{ma2016scalable, ma2018alamo}. In the Catapult project, FPGA boards were integrated into data center applications and achieved speedup. Microsoft Research’s Catapult utilized multi-banked input buffer and kernel weight buffer to provide an efficient buffering scheme of feature maps and weights, respectively. To minimize the off-chip memory traffic, a specialized network on-chip was designed to re-distribute the output feature maps on the multi-banked input buffer instead of transferring them to the external memory~\cite{ovtcharov2015accelerating}.

To further reduce memory footprint and bandwidth requirement, optimal fractional length for weights
and feature maps in each layer  are used. Singular value decomposition (SVD) has also been applied to the weight matrix of FC layer in order to reduce memory footprint at this layer~\cite{qiu2016going}. 
Tiling techniques have been proposed where large-scale input data is partitioned into small subsets or tiles whose size is configured to leverage the trade-off between the hardware cost and the speedup~\cite{wang2017dlau}. 
 
Automation tools have been developed that automatically build neural networks with optimized performance~\cite{wang2016deepburning}. They employ pre-constructed register transfer level (RTL) module library that holds hardware (including logical and arithmetic operations) and configuration scripts. DeepBurning, for example, generates the hardware description for neural network scripts. Another modularized RTL compiler,   ALAMO, integrates both the RTL finer level optimization and the flexibility of high-level synthesis (HLS) to generate efficient Verilog parameterized RTL scripts for ASIC or FPGA platform under the available number of parallel computing resources (i.e., the number of multipliers)~\cite{ma2016scalable, ma2018alamo}. Acceleration is achieved by employing loop unrolling technique for CONV layer operations. Some of the reviewed techniques also help minimize the size of FPGA on-chip memories to optimize energy and area usage~\cite{peemen2013memory},~\cite{beric2008memory}.

In Table \ref{tab_comparison_tech_1} and  Table \ref{tab_comparison_tech_2}, we list the optimization mechanisms utilized by each of the reviewed techniques to maximize performance 
and throughput of FPGA-based deep learning networks.

\begin{sidewaystable*}[ph!]
	\renewcommand{\arraystretch}{1}
	\centering
	\captionsetup{justification=centering}
	\caption{Optimization Mechanisms Employed for FPGA-based Acceleration of Deep Learning Networks.}
	\resizebox{\textwidth}{!}{%
	\begin{tabular}{|l||c|c|c|c|c|c|c|c|c|c|c|c|c|c|}
		\hline
		\textbf{Technique} & 
		\textbf{\makecell{VIP\\\cite{cloutier1996vip}}} & 
		\textbf{\makecell{CNP\\\cite{farabet2009cnp}}} & 
		\textbf{\makecell{CONV\\Coprocessor\\Accelerator~\cite{sankaradas2009massively}}} & 
		\textbf{\makecell{MAPLE\\\cite{cadambi2010programmable}}} & 
		\textbf{\makecell{DC-CNN\\\cite{chakradhar2010dynamically}}} & 
		\textbf{\makecell{NeuFlow\\\cite{farabet2011neuflow}}} & 
		\textbf{\makecell{Memory-Centric\\Accelerator\\\cite{peemen2013memory}}} & 
		\textbf{\makecell{nn-X\\\cite{gokhale2014240}}} & 
		\textbf{\makecell{Roofline-based\\FPGA\\Accelerator~\cite{zhang2015optimizing}}} &
		\textbf{\makecell{Embedded\\FPGA\\Accelerator~\cite{qiu2016going}}} & 
		\textbf{\makecell{DeepBurning\\\cite{wang2016deepburning}}} & 
		\textbf{\makecell{OpenCL-based\\FPGA\\Accelerator~\cite{suda2016throughput}}} & 
		\textbf{\makecell{Caffeine\\\cite{zhang2016caffeine, zhang2018caffeine}}} & 
		\textbf{\makecell{fpgaConvNet\\\cite{venieris2016fpgaconvnet}}}\\
		\hline
		\hline
		\textbf{\makecell[l]{Loop\\Unrolling}} &  &  &  &  &  &  &  &  & $\bigtimes$ & &  & $\bigtimes$ & $\bigtimes$ &  $\bigtimes$ \\ 
		\hline
		\textbf{\makecell[l]{Loop\\Tiling}} & $\bigtimes$ &  &  &  &  & $\bigtimes$ & $\bigtimes$ &  & $\bigtimes$ & $\bigtimes$ & $\bigtimes$ &  & $\bigtimes$ &  \\ 
		\hline
		\textbf{\makecell[l]{Loop\\Interchange}} &  &  &  &  &  &  & $\bigtimes$ &  &  & &  &  &  &  \\ 
		\hline
		\multirow{2}{*}{\textbf{Pipelining}} & \multirow{2}{*}{$\bigtimes$} & \multirow{2}{*}{$\bigtimes$} & \multirow{2}{*}{$\bigtimes$} & \multirow{2}{*}{$\bigtimes$} & \multirow{2}{*}{$\bigtimes$} & \multirow{2}{*}{$\bigtimes$} & \multirow{2}{*}{$\bigtimes$} & \multirow{2}{*}{$\bigtimes$} & \multirow{2}{*}{$\bigtimes$} & \multirow{2}{*}{} & \multirow{2}{*}{} & \multirow{2}{*}{$\bigtimes$} & \multirow{2}{*}{$\bigtimes$} & \multirow{2}{*}{$\bigtimes$} \\ 
		 &  &  &  &  &  &  &  &  &  & &  &  &  &  \\ 
		\hline
		\multirow{2}{*}{\textbf{Batching}} & \multirow{2}{*}{} & \multirow{2}{*}{} & \multirow{2}{*}{} & \multirow{2}{*}{} & \multirow{2}{*}{} & \multirow{2}{*}{} & \multirow{2}{*}{} & \multirow{2}{*}{} & \multirow{2}{*}{} & \multirow{2}{*}{} & \multirow{2}{*}{} & \multirow{2}{*}{} & \multirow{2}{*}{$\bigtimes$} & \multirow{2}{*}{} \\ 
		 &  &  &  &  &  &  &  &  &  & &  &  &  &  \\ 
		 \hline
		\multirow{2}{*}{\textbf{Multi-CLPs}} & \multirow{2}{*}{} & \multirow{2}{*}{} & \multirow{2}{*}{} & \multirow{2}{*}{} & \multirow{2}{*}{} & \multirow{2}{*}{} & \multirow{2}{*}{} & \multirow{2}{*}{} & \multirow{2}{*}{} & \multirow{2}{*}{} & \multirow{2}{*}{} & \multirow{2}{*}{} & \multirow{2}{*}{} & \multirow{2}{*}{$\bigtimes$} \\ 
		 &  &  &  &  &  &  &  &  &  & &  &  &  &  \\ 
		\hline
		\textbf{\makecell[l]{Fixed-Point\\Precision}} & $\bigtimes$ & $\bigtimes$ & $\bigtimes$ & $\bigtimes$ & $\bigtimes$ & $\bigtimes$ & $\bigtimes$ & $\bigtimes$ &  & $\bigtimes$ & $\bigtimes$ & $\bigtimes$ & $\bigtimes$ & $\bigtimes$ \\ 
		\hline
		\textbf{\makecell[l]{Per-Layer\\Quantization}} &  &  &  &  &  &  &  &  &  & $\bigtimes$ &  &  &  &  \\ 
		\hline
		\textbf{\makecell[l]{Singular Value\\Decomposition}} &  &  &  &  &  &  &  &  &  & $\bigtimes$ &  &  &  &  \\ 
		\hline
		\multirow{2}{*}{\textbf{Prefetching}} & \multirow{2}{*}{} & \multirow{2}{*}{} & \multirow{2}{*}{} & \multirow{2}{*}{} & \multirow{2}{*}{$\bigtimes$} & \multirow{2}{*}{$\bigtimes$} & \multirow{2}{*}{} & \multirow{2}{*}{} & \multirow{2}{*}{} & \multirow{2}{*}{} & \multirow{2}{*}{} & \multirow{2}{*}{} & \multirow{2}{*}{$\bigtimes$} & \multirow{2}{*}{} \\ 
		 &  &  &  &  &  &  &  &  &  & &  &  &  &  \\ 
		\hline
		\textbf{\makecell[l]{Rearranging\\Memory Data}} &  &  &  &  &  &  &  &  &  & $\bigtimes$ & $\bigtimes$ & $\bigtimes$ & $\bigtimes$ &  \\ 
		\hline
		\textbf{\makecell[l]{In-Memory\\Processing}} &  &  &  & $\bigtimes$ &  &  &  &  &  & &  &  &  &  \\ 
		\hline
		\textbf{\makecell[l]{Line\\Buffer}} &  &  &  &  &  &  &  &  &  & $\bigtimes$ &  &  &  &  \\ 
		\hline
		\textbf{\makecell[l]{Double\\Buffering}} &  &  &  &  &  &  &  &  & $\bigtimes$ & &  &  & $\bigtimes$ &  \\ 
		\hline
		\textbf{\makecell[l]{Approximating\\Non-Linear AF}} &  & $\bigtimes$ & $\bigtimes$ &  &  & $\bigtimes$ &  & $\bigtimes$ &  & & $\bigtimes$ & $\bigtimes$ &  & $\bigtimes$ \\ 
		\hline
		\textbf{\makecell[l]{Eliminating\\FC Layer}} &  &  &  $\bigtimes$ &  & $\bigtimes$ &  &  & $\bigtimes$ & $\bigtimes$ & &  &  &  &  \\ 
		\hline
		\textbf{\makecell[l]{Roofline\\Model}} &  &  &  &  &  &  &  &  & $\bigtimes$ & &  &  & $\bigtimes$ &  \\ 
		\hline
		\textbf{\makecell[l]{Polyhedral\\Optimization}} &  &  &  &  &  &  &  &  & $\bigtimes$ & &  &  & $\bigtimes$ &  \\ 
		\hline
		\textbf{\makecell[l]{Dynamic\\Programming}} &  &  &  &  & $\bigtimes$ &  &  &  &  & &  &  &  &  \\ 
		\hline
		\textbf{\makecell[l]{Graph\\Partitioning}} &  &  &  &  &  &  &  &  &  & &  &  &  & $\bigtimes$ \\ 
		\hline
\end{tabular}}
\label{tab_comparison_tech_1}
\end{sidewaystable*}

\begin{sidewaystable*}[ph!]
	\renewcommand{\arraystretch}{1}
	\centering
	\captionsetup{justification=centering}
	\caption{Optimization Mechanisms Employed for FPGA-based Acceleration of Deep Learning Networks.}
	\resizebox{\textwidth}{!}{%
	\begin{tabular}{|l||c|c|c|c|c|c|c|c|c|c|c|c|c|c|c|c|c|}
		\hline
		\textbf{Technique} & 
		\textbf{\makecell{ALAMO\\\cite{ma2016scalable, ma2018alamo}}} & 
		\textbf{\makecell{Throughput-\\Optimized\\FPGA\\Accelerator~\cite{liu2017throughput}}} & 
		\textbf{\makecell{FP-DNN\\\cite{guan2017fp}}} & 
		\textbf{\makecell{FINN\\\cite{umuroglu2017finn}}} & 
		\textbf{\makecell{Customized\\CONV Loop\\Accelerator\\\cite{ma2017optimizing}}} & 
		\textbf{\makecell{Latency-Driven\\Design for\\FPGA-based\\CNNs~\cite{venieris2017latency}}} & 
		\textbf{\makecell{DLA\\\cite{aydonat2017opencl}}} & 
		\textbf{\makecell{Winograd-\\based CNN\\Accelerator\\\cite{lu2017evaluating}}} & 
		\textbf{\makecell{OpenCL-based\\Architecture for\\Accelerating\\CNNs~\cite{zhang2017improving}}} &
		\textbf{\makecell{Multi-CLP\\Accelerator\\for CNNs\\\cite{shen2017maximizing}}} & 
		\textbf{\makecell{Automated\\Systolic Array\\Architecture\\for CNN~\cite{wei2017automated}}} & 
		\textbf{\makecell{End-to-End\\Scalable FPGA\\Accelerator\\\cite{ma2017end}}} & 
		\textbf{\makecell{DLAU\\\cite{wang2017dlau}}} & 
		\textbf{\makecell{An Automatic\\RTL Compiler for\\High-Throughput\\Deep CNNs~\cite{ma2017automatic}}} &
		\textbf{\makecell{Intel's\\DLA\\\cite{abdelfattah2018dla}}} & 
		\textbf{\makecell{Angel-Eye\\\cite{guo2018angel}}} & 
		\textbf{\makecell{Optimizing the\\CONV Operation\\to Accelerate DNNs\\on FPGA~\cite{ma2018optimizing}}} \\
		\hline
		\hline
		\textbf{\makecell[l]{Loop\\Unrolling}} & $\bigtimes$ & $\bigtimes$ &  &  & $\bigtimes$ & $\bigtimes$ &  & $\bigtimes$ &  & $\bigtimes$ &  & $\bigtimes$ &  & $\bigtimes$ & & & $\bigtimes$\\ 
		\hline
		\textbf{\makecell[l]{Loop\\Tiling}} &  &  & $\bigtimes$ & $\bigtimes$ & $\bigtimes$ &  & $\bigtimes$ & $\bigtimes$ &  & $\bigtimes$ & $\bigtimes$ & $\bigtimes$ & $\bigtimes$ & $\bigtimes$ & & & $\bigtimes$\\ 
		\hline
		\textbf{\makecell[l]{Loop\\Interchange}} &  &  &  &  & $\bigtimes$ &  &  &  &  & &  &  &  &  & & & $\bigtimes$ \\ 
		\hline
		\multirow{2}{*}{\textbf{Pipelining}} & \multirow{2}{*}{$\bigtimes$} & \multirow{2}{*}{$\bigtimes$} & \multirow{2}{*}{} & \multirow{2}{*}{$\bigtimes$} & \multirow{2}{*}{} & \multirow{2}{*}{$\bigtimes$} & \multirow{2}{*}{$\bigtimes$} & \multirow{2}{*}{$\bigtimes$} & \multirow{2}{*}{} & \multirow{2}{*}{$\bigtimes$} & \multirow{2}{*}{$\bigtimes$} & \multirow{2}{*}{} & \multirow{2}{*}{$\bigtimes$} & \multirow{2}{*}{} & \multirow{2}{*}{} & \multirow{2}{*}{} & \multirow{2}{*}{} \\ 
		&  &  &  &  &  &  &  &  &  & &  &  &  &  & & &\\  
		\hline
		\textbf{\makecell[l]{Input\\Batching}} &  &  &  &  &  &  &  &  &  & $\bigtimes$ &  &  &  &  & & &\\  
		 \hline
		 \textbf{\makecell[l]{FC Layer\\Batching}} &  &  & $\bigtimes$ &  &  &  & $\bigtimes$ & $\bigtimes$ &  & &  &  &  &  & & &\\  
		 \hline
		 \multirow{2}{*}{\textbf{Multi-CLPs}} & \multirow{2}{*}{} & \multirow{2}{*}{$\bigtimes$} & \multirow{2}{*}{} & \multirow{2}{*}{} & \multirow{2}{*}{} & \multirow{2}{*}{} & \multirow{2}{*}{} & \multirow{2}{*}{} & \multirow{2}{*}{} & \multirow{2}{*}{$\bigtimes$} & \multirow{2}{*}{} & \multirow{2}{*}{} & \multirow{2}{*}{} & \multirow{2}{*}{} & \multirow{2}{*}{} & \multirow{2}{*}{$\bigtimes$} & \multirow{2}{*}{} \\ 
		&  &  &  &  &  &  &  &  &  & &  &  &  &  & & &\\  
		\hline
		\textbf{\makecell[l]{Binarized\\CNN}} &  &  &  & $\bigtimes$ &  &  &  &  &  & &  &  &  &  & & &\\ 
		\hline
		\textbf{\makecell[l]{Fixed-Point\\Precision}} & $\bigtimes$ &  $\bigtimes$ &  $\bigtimes$ &  & $\bigtimes$ & $\bigtimes$ &  & $\bigtimes$ & $\bigtimes$ & $\bigtimes$ & $\bigtimes$ & $\bigtimes$ &  & $\bigtimes$ & & $\bigtimes$ & $\bigtimes$\\ 
		\hline
		\textbf{\makecell[l]{Per-Layer\\Quantization}} &  &  &  &  &  &  &  &  &  & &  &  &  &  & & $\bigtimes$ & $\bigtimes$\\  
		\hline
		\multirow{2}{*}{\textbf{Prefetching}} & \multirow{2}{*}{$\bigtimes$} & \multirow{2}{*}{} & \multirow{2}{*}{} & \multirow{2}{*}{} & \multirow{2}{*}{} & \multirow{2}{*}{} & \multirow{2}{*}{$\bigtimes$} & \multirow{2}{*}{$\bigtimes$} & \multirow{2}{*}{} & \multirow{2}{*}{} & \multirow{2}{*}{} & \multirow{2}{*}{} & \multirow{2}{*}{} & \multirow{2}{*}{} & \multirow{2}{*}{$\bigtimes$} & \multirow{2}{*}{} & \multirow{2}{*}{} \\ 
		&  &  &  &  &  &  &  &  &  & &  &  &  &  & & &\\  
		\hline
		\textbf{\makecell[l]{Rearranging\\Memory Data}} &  &  &  &  &  &  &  &  & $\bigtimes$ & &  & $\bigtimes$ &  &  & & &\\ 
		\hline
		\textbf{\makecell[l]{Line\\Buffer}} &  &  &  &  &  &  &  & $\bigtimes$ & $\bigtimes$ & &  &  &  &  & & & $\bigtimes$ \\ 
		\hline
		\textbf{\makecell[l]{Double\\Buffering}} &  &  & $\bigtimes$ &  & $\bigtimes$ &  & $\bigtimes$ & $\bigtimes$ &  & $\bigtimes$ & $\bigtimes$ &  & $\bigtimes$ &  & $\bigtimes$ & &\\  
		\hline
		\textbf{\makecell[l]{Padding\\Optimizations}} &  &  &  &  &  &  &  &  &  & &  & $\bigtimes$ &  &  & & & $\bigtimes$ \\ 
		\hline
		\textbf{\makecell[l]{Winograd\\Algorithm}} &  &  &  &  &  &  & $\bigtimes$ & $\bigtimes$ &  & &  &  &  &  & & &\\
		\hline
		\textbf{\makecell[l]{Approximating\\Non-Linear AF}} &  &  &  & $\bigtimes$ &  & $\bigtimes$ &  &  &  & &  &  & $\bigtimes$ &  & & &\\
		\hline
		\textbf{\makecell[l]{Roofline\\Model}} &  &  &  & $\bigtimes$ &  &  &  &  & $\bigtimes$ & &  &  &  &  & & &\\ 
		\hline
		\textbf{\makecell[l]{Polyhedral\\Optimization}} &  &  &  &  &  &  &  &  &  & & $\bigtimes$ &  &  &  & & &\\ 
		\hline
		\textbf{\makecell[l]{Dynamic\\Programming}} &  &  &  &  &  &  &  &  &  & $\bigtimes$ &  &  &  &  & & &\\ 
		\hline
		\textbf{\makecell[l]{Graph\\Coloring}} &  &  & $\bigtimes$ &  &  &  &  &  &  & &  &  &  &  & & &\\ 
		\hline
		\textbf{\makecell[l]{Graph\\Partitioning}} &  &  &  &  &  & $\bigtimes$ &  &  &  & &  &  &  &  & $\bigtimes$ & &\\ 
		\hline
		\textbf{\makecell[l]{Pattern\\Matching}} &  &  &  &  &  & $\bigtimes$ &  &  &  & &  &  &  &  & & &\\ 
		\hline
\end{tabular}}
\label{tab_comparison_tech_2}
\end{sidewaystable*}

To enhance utilization of FPGAs in CNNs acceleration and to maximize their effectiveness, we recommend
the development of a framework that includes a user-friendly interface that allows the user to
easily specify the CNN model to be accelerated. This includes specifying the CNN model parameters in terms
of number of convolution layers and their sizes, and number of fully connected layers along with other intermediate operations. The specified CNN model weights will be read from a file. 
In addition, the user should have the option of specifying the 
FPGA platform that will be used for implementing the CNN accelerator and the maximum 
tolerable error, along with the selection of a library from a set of applications to be used for model optimization
and evaluation. The framework then should perform optimizations to find the minimum 
number of bits that need to be used for representing the weights and feature maps and the number of fraction
bits to be used for each layer. In addition, optimization
of fully connected layers is performed to minimize the memory requirements. All such optimizations are
carried out bounded by the maximum error specified by the user for the specified application library.

The framework should be designed based on the development of a scalable hardware architecture that works
for any given FPGA platform and achieves higher speedup with the availability of higher resources.
Based on the available resources, specified by the FPGA platform, the tool will perform optimizations
to maximize parallelism and data reuse, given the resource limitations.
The tool will then automatically generate the CNN model that will fit on the given FPGA platform and will
allow the user to evaluate the performance based on the chosen application library. This will allow
the user to evaluate the performance gains while evaluating different FPGA platforms with different resources.
The tool should have the option to generate performance measures based on different performance metrics
as selected by the user such as number of frames processed per second or number of operations performed per
second. In addition, the tool will report other design metrics such as resource utilization, memory
sizes and bandwidth, and power dissipation.

Furthermore, it is desired to have the option for the user to specify the desired performance 
for a given CNN model and have the tool perform necessary analysis and evaluation and recommend to the user
candidate FPGA platforms for achieving the desired performance levels. This will require the development of 
reasonably accurate analytical  models that will estimate the needed resources for achieving the desired
performance. The user can then choose the recommended FPGA platform  and perform complete evaluation to verify that
the desired performance levels are met.

\section{Conclusion} \label{sec:Conclusion}

In this paper, we  reviewed  recent developments in the area of acceleration of deep learning networks and, in particular, convolution  neural  networks (CNNs) on field programmable gate arrays (FPGAs). 
The paper begins with a brief overview of deep learning techniques highlighting their importance, key operations, and applications.  Special emphasis is given on CNNs as they have wide applications in the area of   image detection and recognition and  require both CPU and memory intensive operations that can be 
effectively accelerated utilizing FPGA inherent ability to maximize parallelism of operations.

While the paper  briefly touches upon the acceleration techniques for deep learning algorithms and CNNs from both software and hardware perspectives, the core of this article has been the review of recent techniques employed in the acceleration of CNNs on FPGAs. A thorough up-to-date review is provided that illustrates the employment of various possibilities and techniques such as exploitation of parallelism utilizing loop tiling and loop unrolling, 
effective use of internal memory to maximize data reuse, operation pipelining, 
and effective use of data sizes to minimize memory footprint, and, to optimize FPGA resource utilization.

The paper  also presented the use of tools for  generating register transfer level (RTL) scripts that not only help in automating the design process, but also help in exploring the design space and suggesting efficient hardware.  The paper   discusses the use of analytics such as: (i) work load analysis in determining the computations that can be parallelized, (ii) optimal loop unrolling factors, (iii) determining access patterns to improve data locality, etc.  In addition, a brief review of the use of non-deterministic heuristics in solving NP-hard combinatorial optimization problems in the design and
implementation of CNNs has been presented.  
Finally, the paper   summarizes the key features employed by the various FPGA-based CNN acceleration techniques
and provided recommendations for enhancing the effectiveness of utilizing FPGAs in CNNs acceleration.
\section*{Acknowledgment}

Authors acknowledge  King Fahd University of Petroleum \& Minerals, Dhahran, Saudi Arabia  for all support.
We also like to acknowledge Dr. Blair P. Bremberg and Ms. Sumaiya Hussain Sadiq for their help in professional English editing of this manuscript.



\begin{IEEEbiography}[{\includegraphics[width=1in,height=1.25in,clip,keepaspectratio]{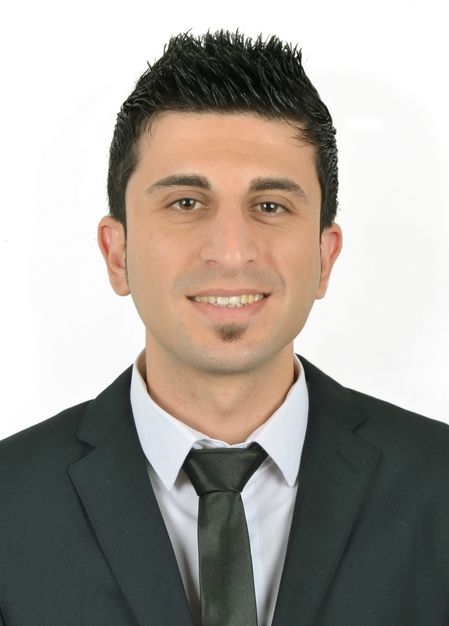}}]{Ahmad Shawahna} obtained M.S. in computer engineering from King Fahd University of Petroleum and Minerals (KFUPM), Saudi Arabia, in 2016. He also received the B.Sc degree in computer engineering from An-Najah National University, Palestine, in 2012. Ahmad Shawahna is a Ph.D. student in the department of computer engineering of KFUPM. In addition, he is currently working at the Center for Communications and IT Research (CCITR), KFUPM. His research interests include hardware accelerator, deep learning, CNNs, FPGA, wireless security, network security, Internet of Things (IoT), and cloud computing.
\end{IEEEbiography}


\begin{IEEEbiography}[{\includegraphics[width=1in,height=1.65in,clip,keepaspectratio]{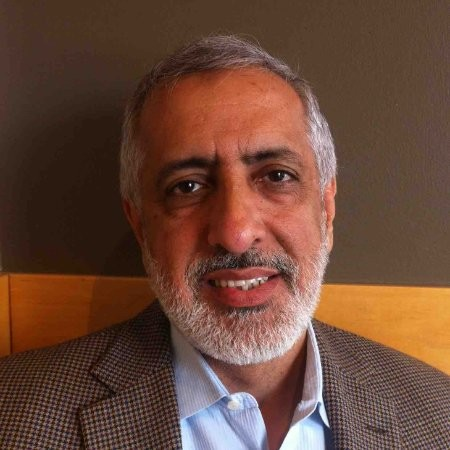}}]{Sadiq M. Sait} (Senior Member, IEEE) obtained his Bachelor's degree in Electronics Engineering from Bangalore University in 1981, and Master's and Ph.D. degrees in Electrical Engineering from KFUPM in 1983 and 1987, respectively. In 1981 Sait received the best Electronic Engineer award from the Indian Institute of Electrical Engineers, Bangalore (where he was born). Sait has authored over 200 research papers, contributed chapters to technical books, and lectured in over 25 countries. Sadiq M. Sait is also the principle author of two books. He is currently Professor of Computer Engineering and the Director of the Center for Communications and IT Research of KFUPM.
\end{IEEEbiography}


\begin{IEEEbiography}[{\includegraphics[width=1in,height=1.25in,clip,keepaspectratio]{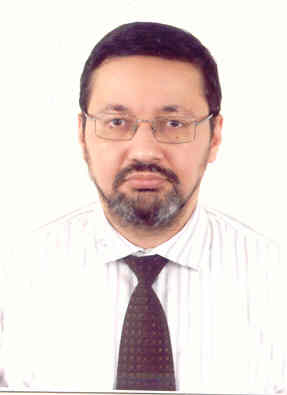}}]{Aiman El-Maleh}
is a Professor in the Computer Engineering Department at King Fahd University of Petroleum \& Minerals. He holds a B.Sc. in Computer Engineering, with first honors, from King Fahd University of Petroleum \& Minerals in 1989, a M.A.SC. in Electrical Engineering from University of Victoria, Canada, in 1991, and a Ph.D in Electrical Engineering, with dean’s honor list, from McGill University, Canada, in 1995. He was a member of scientific staff with Mentor Graphics Corp., a leader in design automation, from 1995-1998. Dr. El-Maleh received the Excellence in Teaching award from KFUPM in 2001/2002, 2006/2007 and 2011/2012, the Excellence in Advising award from KFUPM in 2013/2014 and 2017/2018, the Excellence in Research award from KFUPM in 2010/2011 and 2015/2016, and the First Instructional Technology award from KFUPM in 2009/2010. Dr. El-Maleh's research interests are in the areas of synthesis, testing, and verification of digital systems. In addition, he has research interests in defect and soft-error tolerance design, VLSI design, design automation and efficient FPGA implementations of deep learning algorithms and data compression techniques. Dr. El-Maleh is the winner of the best paper award for the most outstanding contribution in the field of test at the 1995 European Design \& Test Conference. His paper presented at the 1995 Design Automation Conference was also nominated for best paper award. He holds five US patents.
\end{IEEEbiography}

\EOD
\end{document}